\documentclass[sn-mathphys-num]{sn-jnl}

\usepackage{graphicx}%
\usepackage{multirow}%
\usepackage{amsmath,amssymb,amsfonts}%
\usepackage{amsthm}%
\usepackage{mathrsfs}%
\usepackage[title]{appendix}%
\usepackage{xcolor}%
\usepackage{textcomp}%
\usepackage{manyfoot}%
\usepackage{booktabs}%
\usepackage{algorithm}%
\usepackage{algorithmicx}%
\usepackage{algpseudocode}%
\usepackage{listings}%
\usepackage{pgfplots}
\usepackage{natbib}

\usepackage{subcaption} 

\theoremstyle{thmstyleone}%
\theoremstyle{thmstyletwo}%

\theoremstyle{thmstylethree}%

\begin{document}

\title[Article Title]{Stratify: Unifying Multi-Step Forecasting Strategies}


\author*[1]{\fnm{Riku} \sur{Green}}\email{riku.green@bristol.ac.uk}

\author[2,1]{\fnm{Grant} \sur{Stevens}}\email{grant.stevens@bristol.ac.uk}

\author[3]{\fnm{Zahraa} \sur{Abdallah}}\email{zahraa.abdallah@bristol.ac.uk}

\author[3]{\fnm{Telmo} \sur{M. Silva Filho}}\email{telmo.silvafilho@bristol.ac.uk}


\affil*[1]{\orgdiv{School of Computer Science}, \orgname{University of Bristol}, \orgaddress{\street{Merchant Venturers Building, Woodland Road}, \city{Bristol}, \postcode{BS8 1UB}, \country{United Kingdom}}}

\affil[2]{\orgdiv{School of Physics}, \orgname{University of Bristol}, \orgaddress{\street{HH Wills Physics Laboratory, Tyndall Avenue}, \city{Bristol}, \postcode{BS8 1TL}, \country{United Kingdom}}}

\affil[3]{\orgdiv{School of Engineering Mathematics and Technology}, \orgname{University of Bristol}, \orgaddress{\street{Ada Lovelace Building, Tankard's Close}, \city{Bristol}, \postcode{BS8 1TW}, \country{United Kingdom}}}




\abstract{
A key aspect of temporal domains is the ability to make predictions multiple time steps into the future, a process known as multi-step forecasting (MSF).
At the core of this process is selecting a forecasting strategy, however, with no existing frameworks to map out the space of strategies, practitioners are left with ad-hoc methods for strategy selection.
In this work, we propose \textit{Stratify}, a parameterised framework that addresses multi-step forecasting, unifying existing strategies and introducing novel, improved strategies. 
We evaluate \textit{Stratify} on 18 benchmark datasets, five function classes, and short to long forecast horizons (10, 20, 40, 80). In over \textbf{$~84\%$ of 1080 experiments}, novel strategies in \textbf{\textit{Stratify} improved performance} compared to all existing ones.
Importantly, we find that no single strategy consistently outperforms others in all task settings, highlighting the need for practitioners explore the \textit{Stratify} space to carefully search and select forecasting strategies based on task-specific requirements.
Our results are the most comprehensive benchmarking of known and novel forecasting strategies. We make \href{https://github.com/zs18656/Stratify_MSF}{code available} to reproduce our results.
}




\maketitle

\section{Introduction}\label{intro}

\begin{figure}[hbt!]
    \centering
    \includegraphics[width=\linewidth]{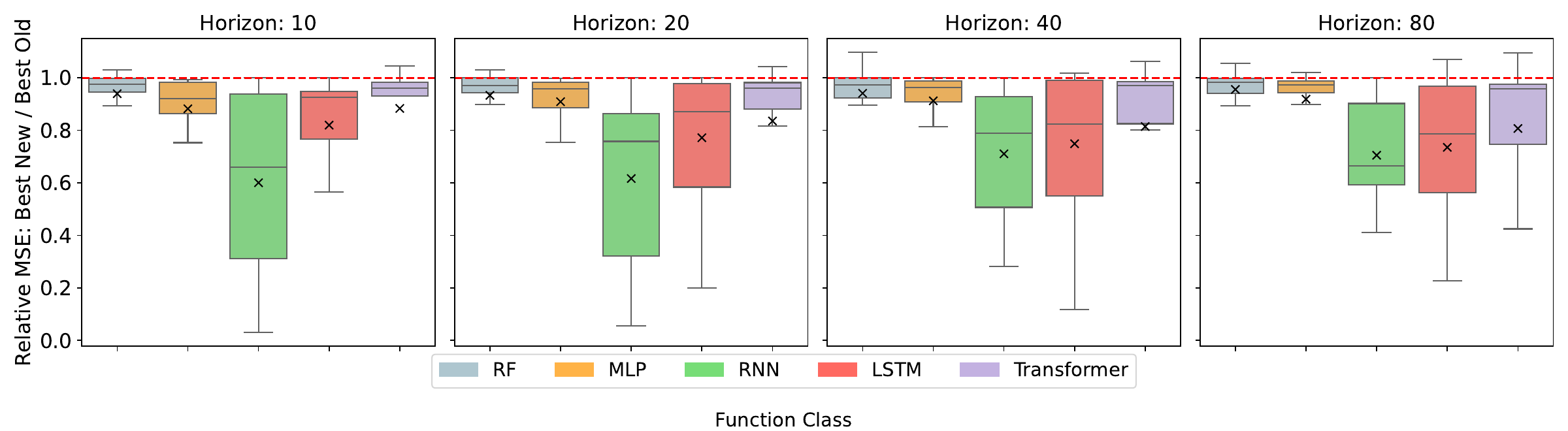}
    \caption{Relative MSE over 18 Benchmark Datasets and multistep horizons 10, 20, 40, 80 for five function classes. Exploring \textit{Stratify}'s novel strategies consistently outperform existing strategies within the unified space. The best performing novel strategy is compared to the best performing existing strategy, both of which are accessible in the \textit{Stratify} space. Consistent relative MSE below the dashed line show that exploring \textit{Stratify} is beneficial across short to long horizons.}
    \label{fig:firstboxplots}
\end{figure}

\noindent Time series forecasting plays a critical role in numerous real-world applications such as in healthcare \citep{morid2023time_health}, transport networks \citep{nguyen2018deep_transport}, geographical systems \citep{rajagukguk2020review_geog}, and financial markets \citep{sezer2020financialsurvey}. Multistep forecasting, which involves predicting a consecutive sequence of future time steps, remains a significant challenge in time series analysis \citep{masini2023machine_advances_good, theoryandpractice}. Multistep forecasting (MSF) strategies have consistently received attention in the time series literature given their necessity for long-term predictions in any dynamic domain \citep{lim2021timesurvey, Ji2017Recmo, noa2024dirrecmo}. 

The unique complexity of multistep forecasting is due to the trade-off between variance and bias in selecting a forecasting strategy \citep{taieb2014machine}. Classical analysis of MSF concerns when it is appropriate to incorporate a recursive strategy (high bias) or a direct strategy (high variance). The recursive strategy predicts auto-regressively on a single model's own predictions until the desired horizon length is obtained. In contrast, direct strategies require fitting separate models to predict each fixed length, which is expensive and often results in model inconsistencies \citep{taieb2014machine}. 

To bridge the gap between recursive and direct strategies, \emph{hybrid strategies} have been developed. These are the DirectRecursive (DirRec) \citep{taieb2012reviewmultistep} and Rectify \citep{rectify} strategies. The multi-input multi-output (MIMO) strategy inspired the development of other multi-output strategies, such as Recursive Multi-output (RecMO) \citep{Ji2017Recmo}, Direct Multi-output (DirMO) \citep{taieb2010multiple}, and DirrecMO \citep{noa2024dirrecmo}. Multi-output (MO) strategies allow for tuning of output dimension of models to find an optimal balance in bias, variance, and computational efficiency. Hybrid methods have been shown to allow for more flexibility and improve the state of the art in MSF \citep{An_comp}. However, which strategy is generally optimal remains an open problem, as it often depends on the domain and function class of the forecasting model \citep{Ji2017Recmo, noa2024dirrecmo, An_comp, taieb2012reviewmultistep}.

The multi-output parameterisation of recursive, direct, and DirRec strategies offers a framework where they become equivalent to MIMO when their parameter value equals the multi-step horizon length \citep{noa2024dirrecmo}. However, little progress has been made to unify or represent MSF strategies. The lack of a unifying framework to represent MSF strategies has precluded a deeper understanding of how the variance and bias induced by the parameter selection of a strategy affects the downstream performance. This leaves practitioners selecting strategies with ad-hoc methods for strategy selection.

In this work, we introduce a novel approach to multi-step forecasting by parameterising and generalising the rectify strategy. This leaves the literature complete in terms of converting widely known single-output strategies into multi-output ones. \textit{Stratify} defines a broader function space for forecasting strategies, encompassing both existing strategies and new ones that have not been previously explored. We highlight the benefits of exploring \textit{Stratify} in Figure \ref{fig:firstboxplots}, where the relative errors of existing methods are compared novel strategies in \textit{Stratify}. By framing these strategies in a parameterised and generalisable function space, we offer practitioners a systematic way to investigate and explore the space of known forecasting strategies for different tasks and datasets. 

Our extensive experiments find that previously unknown strategies, now explored through Stratify, are consistently and often significantly, better performing than the best existing strategies. We make our evaluations on 18 benchmark datasets \citep{shao2024basicts} and multi-step horizon lengths of 10, 20, 40, and 80. 

Our main contributions and novelty of \textit{Stratify} include:
\begin{itemize}
    \item \textbf{A Unified Framework:} \textit{Stratify} is a unified representation of forecasting strategies, facilitating a systematic exploration of all existing strategies as well as novel strategies which can be significantly higher performing.
    \item \textbf{Novel MSF Strategies}: Through Stratify, we discover novel strategies that consistently outperform all existing ones.
    \item \textbf{Improved Performance:} Experimental validation on  18 benchmark datasets and multiple function classes demonstrates that \textit{Stratify} consistently outperforms state-of-the-art strategies across diverse forecast horizons.
    \item \textbf{Optimisation/Visualisation Insights:} We show that the \textit{Stratify} representation of MSF strategy performance is often relatively smooth.  The smoothness of Stratify’s function space highlights the possibility of efficient optimisation over the space, highlighting a further practical utility.
\end{itemize}

\noindent The rest of this paper is organised as follows: Section \ref{relatedwork} presents the related work on multi-step time series forecasting strategies; Section \ref{prelim} covers the preliminaries; Section \ref{method} describes \textit{Stratify}; Section \ref{results} presents our results and experimental setup; and Section \ref{discussion} presents the discussion, future work, and conclusion.

\newpage
\section{Related work}\label{relatedwork}

\begin{figure*}[hbt!]
    \centering
    \includegraphics[width=0.8\linewidth]{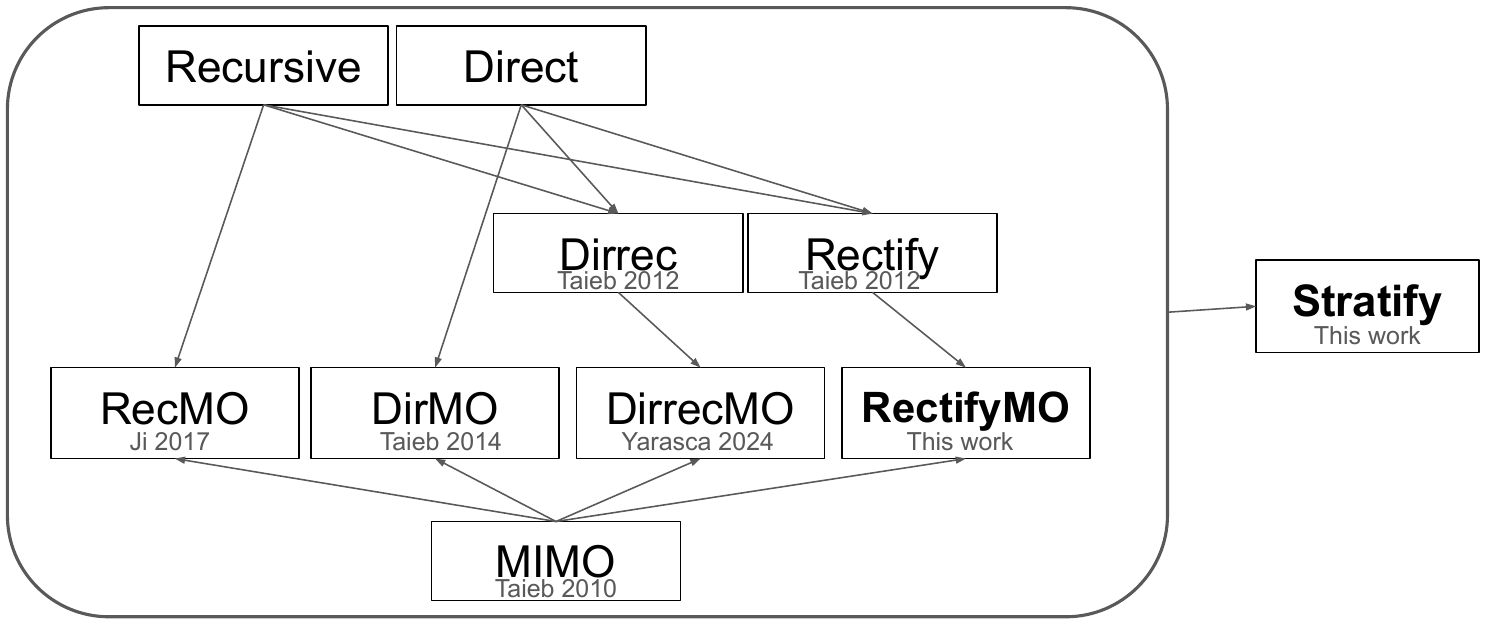}
    \caption{Summary of the strategies in MSF. In bold are our contributions. We extend the single-output Rectify strategy into its multi-output \citep{taieb2010multiple} variant, analogous to RecMO, DirMO \citep{taieb2014machine}, and DirRecMO \citep{noa2024dirrecmo}. \textit{Stratify} is a framework which generalises all existing strategies and introduces novel strategies with improved performance. Lines show the evolution and fusion of previous strategies to form new ones.}
    \label{fig:strategyhistory}
\end{figure*}

Multi-step time series forecasting strategies are designed to predict multiple future points in a sequence, and they have evolved to address various challenges inherent in this task. We show this evolution in Figure \ref{fig:strategyhistory}. The recursive strategy involves training a single model for one-step-ahead forecasting and then iteratively applying it to predict multiple steps ahead by feeding each prediction back into the model as input \citep{taieb2014machine}. While straightforward, this method can suffer from error accumulation over longer horizons, as inaccuracies compound with each step. In contrast, the direct strategy trains separate models for each forecasting horizon, predicting future values directly from observed data. This approach avoids the issue of error propagation but may neglect the dependencies between future time points and results in higher variance, especially when data is limited. 

To overcome the limitations of these basic strategies, hybrid and multiple-output (MO) approaches have been developed. Hybrid strategies like DirRec \citep{taieb2012reviewmultistep} and Rectify \citep{rectify} combine elements of both recursive and direct methods by using multiple models where each model's input includes previous predictions, aiming to balance error accumulation and independence assumptions. The MIMO (Multiple Input Multiple Output) \citep{taieb2014machine} strategy employs a single model to predict the entire sequence of future values simultaneously, preserving dependencies between them and potentially improving efficiency. MO variants such as RecMO, DirMO, and DirRecMO \citep{noa2024dirrecmo} extend the recursive, direct, and DirRec strategies by forecasting multiple future steps at once in segments. 
These methods aim to effectively mitigate error propagation and increase computational efficiency. We contribute a method for parameterising the Rectify strategy, named RectifyMO, and show where our contribution relates to other works in Figure \ref{fig:strategyhistory}.

The recursive strategy is known to be asymptotically biased \citep{brown1984residual}. It is shown that minimising one-step-ahead forecast errors does not guarantee the minimum for multi-step-ahead errors \citep{taieb2014machine}. This highlights that MSF strategies represent an important assumption in the modelling of the underlying data-generating process. Although the direct strategy is shown to be unbiased, as the model objective is identical to the MSF objective, there are no guarantees of consistency within the direct strategy \citep{taieb2014machine}. Rectify was developed to be asymptotically unbiased \citep{rectify}. Since its base model is recursive, the majority of dynamics are consistent over its forecast, and the variance of its direct component is reduced.

Although direct methods are more theoretically motivated, at least in the large data limit, it is not obvious which MSF strategy to use in practice. Multiple studies compare the performances of different MSF strategies, and their findings are not entirely consistent: \citet{atiya1999comparison} favour direct strategies, \citet{taieb2012reviewmultistep} favour multi-output strategies, \citet{An_comp, noa2024dirrecmo} favour dirrec, whereas \citet{Ji2017Recmo} favour recursive strategies.

Few studies have extensively compared their performance across multiple datasets. For example, \citet{An_comp} focuses solely on Multi-Layer Perceptrons (MLPs), and \citet{taieb2012reviewmultistep} analyses performance within a single domain. Although \citet{taieb2014machine} provides an extensive theoretical and empirical analysis, specifically on variance-bias decomposition, newer strategies such as RecMO and DirRecMO have since emerged. To our knowledge, no research has comprehensively explored the breadth of MSF strategies across multiple domains and horizon lengths.

Our review of the literature highlights three main gaps. Firstly, the Rectify strategy remains to be made into a multi-output strategy. Secondly, there is no unifying framework for understanding and exploring MSF strategies. Lastly, the most recent evaluations of MSF strategies \citep{An_comp, taieb2012reviewmultistep,noa2024dirrecmo} do not include the most recent additions, or multiple datasets, and is therefore out-of-date. \textit{Stratify} is motivated to fill these gaps by unifying all existing strategies, as well as introduce effective novel strategies, and provide an intuitive framework for representing the strategy space. Our experiments are the most comprehensive benchmarking of strategies over datasets and function classes considered to date.

\section*{Preliminaries} \label{prelim}

Multi-step forecasting involves predicting future values of a time series based on historical observations. Given a univariate time series $\{ y_1, y_2, \dots, y_T \}$ consisting of $T$ observations, the goal is to forecast the next $H$ observations $\{ y_{T+1}, y_{T+2}, \dots, y_{T+H} \}$, where $H$ is the forecasting horizon. This problem can be formulated as:

\begin{equation}
\label{slidingeq}
\{ y_{T+1}, y_{T+2}, \dots, y_{T+H} \} = \mathcal{D}( y_1, y_2, \dots, y_T ),
\end{equation}

where $\mathcal{D}$ represents the unknown function that maps past observations to future values, as discussed in \citet{vapnik_overview}.

\noindent \textbf{Notation} For the time series $\{ y_1, y_2, \dots, y_T \}$, the subscript $y_{2:5}$ refers to the sequence from $\{ y_2, \dots, y_5 \}$, and $y_{:j}$ refers be all points up to the index $j$. We define the function $\varsigma(H) = \{ \sigma \in \mathbb{Z}^+ \mid H \mod \sigma = 0 \}$, where $\varsigma(H)$ returns the set of all numbers that divide the value $H$ with no remainder. For a parameterisable strategy, $G$, we use $G$-$\sigma$ to refer to $G$ being parameterised with the value of $\sigma$.

\subsection{Single-Output Strategies}

The \textit{recursive strategy} iteratively applies a single-output model $f$ to predict one step ahead, using each new prediction as input for the next forecast. The model is defined as:

\begin{equation}
\label{recursive}
\hat{y}_{T+1} = f\left( y_t, y_{t-1}, \dots, y_{t-w+1} \right),
\end{equation}
where \( w \) is the window size (the number of past observations considered), and \( \hat{y}_{t+1} \) is the predicted value at time \( t+1 \).

Let $x = y_{T-1}, \dots, y_{T-w+1}$ be the $w$ most recent observations of the time series. To forecast $H$ steps ahead starting from time $T$, the predictions are obtained as:

\begin{equation}
\label{recursive_msf}
\hat{y}_{T+h} = 
\begin{cases} 
f\left( x \right), & \text{if } h = 1, \\[1ex]
f\left( \hat{y}_{T+h-1}, \hat{y}_{T+h-2}, \dots, x_{:w-h} \right), & \text{if } 1 < h \leq w, \\[1ex]
f\left( \hat{y}_{T+h-1}, \hat{y}_{T+h-2}, \dots, \hat{y}_{T+h-w} \right), & \text{if } h > w.
\end{cases}
\end{equation}

\noindent The \textit{direct strategy} employs $H$ distinct models, each predicting a specific horizon directly from the observed data:

\begin{equation}
\label{direct}
\hat{y}_{T+H} = \text{concat} (f_1(x), f_2(x), \dots, f_{H}(x)), \quad 
\text{where} \quad f_i : x \mapsto \hat{y}_{(i-1) : i},
\end{equation}

with \( i \in \mathcal{I} \) such that \( \mathcal{I} = \left\{ 1, 2, \dots, H \right\} \).

\noindent The \textit{DirRec strategy} (Direct-Recursive) combines elements of both recursive and direct methods by using $H$ models, each incorporating previous predictions as inputs. DirRec forecasts also concatenate outputs of a set of functions, $\{f_1, \dots, f_H\}$, except they are defined as follows:

\begin{equation}
f_i : 
\begin{cases}
x \mapsto \hat{y}_{(i-1) : i}, & \text{if } i = 1, \\[1ex]
\text{concat}\left(\hat{y}_{(i-2) : (i-1)}, x\right) \mapsto \hat{y}_{(i-1) : i}, & \text{if } i > 1.
\end{cases}
\end{equation}

\noindent with \( i \in \mathcal{I} \) such that \( \mathcal{I} = \left\{ 1, 2, \dots, H \right\} \). Since the input space of $f_{i+1}$ depends on $f_{i}$, forecasts cannot be produced in parallel.

\begin{figure}[t!]
    \centering
    \includegraphics[width=\linewidth]{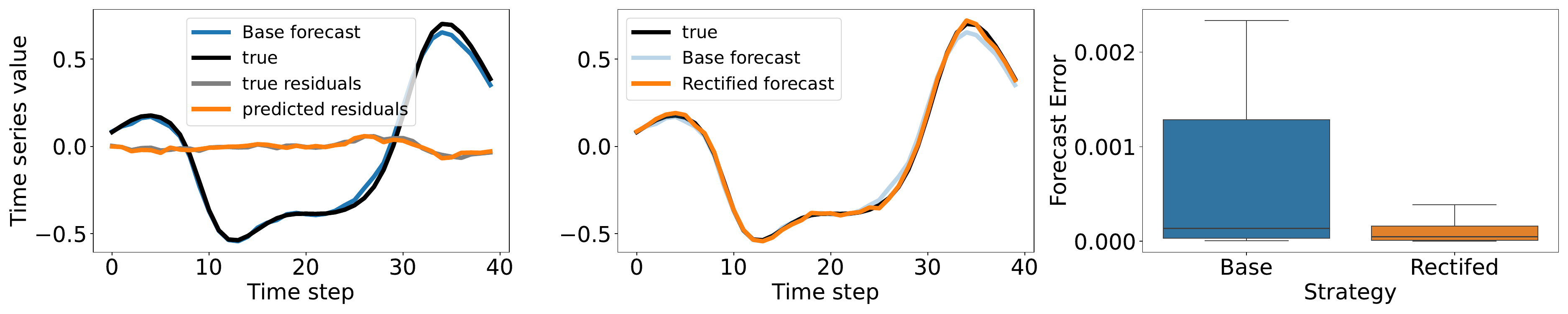}
    \caption{An example of Rectify, the recurisve forecast captures the majority of the dynamics and the variance in residuals is modelled effectively by the direct forecast.}
    \label{fig:rectifyexample}
\end{figure}

\noindent The \textit{rectify strategy} is a two-stage multi-step forecasting strategy that combines the strengths of both recursive and direct forecasting approaches by reducing bias while controlling variance. The rectify strategy involves two steps. First, for some target time series $\{ y_1, y_2, \dots, y_T \}$, and where $x$ is defined as the $w$ recent values of the time series, a single-output recursive strategy, denoted as $b$, is trained as in \autoref{recursive}. Forecasts from the base model,  $\beta_{T:T+H}$ are produced via \autoref{recursive_msf} and these are referred to as the base forecasts. Next, a new time series, $\{ \eta_1, \dots, \eta_T \}$ is generated where:
\begin{equation}
    \label{eta_gen}
\eta_i = y_i - \beta_i, \forall i \in [1,T]. 
\end{equation}

A direct strategy is then trained to forecast $\eta_{T:T+H}$ with $H$ distinct models, denoted as $\{r_1, \dots, r_{H}\}$, similar to \autoref{direct}:

\begin{equation}
\label{direct}
\hat{\eta}_{T+H} = \text{concat}\left(r_1(x), \dots, r_{H}(x)\right), \quad 
\text{where} \quad f_i : x \mapsto \hat{\eta}_{(i-1) : i},
\end{equation}
with \( i \in \mathcal{I} \) such that \( \mathcal{I} = \left\{ 1, 2, \dots, H \right\} \).

The final forecast for the rectify strategy, $y_{T:T+H}$ is produced via obtaining $\beta_{H:H+T}$ as from \autoref{recursive_msf} and $\eta_{T:T+H}$ from \autoref{direct} and their element-wise addition:
\begin{equation}
    \label{rectify}
    \hat{y}_{T:T+H} = \beta_{H:H+T}  + \eta_{H:H+T}
\end{equation}

\noindent The rectify strategy adjusts predictions from the base model by modelling its residuals, $\eta$. The base model is responsible for capturing the primary dynamics of the time series, while the rectifying model compensates for its high bias. By combining the two, the rectify strategy aims to produce accurate forecasts with reduced bias and variance. An example is shown in Figure \ref{fig:rectifyexample}.

\subsection{Multiple-Output Strategies}

\noindent The \textit{Multiple Input Multiple Output (MIMO)} method uses a single model $f$ to predict all $H$ future values simultaneously:

\[
\hat{y}_{T:T+H} = f( y_T, y_{T-1}, \dots, y_{T-w+1} )
\]

\noindent where $f: \mathbb{R}^w \rightarrow \mathbb{R}^H$.
\\
Recursive, direct, and DirRec strategies have been extended to multiple outputs, parameterised by $\sigma$, the number of steps predicted at each iteration for producing a forecast.

\begin{figure}[hbt!]
    \centering
    \includegraphics[width=\linewidth]{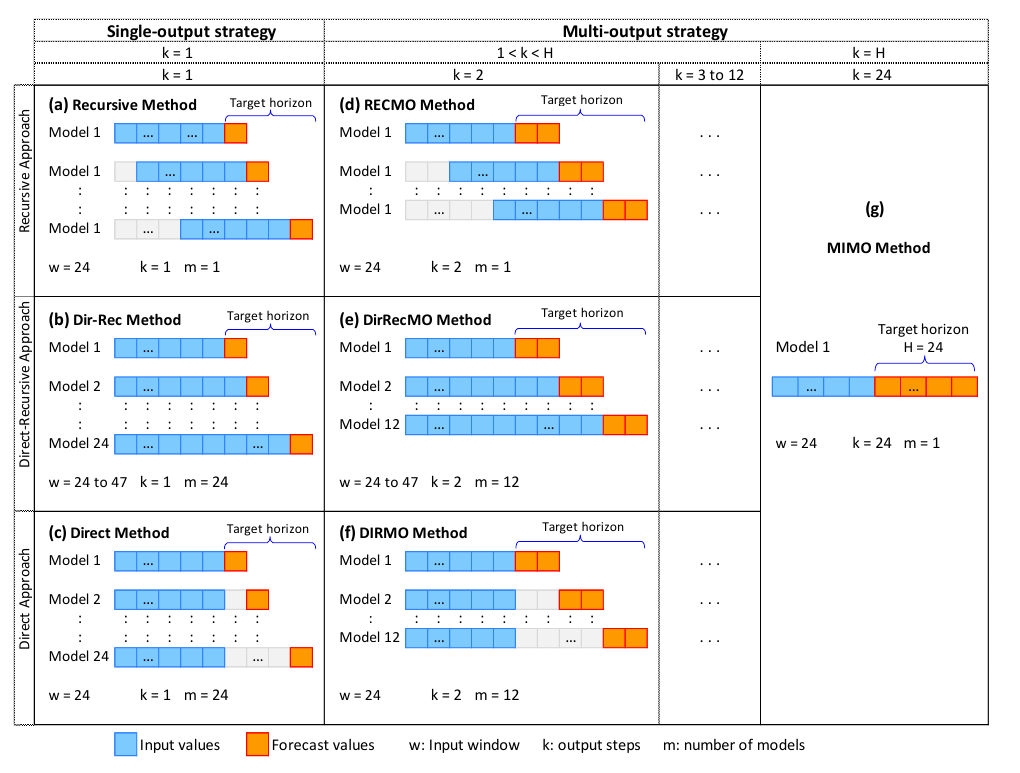}
    \caption{We show Figure 3 from \citet{noa2024dirrecmo}. Forecasts over $H = 24$ for each forecasting strategy are shown, where they are parameterised by $k$, instead of $\sigma$ as used in this work. Recursive/RecMO strategies consist of a single model that iterates forwards $k$ values using its own predictions as future inputs until the horizon is reached. Direct methods train an independent model for each $k$ values. DirRec methods also train independent models for each $k$ values but use the previous predictions as input into the subsequent prediction. All methods become equivalent when $k=H$.}
    \label{fig:dirrecmofigure}
\end{figure}

\noindent The \textit{RecMO} (Recursive Multiple-Output) strategy uses a single model $f$ to predict $\sigma$ steps ahead recursively:

\begin{equation}
\hat{y}_{T + (i-1)\sigma : T + i\sigma} = 
\begin{cases}
f(x), & \text{if } i = 1, \\[1ex]
f\left(\hat{y}_{T + (i-2)\sigma : T + (i-1)\sigma}, x_{w - (i-1)\sigma} \right), & \text{if } i > 1,
\end{cases}
\end{equation}
with \( i \in \mathcal{I} \) such that \( \mathcal{I} = \left\{ 1, 2, \dots, \frac{H}{\sigma} \right\} \). The total multi-step forecast is simply the concatenation of $\hat{y}_{T + (i-1)\sigma : T + i\sigma}$ for all $i \in \mathcal{I}$. Since $\hat{y}_{T + (i)\sigma : T + (i+1)\sigma}$ depends on $\hat{y}_{T + (i-1)\sigma : T + i\sigma}$, forecasts cannot be produced in parallel.

\noindent The \textit{DirMO} (Direct Multiple-Output) strategy trains $\frac{H}{\sigma}$ models $\{ f_1, f_2, \dots, f_{H/\sigma} \}$, each predicting $\sigma$ future values directly:

\begin{equation}
\hat{y}_{T:T+H} = \text{concat}\left(f_1(x), f_2(x), \dots, f_{H/\sigma}(x)\right), \quad 
\text{where} \quad f_i : x \mapsto \hat{y}_{(i-1)\sigma : i\sigma},
\end{equation}
with \( i \in \mathcal{I} \) such that \( \mathcal{I} = \left\{ 1, 2, \dots, \frac{H}{\sigma} \right\} \) 
and \( \sigma \mid H \). Since the models $f_i$ have identical input spaces, forecasts using the DirMO can still be produced in parralel.

\noindent The \textit{DirRecMO} (Direct-Recursive Multiple-Output) strategy combines the DirRec and MO approaches. Similar to DirMO, it uses $\frac{H}{\sigma}$ models $\{ f_1, f_2, \dots, f_{H/\sigma} \}$ with each predicting $\sigma$ steps ahead, but with inputs incorporating previous predictions. The DirRecMO strategy's $f_i$ are defined as follows:

\begin{equation}
f_i : 
\begin{cases}
x \mapsto \hat{y}_{(i-1)\sigma : i\sigma}, & \text{if } i = 1, \\[1ex]
\text{concat}\left(\hat{y}_{(i-2)\sigma : (i-1)\sigma}, x\right) \mapsto \hat{y}_{(i-1)\sigma : i\sigma}, & \text{if } i > 1.
\end{cases}
\end{equation}

with \( i \in \mathcal{I} \) such that \( \mathcal{I} = \left\{ 1, 2, \dots, \frac{H}{\sigma} \right\} \) 
and \( \sigma \mid H \). Since the input space of $f_{i+1}$ depends on $f_{i}$, forecasts cannot be produced in parallel.

\section*{Stratify: A Unified Framework for MSF Strategies}\label{method}
The \textit{Stratify} framework is constructed by first parameterising Rectify, completing the literature with multi-output variants of the well-known single-output strategies.Then we extend the multi-output rectify (RectifyMO) to Stratify, our main contribution.

\subsection{RectifyMO: Extending Rectify to Multi-Output}

The \textit{Rectify} strategy strictly produces base forecasts with the recursive strategy and a residual forecast with the direct strategy. Extending Rectify into a multi-output strategy, referred to as \textit{RectifyMO}, is relatively straightforward. For some $\sigma$, we simply substitute the base model, $b$ with a RectMO strategy with parameterisation equal to $\sigma$, as well as the direct residual forecasting strategy with a DirMO strategy with the same parameterisation, $\sigma$. For RectifyMO, the base and rectifying models are denoted as $b_\sigma$ and $r_\sigma$ respectively.
 
 The RectifyMO$_\sigma$ strategy involves the following two steps, for some $\sigma$: a $\sigma$-RecMO base strategy is trained and produces the $\beta_{H:H+T}$ forecast in $\sigma$-steps. As with Rectify, the residual time series $\eta$ is generated using \autoref{eta_gen} where $b_\sigma$ is used to generate the $\beta_i$. The $\sigma$-DirMO strategy then predicts $\eta_i$ from $x$. The resulting forecast for is the summation of $\hat{y}_{T:T+H} = \beta_{H:H+T}  + \eta_{H:H+T}$, identical to Rectify in \autoref{rectify}.

 The multi-output formulation, RectifyMO, allows further balancing of variance and bias via parameterisation the base and residual forecasters by $\sigma$. Similar to other MO-parameterisations, at $\sigma = 1$ we have the original single-output Rectify, and at $\sigma = H$ we have the MIMO strategy (as the base and rectifier).

\subsection{Generalising RectifyMO into Stratify}
The motivation for Rectify is to use a base forecaster to capture general dynamics to simplify the problem for its residual forecaster \citep{rectify}. With RectifyMO we followed the same motivation where the base dynamics are forecasted by a biased estimator in a RecMO strategy, for some $\sigma$, and an unbiased  estimator for the residual forecaster with the same $\sigma$. One way to generalise RectifyMO further is by allowing for the base and the residual forecasters to have different $\sigma$-parameters. However, by also by allowing for any combination of RecMO, DIRMO, or DirRec strategies as the base or rectifier, we have the most generalised framework for MSF strategies. We name our general framework \emph{Stratify} for its resemblance to the Rectify strategy.

For forecasting horizon $H$, and let the set $\varsigma(H) = \{ \sigma \in \mathbb{Z}^+ \mid H \mod \sigma = 0 \}$. We define a list of strategies $S_{H}$:
\begin{equation}
    \label{bigS}
S(H) = \bigcup_{\sigma \in \varsigma(H)} \{ \text{RecMO}(\sigma), \text{DIRMO}(\sigma), \text{DirRec}(\sigma) \}.
\end{equation}

The set of strategies $S$ represents every possible multi-output strategy for horizon $H$, with redundancy where RecMO-$\sigma$ $\equiv$ DirMO-$\sigma$ $\equiv$ DirRec-$\sigma$ at $\sigma = H$. The \textit{Stratify} framework follows a similar method to RectifyMO, where there is a base strategy and a rectifying one. However, instead of indexing strategies with $\sigma$, they are indexed using a three-dimensional vector $[\rho_\sigma, \delta_\sigma, \iota_\sigma]$ where $\rho_\sigma, \delta_\sigma, \iota_\sigma \in \varsigma(H)$. First, the base strategy is selected with index $S(H)_i$, which produces $\beta_{H:H+T}$. Secondly, the rectifying strategy is selected with index $S(H)_j$ which produces $\eta_{H:H+T}$. The final forecast is then produced under \autoref{rectify} with the respective base and rectifying strategies.
The \textit{Stratify} framework offers practitioners with an exhaustive list of existing multi-step forecasting strategies. Through this framework we explore novel strategies. 
\begin{table}[b!]
\caption{A summary of forecasting strategies: forecasts produced in series use predicted values as inputs. Biased forecasts are ones where they do not converge to zero error in the infinite data limit. Strategies are generally parameterised by the size of the output space of their models.}\label{tab1}%
\footnotesize
\begin{tabular}{@{}llll@{}}
\toprule
Strategy & Uses Predicted Values  & Biased & Parameters \\
\midrule
Recursive    & Yes   & Yes  & 0  \\
Direct    & No   & No  & 0  \\
Rectify    & Yes   & No  & 0 \\
DirRec  & Yes   & No  & 0 \\
RecMO \footnotemark[1]   & Yes  & Yes  & 1 \\
DIRMO   & No   & No  & 1 \\
DIRRECMO \footnotemark[1]   & Yes   & No  & 1 \\
\textbf{RectifyMO}    & Yes   & No  & 1 \\
\textbf{Stratify} & Flexible & Flexible & 2 \\

\botrule
\end{tabular}
\footnotetext[1]{For $s \neq H$}

\end{table}
\\
\noindent \textbf{Strategy Notation} Whilst strategies are typically referenced by their parameter value \citep{An_comp}, we represent strategies as a percentage of their total forecasting horizon length. We make the claim that, by not normalising the parameter by the task horizon, the current approach precludes fair comparison across strategy types. For example, a RecMO-5 strategy on a horizon of 10 would require two recursive steps to complete its forecast, but would require four steps if used on a horizon of 20. To avoid this inconsistency, we instead represent strategies as parameterised by the percentage of the task horizon. So RecMO-50\% would equate to RecMO-5 on a horizon of 10, and RecMO-10 for a horizon of 20. In this work, we index \textit{Stratify} strategies using $\rho, \delta, \iota$ followed by $:X$ where $X$ is the percentage of the horizon forecasted under the RecMO, DirMO, DirRec strategy, respectively. As an example, $\rho$:10\%$\delta$:10\% on a horizon of 10 would equate to a base strategy of RecMO-1 and a rectifier of DirMO-1 (the original rectify strategy).

\subsubsection{Theoretical Considerations}
We know from \citet{taieb2014machine} showing that recursive forecasts are biased. Hence, we note that using a RecMO strategy in the base and in the residual forecaster results in a biased strategy. 
However, using an unbiased strategy as the base makes a \textit{Stratify} strategy unbiased for any rectifier selected. This follows from the residuals of the base converging to zero in the infinite data limit, resulting in a trivial task for any rectifier.
Our full generalisation of Rectify allows for a more flexible framework to select between the variance and bias of MSF strategies at both the base and rectifier level.


\section{Results}\label{results}
Using a diverse set of time series forecasting benchmarks, this section proposes experimental contributions to the following research questions:

\begin{itemize}
  
    \item (R1) To what extent does exploring the Stratify space aid multi-step forecasting?

    \item (R2) How do all known strategies compare to each other?
    
    \item (R3) How can we practically represent the space of strategies?
\end{itemize}

\subsection{Experimental Setup}\label{expsetup}
\textbf{Datasets} To evaluate the \textit{Stratify} space, we conduct experiments using the BasicTS benchmark suite \citep{shao2024basicts}, a comprehensive platform designed for fair and reproducible comparisons in multivariate time series (MTS) forecasting. We collapse along feature dimensions for multivariate time series and show their characteristics in Table \ref{datasetstable}. We include a synthetic chaotic time series of length 10,000 from the Mackey Glass equations from \citet{chandra2021evaluation_multi_deeplearning}, denoted $mg\_10000$.

\begin{table}[hbt!]
\footnotesize
\caption{Datasets from BasicTS benchmarking for multistep forecasting.}
\begin{tabular}{@{}llllll}
\toprule
Dataset           &  Domain&Length    & Mean       & Variance  & Range     \\ \midrule
Traffic           &  Transport&1.754e+04 & 5.700e-02  & 1.000e-03 & 1.520e-01 \\
METR-LA           &  Transport&3.427e+04 & 5.372e+01  & 2.242e+02 & 6.614e+01 \\
Illness           &  Physiological&9.660e+02 & 9.644e+04  & 2.716e+09 & 2.409e+05 \\
mg\_10000         &  Synthetic&9.999e+03 & 0.000e+00  & 1.200e-01 & 1.509e+00 \\
ExchangeRate      &  Finance&7.588e+03 & 6.950e-01  & 6.000e-03 & 3.450e-01 \\
ETTm1             &  Energy&5.760e+04 & 4.794e+00  & 6.466e+00 & 1.788e+01 \\
ETTh2             &  Energy&1.440e+04 & 1.763e+01  & 2.161e+01 & 4.003e+01 \\
Pulse             &  Physiological&2.000e+04 & 3.200e-02  & 3.100e-02 & 1.000e+00 \\
PEMS04            &  Transport&1.699e+04 & 2.117e+02  & 1.129e+04 & 3.302e+02 \\
PEMS03            &  Transport&2.621e+04 & 1.793e+02  & 8.506e+03 & 3.148e+02 \\
PEMS-BAY          &  Transport&5.212e+04 & 6.262e+01  & 2.718e+01 & 4.690e+01 \\
BeijingAirQuality &  Environment&3.600e+04 & 5.391e+01  & 1.297e+03 & 4.074e+02 \\
Weather           &  Environment&5.270e+04 & 1.889e+02  & 3.129e+03 & 1.219e+03 \\
ETTh1             &  Energy&1.440e+04 & 4.780e+00  & 6.430e+00 & 1.742e+01 \\
ETTm2             &  Energy&5.760e+04 & 1.764e+01  & 2.164e+01 & 4.031e+01 \\
PEMS07            &  Transport&2.822e+04 & 3.085e+02  & 1.670e+04 & 4.448e+02 \\
Electricity       &  Energy&2.630e+04 & 2.539e+03  & 9.849e+05 & 5.572e+03 \\
PEMS08            &  Transport&1.786e+04 & 2.307e+02  & 8.452e+03 & 3.031e+02 \\ \bottomrule
\end{tabular}
\label{datasetstable}

\end{table}

\noindent \textbf{Task settings} We consider forecast horizon lengths of 10, 20, 40, and 80 to examine the adaptability of our method to varying temporal prediction requirements. We use 80\%, 10\%, and 10\% splits for train, validation, and test, respectively. We use a fixed window length across all 1080 experiments of $w = 160$. This is selected since the minimum number of lagged values is at least two times the forecast length for the recursive strategy to have at least one real value in its input, which matches our longest horizon of 80.


\noindent \textbf{Function classes} To ensure a comprehensive evaluation of the proposed \textit{Stratify} framework, five function classes are selected for comparison: Multilayer Perceptron (MLP), Recurrent Neural Network (RNN), Long Short-Term Memory (LSTM), Transformer, and Random Forest (RF). These function classes are selected to reflect a range of methodological paradigms. 

The Random Forest implementation is sourced from SKLearn \citep{scikit-learn}. For the deep learning-based models (MLP, RNN, LSTM, and Transformer), implementations are built using the Pytorch framework \citep{paszke2019pytorch}, known for its flexibility and performance in developing neural network architectures.

Each deep learning model is configured with two hidden layers, each containing 100 hidden units. We find this architecture balances computational efficiency with the capacity to model complex temporal relationships. The models are trained for 1000 epochs, a setting chosen based on exploratory experiments to ensure convergence across datasets. Finally, we employed the Adam optimiser \citep{kingma2014adam} with a learning rate of $0.01$ and a batch size of 1024. The configurations are consistent across all datasets and forecast horizons to ensure a fair comparison.

\noindent \textbf{Computation of the \textit{Stratify} plane across function families} In total we use 216 tasks per function class (18 datasets, 4 horizons, and 3 seeds). We show the computational training time for each strategy in the \textit{Stratify} space in Figure \ref{fig:compute_plane}. All strategies besides RecMO require fitting models for each position in the horizon; we only evaluate the entire \textit{Stratify} space for the MLP. For the remaining functions we only evaluate the \textit{Stratify} plane in the RecMO-RecMO region. We justify this with the following example: for horizon 80, we would need to train over 86,400 transformers just for direct/dirrec methods (20 strategies in Stratify for horizon 80 on 18 datasets with 3 seeds). This does not include training required for the remaining multi-output strategies. One of our core findings highlights that many novel strategies in \textit{Stratify} are better performing whilst also being much less computationally expensive to train.

\noindent \textbf{Evaluation and significance testing} To ensure statistical reliability, each experiment is repeated three times using different random seeds to account for variability in model initialisation and training. Model performance is quantified using the Mean Squared Error (MSE), a widely adopted metric that offers a consistent and interpretable measure of forecasting accuracy in time series tasks \citep{theoryandpractice}. Whilst the use of MSE penalises large errors more than other methods such as (Symmetric) Mean Absolute Percentage Error (S)MAPE, we use MSE as our metric for consistency with the training objective and since (S)MAPE methods are non-symmetric and can be unstable for small time series values.

To evaluate the significance of performance differences across models, we use the Friedman test as a non-parametric method to compare ranks across multiple datasets \citep{nemenyitest}. This test assesses whether there are statistically significant differences among the models' performances, under the null hypothesis that all models perform equally. If the null hypothesis is rejected, we conduct post-hoc pairwise comparisons using the Nemenyi test \citep{nemenyitest}. The Nemenyi test identifies which pairs of models differ significantly and is appropriate for handling multiple comparisons without inflating Type I error rates.

To visualise the results, we use Critical Difference (CD) diagrams, which rank models on a common axis \citep{nemenyitest}. In these diagrams, models connected by a horizontal bar fall within the critical difference threshold, indicating that their performance differences are not statistically significant at the chosen significance level, which is set to $0.05$ in this paper.

\subsection{Novel Strategies in \textit{Stratify} Improve Performance}
\noindent Table~\ref{rel_mse_table} shows the performance of novel strategies in \textit{Stratify} compared to existing strategies (still accessible in \textit{Stratify}) across various datasets and function families. Novel strategies consistently outperform the existing strategies in \textit{Stratify}. This is particularly evident for RNN and LSTM models, achieving mean reductions in error of 28\% and 25\% respectively. The \textit{Stratify} space is consistent across diverse function families. While the optimal strategy remains task-dependent, these results highlight the potential of the \textit{Stratify} space to introduce novel strategies that outperform traditional ones in multi-step forecasting, all of which are unified in \textit{Stratify}.

\begin{table}[!ht]
\caption{For each dataset and function family, we take the lowest MSE of a novel \textit{Stratify} strategy and divide it by the lowest MSE of existing strategies. Values less than 1 show where strategies in \textit{Stratify} outperform the best known previous methods. We show the standard error with $\pm$ calculated over three seeds and all horizons.}
\footnotesize
    \centering
\begin{tabular}{@{}llllll@{}}
\toprule
Dataset           & RF   & MLP  & RNN  & LSTM & Transformer \\ \midrule

        Traffic & 0.98 $\pm$ 0.01 & 0.95 $\pm$ 0.05 & 0.80 $\pm$ 0.18 & 0.93 $\pm$ 0.05 & 0.86 $\pm$ 0.07 \\ 
        METR-LA & 1.00 $\pm$ 0.01 & 0.99 $\pm$ 0.01 & 0.99 $\pm$ 0.02 & 0.95 $\pm$ 0.04 & 0.90 $\pm$ 0.10 \\ 
        Illness & 0.65 $\pm$ 0.12 & 0.64 $\pm$ 0.27 & 1.00 $\pm$ 0.00 & 1.00 $\pm$ 0.00 & 1.00 $\pm$ 0.00 \\ 
        mg\_10000 & 0.74 $\pm$ 0.03 & 0.59 $\pm$ 0.16 & 0.70 $\pm$ 0.27 & 0.22 $\pm$ 0.12 & 0.98 $\pm$ 0.01 \\ 
        ExchangeRate & 1.02 $\pm$ 0.01 & 0.85 $\pm$ 0.11 & 0.59 $\pm$ 1.35 & 0.97 $\pm$ 0.04 & 0.22 $\pm$ 0.27 \\ 
        ETTm1 & 0.97 $\pm$ 0.01 & 0.98 $\pm$ 0.01 & 0.88 $\pm$ 0.06 & 0.97 $\pm$ 0.04 & 0.96 $\pm$ 0.04 \\ 
        ETTh2 & 1.01 $\pm$ 0.03 & 0.99 $\pm$ 0.01 & 0.53 $\pm$ 0.26 & 0.87 $\pm$ 0.22 & 0.34 $\pm$ 0.20 \\ 
        Pulse & 1.00 $\pm$ 0.00 & 0.15 $\pm$ 0.24 & 0.98 $\pm$ 0.11 & 0.36 $\pm$ 0.65 & 1.00 $\pm$ 0.00 \\ 
        PEMS04 & 0.96 $\pm$ 0.01 & 0.87 $\pm$ 0.07 & 0.73 $\pm$ 0.14 & 0.60 $\pm$ 0.06 & 0.91 $\pm$ 0.11 \\ 
        PEMS03 & 0.96 $\pm$ 0.01 & 0.92 $\pm$ 0.03 & 0.87 $\pm$ 0.16 & 0.62 $\pm$ 0.09 & 0.91 $\pm$ 0.07 \\ 
        PEMS-BAY & 0.99 $\pm$ 0.01 & 0.91 $\pm$ 0.09 & 0.53 $\pm$ 0.12 & 0.48 $\pm$ 0.20 & 0.97 $\pm$ 0.03 \\ 
        BeijingAirQuality & 1.00 $\pm$ 0.00 & 0.98 $\pm$ 0.01 & 0.98 $\pm$ 0.01 & 0.99 $\pm$ 0.01 & 0.97 $\pm$ 0.02 \\ 
        Weather & 1.00 $\pm$ 0.00 & 0.96 $\pm$ 0.02 & 0.52 $\pm$ 0.31 & 0.83 $\pm$ 0.15 & 1.05 $\pm$ 0.05 \\ 
        ETTh1 & 0.95 $\pm$ 0.01 & 0.99 $\pm$ 0.01 & 0.80 $\pm$ 0.17 & 0.92 $\pm$ 0.07 & 0.97 $\pm$ 0.02 \\ 
        ETTm2 & 1.05 $\pm$ 0.04 & 0.99 $\pm$ 0.02 & 0.84 $\pm$ 0.12 & 0.99 $\pm$ 0.07 & 0.48 $\pm$ 0.28 \\ 
        PEMS07 & 0.92 $\pm$ 0.02 & 0.92 $\pm$ 0.07 & 0.38 $\pm$ 0.22 & 0.64 $\pm$ 0.14 & 0.85 $\pm$ 0.14 \\ 
        Electricity & 0.93 $\pm$ 0.02 & 0.96 $\pm$ 0.03 & 0.66 $\pm$ 0.00 & 0.62 $\pm$ 0.01 & 1.00 $\pm$ 0.01 \\ 
        PEMS08 & 0.90 $\pm$ 0.01 & 0.91 $\pm$ 0.06 & 0.26 $\pm$ 0.22 & 0.54 $\pm$ 0.08 & 0.96 $\pm$ 0.03 \\ 
\midrule
Mean           & 0.95 $\pm$ 0.10 &  0.86 $\pm$ 0.21 &   0.72 $\pm$ 0.21  &   0.75 $\pm$ 0.24 & 0.85 $\pm$ 0.23 \\
\bottomrule
\end{tabular}
\label{rel_mse_table}
\end{table}

\begin{figure}[ht]
    \centering
    \begin{subfigure}{0.45\textwidth}
         \centering
         \tiny
            \caption{Proportion of experiments where novel \textit{Stratify} strategies outperform the best of previously existing strategies, calculated over three seeds and all horizons.}
            \begin{tabular}{@{}llllll@{}}
            \toprule
            Dataset & RF & MLP & RNN & LSTM & Transformer \\ \midrule
            Traffic & 0.83 & 0.75 & 1.0 & 1.0 & 1.0 \\  
            METR-LA & 0.1 & 1.0 & 0.75 & 0.83 & 1.0 \\  
            Illness & 1.0 & 1.0 & 1.0 & 1.0 & 1.0 \\  
            mg\_10000 & 1.0 & 1.0 & 1.0 & 1.0 & 0.92 \\  
            ExchangeRate & 0.08 & 0.92 & 0.92 & 0.75 & 1.0 \\  
            ETTm1 & 1.0 & 1.0 & 1.0 & 0.83 & 0.92 \\  
            ETTh2 & 0.42 & 0.75 & 1.0 & 0.75 & 1.0 \\  
            Pulse & 0.0 & 1.0 & 1.0 & 1.0 & 0.92 \\  
            PEMS04 & 1.0 & 1.0 & 1.0 & 1.0 & 0.92 \\  
            PEMS03 & 1.0 & 1.0 & 1.0 & 1.0 & 1.0 \\  
            PEMS-BAY & 0.82 & 0.8 & 1.0 & 1.0 & 1.0 \\  
            BeijingAirQuality & 0.6 & 1.0 & 0.92 & 1.0 & 1.0 \\  
            Weather & 0.6 & 1.0 & 0.92 & 0.83 & 0.2 \\  
            ETTh1 & 1.0 & 0.83 & 1.0 & 0.83 & 0.92 \\  
            ETTm2 & 0.08 & 0.6 & 1.0 & 0.58 & 0.92 \\  
            PEMS07 & 1.0 & 0.92 & 1.0 & 1.0 & 0.92 \\  
            Electricity & 1.0 & 1.0 & 1.0 & 1.0 & 0.58 \\  
            PEMS08 & 1.0 & 1.0 & 1.0 & 1.0 & 1.0 \\  
            \midrule
            Mean & 0.70 & 0.92 & 0.97 & 0.91 & 0.90 \\
            \bottomrule
            \end{tabular}
            \label{percentage_table}
    \end{subfigure}
    \hfill
    \begin{subfigure}{0.42\textwidth}
        \centering
            \includegraphics[width=\linewidth]{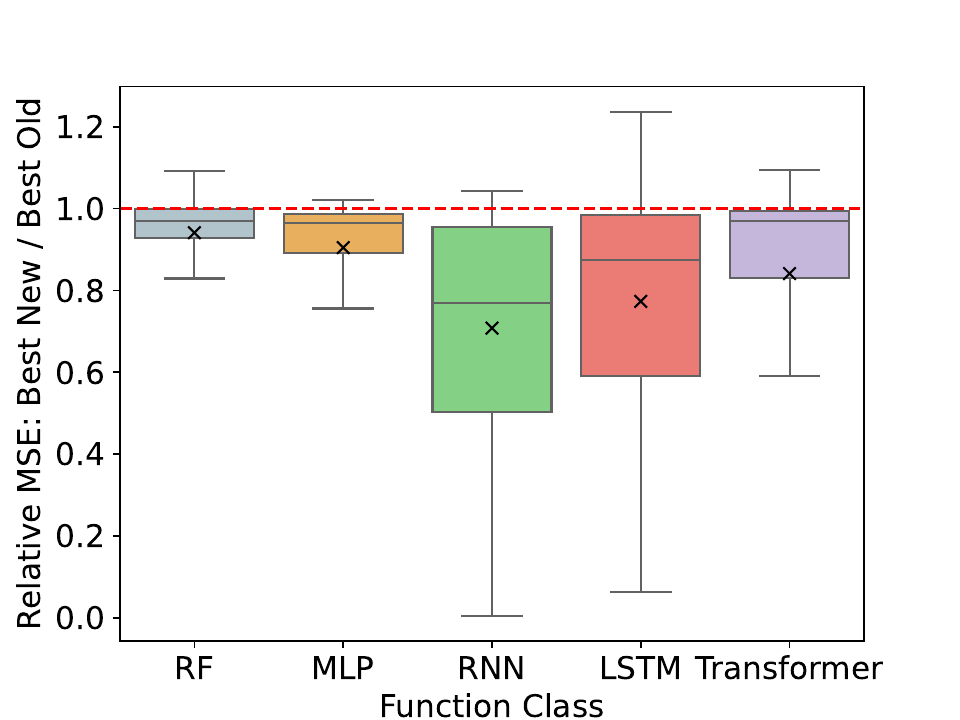}
    \caption{Aggregate results over function families. The lowest MSE of a novel \textit{Stratify} strategy is divided by the lowest MSE of existing strategies. Values less than 1 (shown by dashed red line) highlight task settings where \textit{Stratify} outperforms the best known previous methods. Boxplots (mean shown by `x') show the distribution over every dataset, three seeds, and all horizons.}
            \label{fig:error_ratios_box}
    \end{subfigure}
    \caption{}
\end{figure}

Table~\ref{percentage_table} shows the proportion of experiments in which the best strategy identified within the proposed \textit{Stratify} framework outperformed the best existing strategies across 18 benchmark datasets and five model types. The results demonstrate that \textit{Stratify} consistently contains improved forecasting strategies, with most proportions close to or equal to 1.0, indicating its high likelihood of containing effective methods. On average, \textit{Stratify} outperforms prior methods in 68\% to 87\% of experiments across different models, with RNNs achieving the highest mean proportion (97\%) and RFs the lowest (70\%). Since \textit{Stratify} unifies with existing strategies, the best performing strategy can still be accessed within the framework.

Figure~\ref{fig:error_ratios_box} is a collapsed representation of Table \ref{rel_mse_table} along the dataset and seeds dimension. The box plot quantifies the distribution of improvements by comparing the relative MSE of the best-performing strategies in \textit{Stratify} to the best existing strategies across function families. We report consistent reductions in error, with a relative MSE below 1.0 across all function families. For the RF, MLP and Transformer models, the median reductions are approximately 5\%, while the RNN and LSTM models exhibit more significant improvements $15-20\%$.


\subsubsection{Significant Improvements}

\begin{figure}[hbt!]
    \centering
    \includegraphics[width=\linewidth]{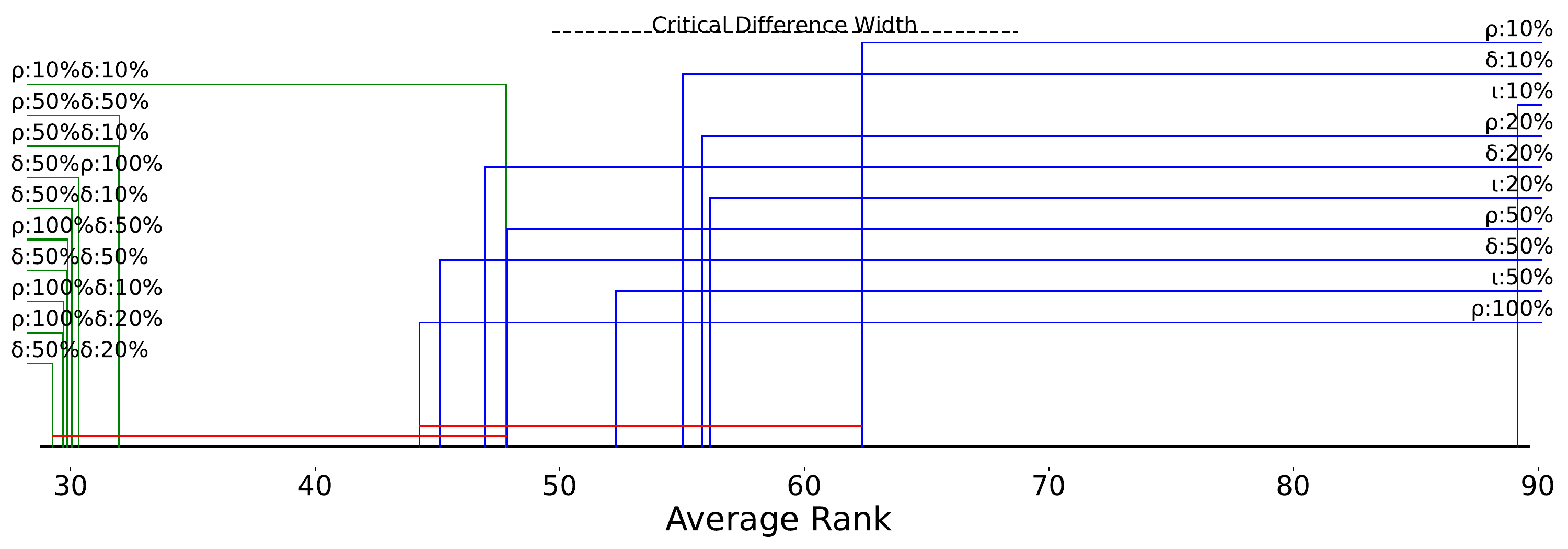}
    \caption{Critical difference diagram for MLP showing the ranking error (lower is better). The top 10 strategies in \textit{Stratify} (green) and all existing strategies (blue). The full diagram is shown in Figure \ref{fig:fullmlp_cd}. Cliques at the 95\% confidence are shown in red.}
    \label{fig:critical_differencemlp}
\end{figure}

Figure~\ref{fig:critical_differencemlp} presents the critical difference diagram that compares the average ranks of the top 10 strategies from the proposed \textit{Stratify} framework (green) against previously known methods (blue) for the MLP model. 
With 95\% confidence, we find that the ten novel strategies shown outperform 60\% of existing strategies and that none of the existing strategies are significantly outperforming other existing strategies.

\begin{figure}[t!]
    \centering
    \includegraphics[width=\linewidth]{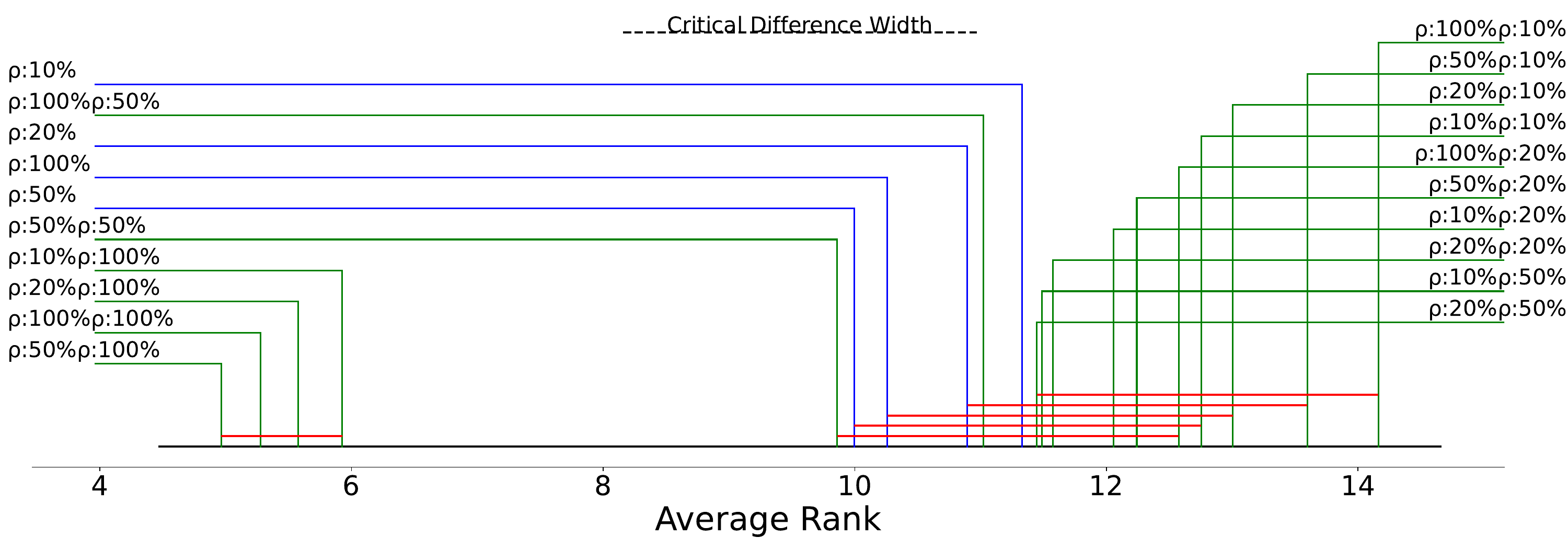}
    \caption{Critical difference diagram for RNN showing the ranking error (lower is better). \textit{Stratify} (green) and all existing strategies (blue). Cliques at the 95\% confidence are shown in red.}
    \label{fig:critical_differencernn}
\end{figure}

Figure \ref{fig:critical_differencernn} presents the critical difference diagrams for the top strategies in \textit{Stratify} using the RNN. The RF, LSTM, and Transformer are shown in the Appendix (Figures \ref{fig:critical_differencerf},  \ref{fig:critical_differencelstm}, and \ref{fig:critical_differencetrans}). For the Transformer and RF, the results are consistent with the MLP findings, where the best strategies in \textit{Stratify} are statistically significantly better than 75\% and 50\% of existing methods. In contrast, for the RNN and LSTM models, the results are more pronounced. The top four strategies, all employing an R:100\% rectifier, achieve statistically significant improvements over all the previous best methods.

\subsection{Representing Forecasting Strategies}

\begin{figure}[t!]
\centering
\begin{tabular}{ccc}
\includegraphics[width=0.31\textwidth]{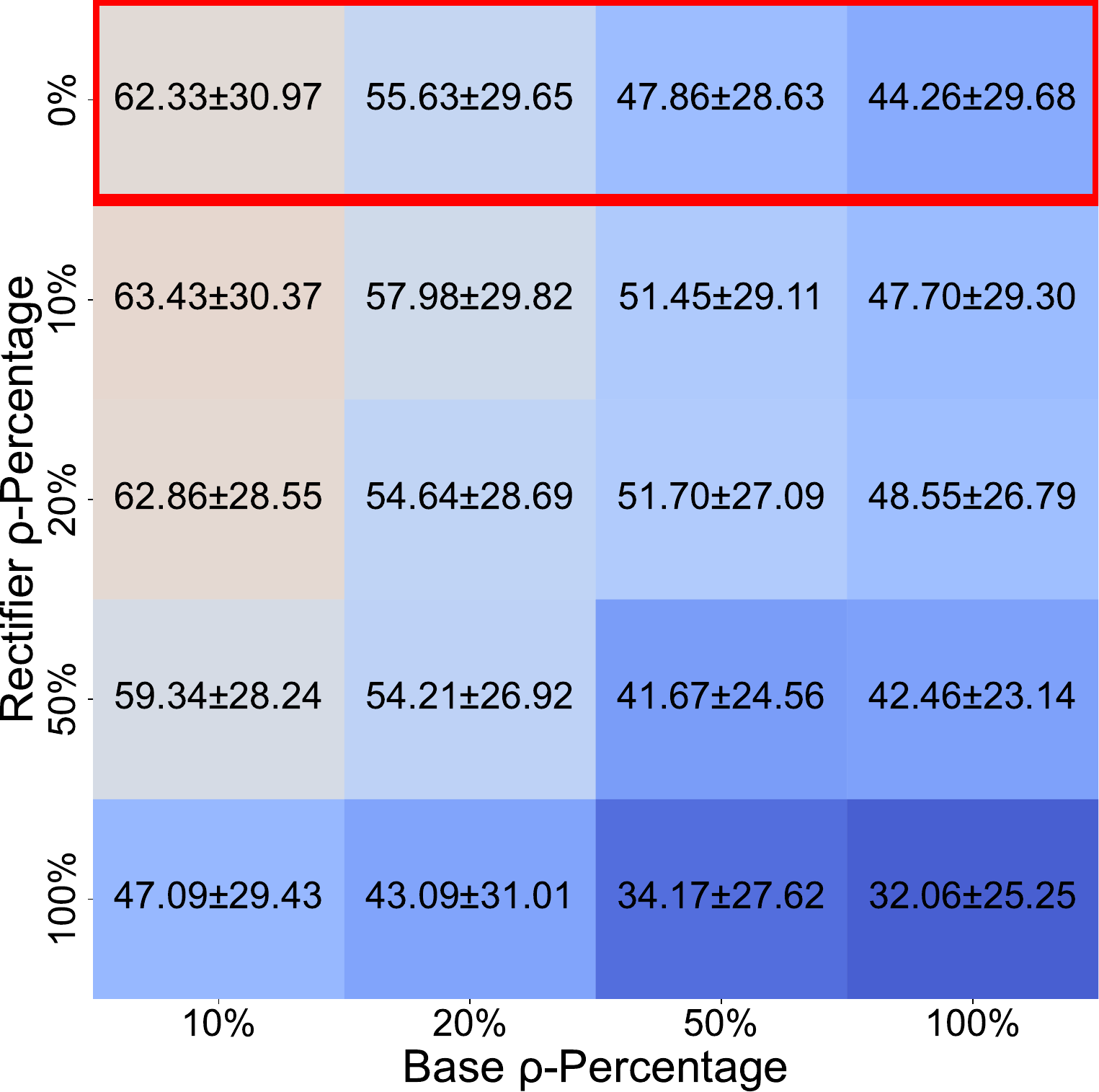} &
\includegraphics[width=0.31\textwidth]{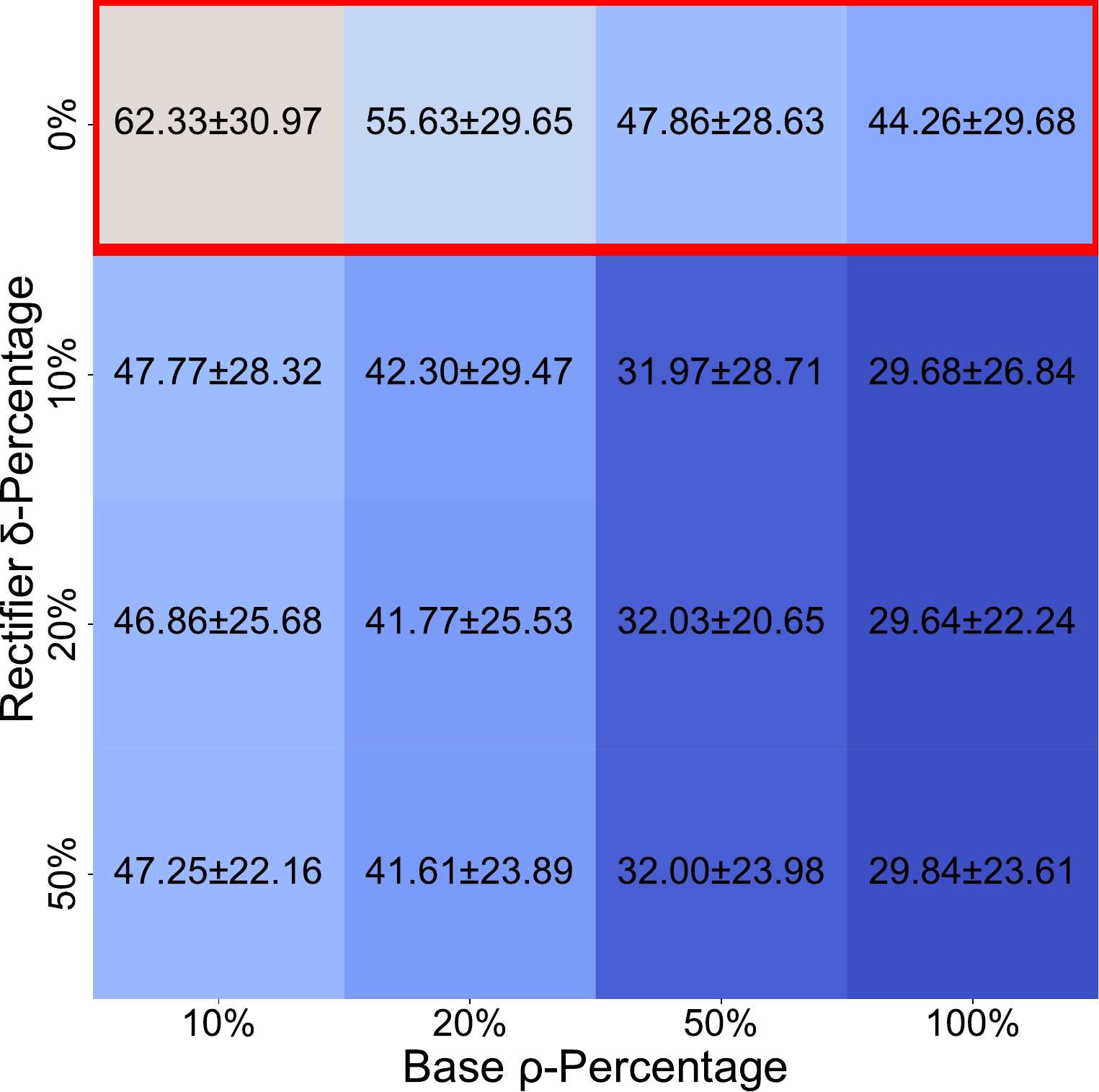} &
\includegraphics[width=0.31\textwidth]{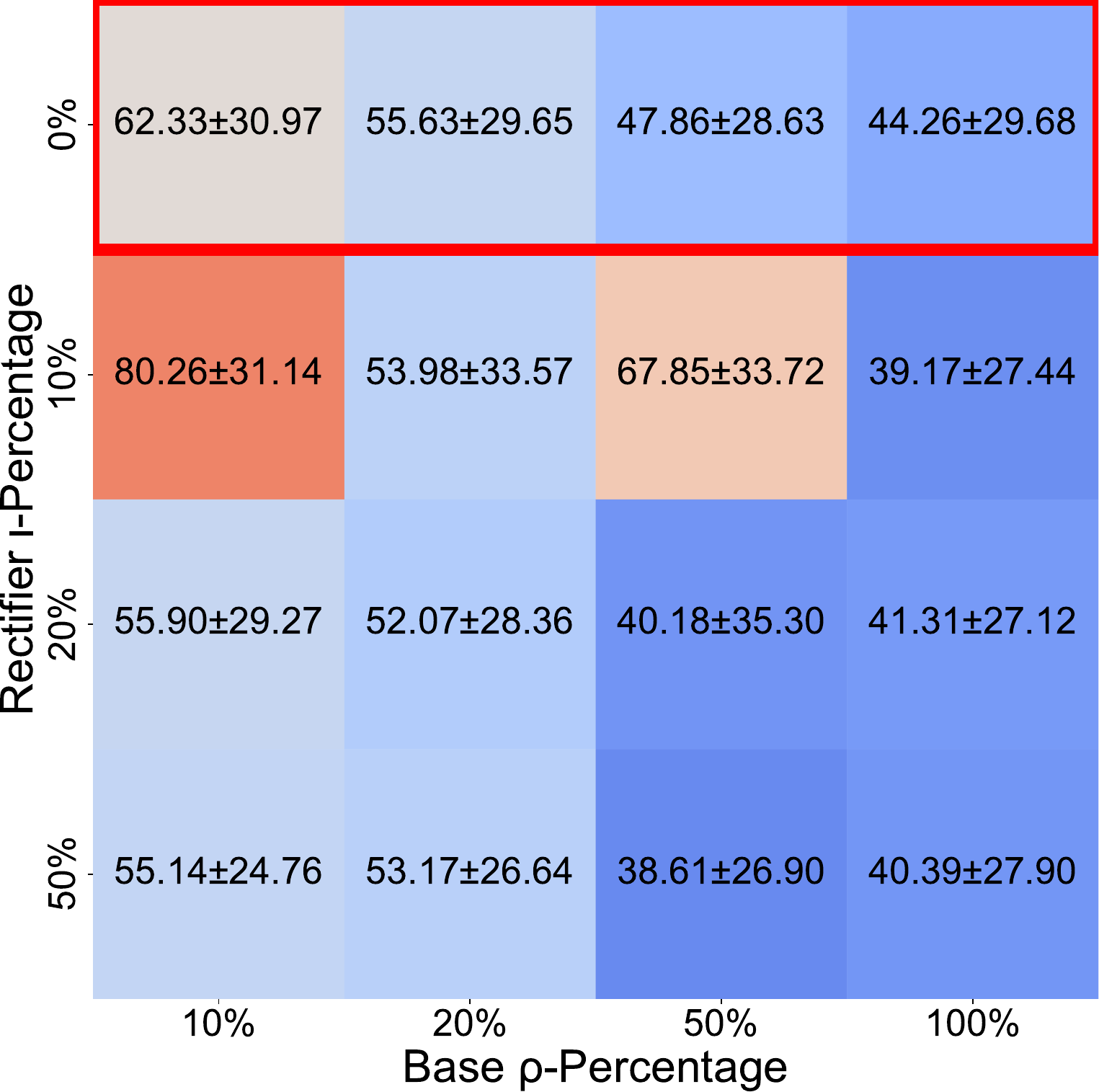} \\
\includegraphics[width=0.31\textwidth]{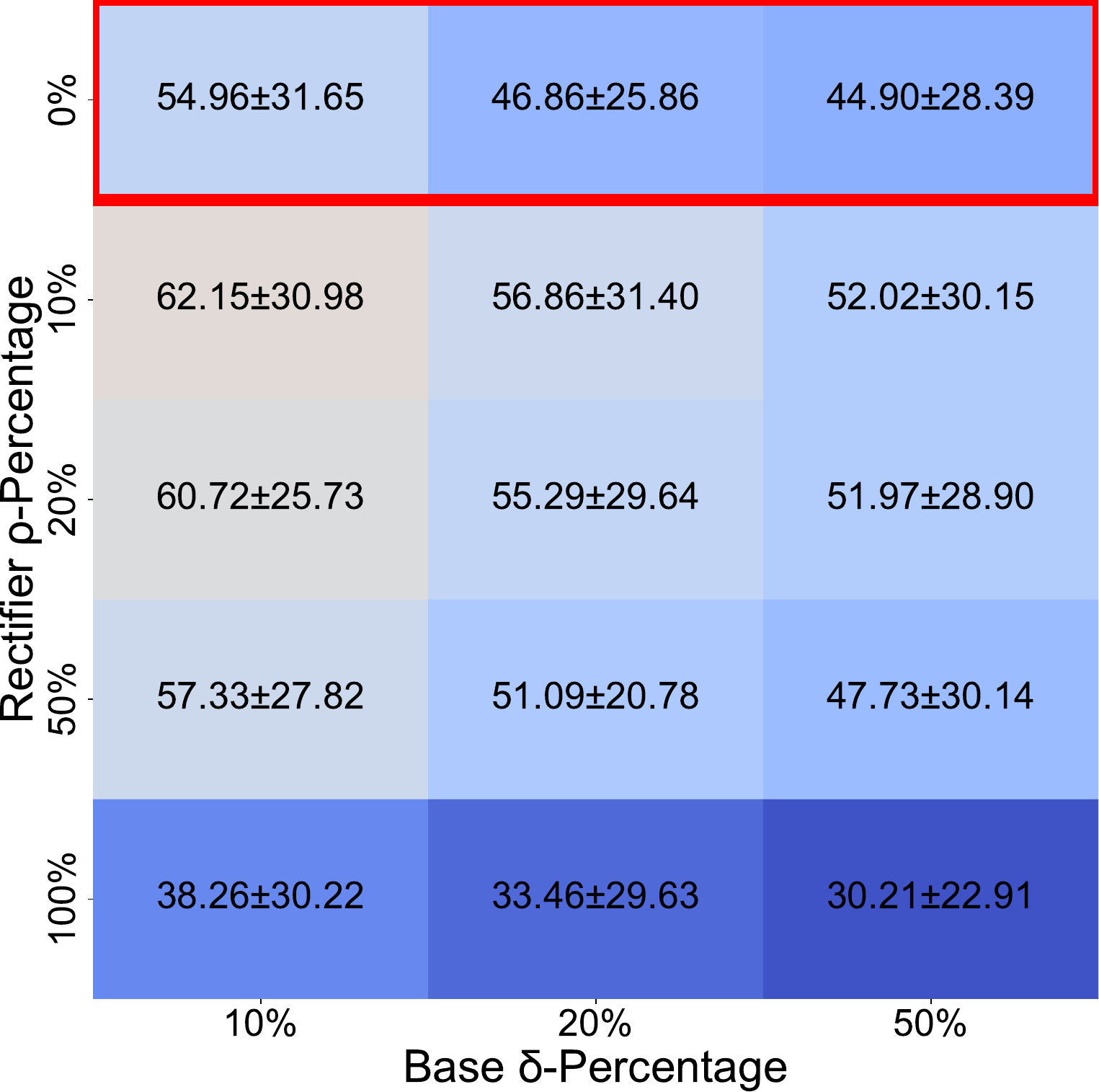} &
\includegraphics[width=0.31\textwidth]{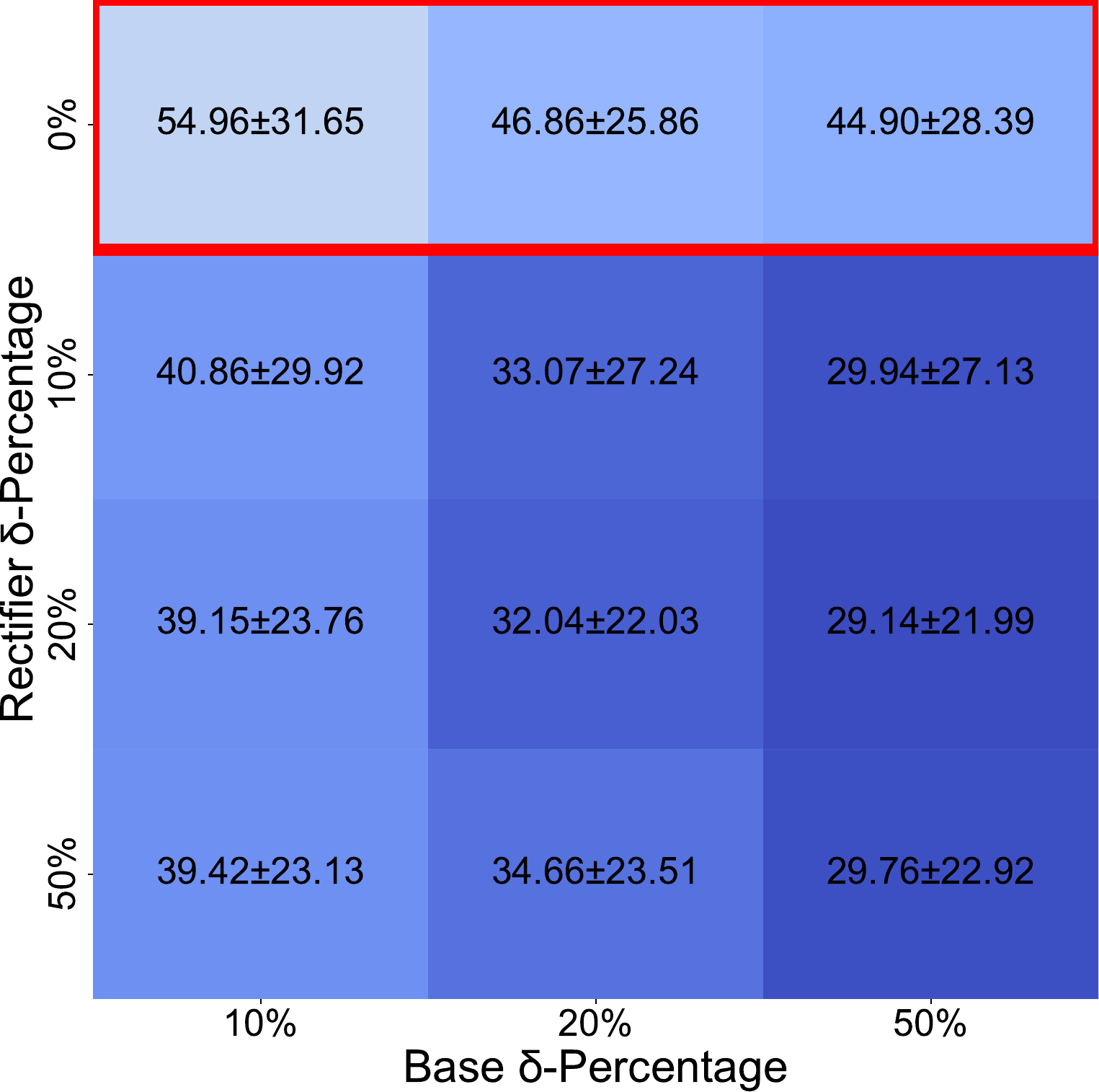} &
\includegraphics[width=0.31\textwidth]{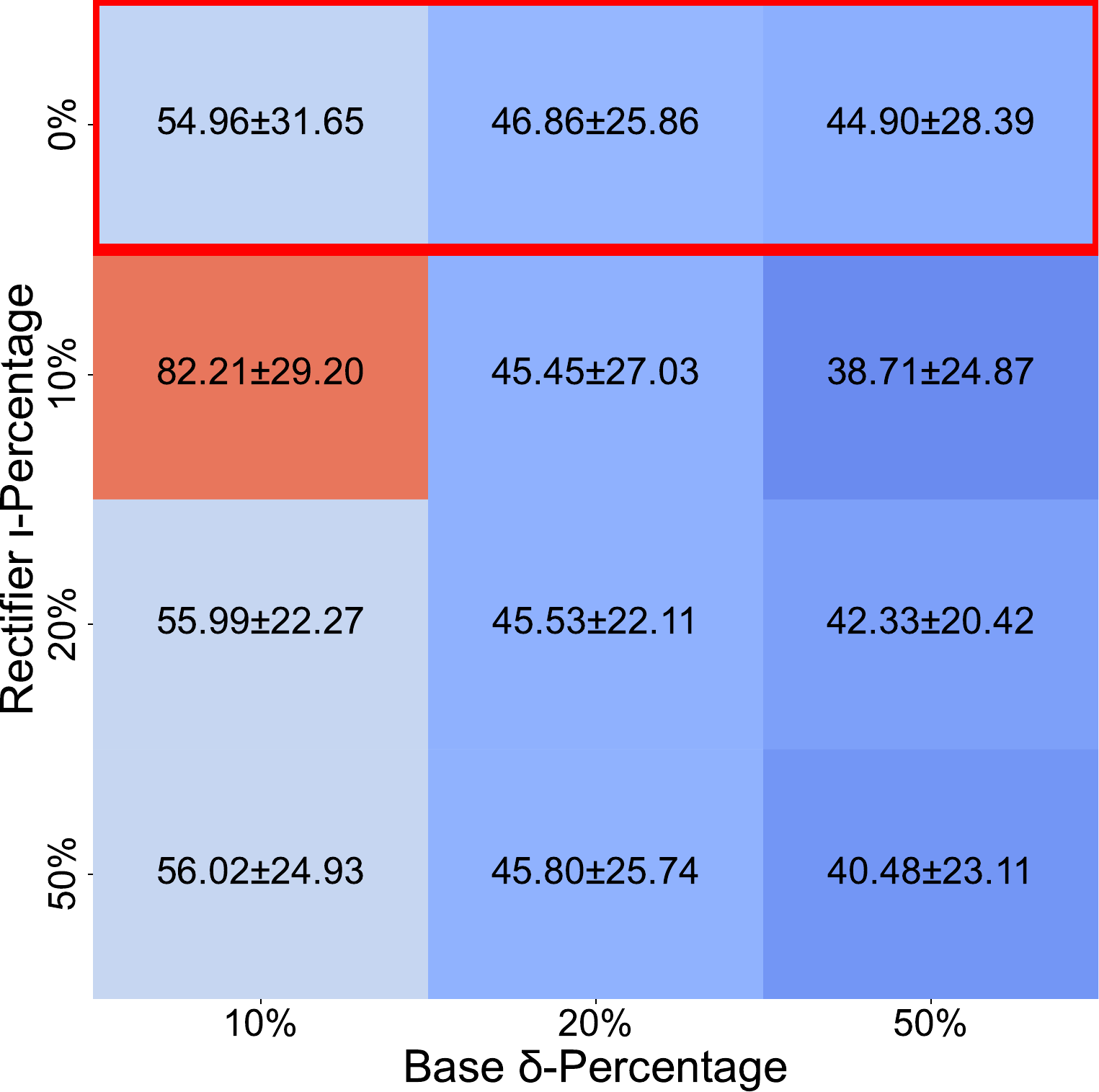} \\
\includegraphics[width=0.31\textwidth]{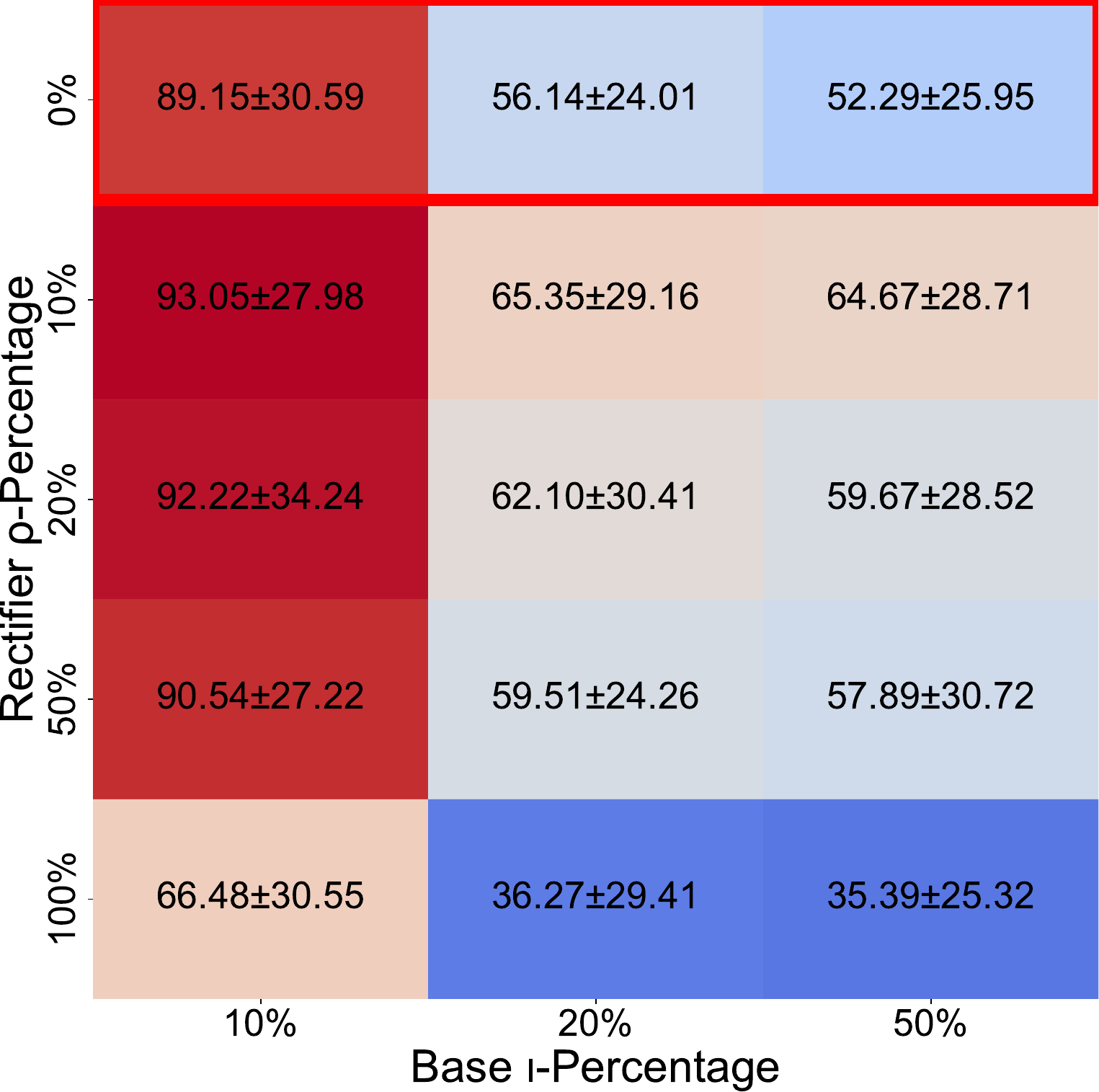} &
\includegraphics[width=0.31\textwidth]{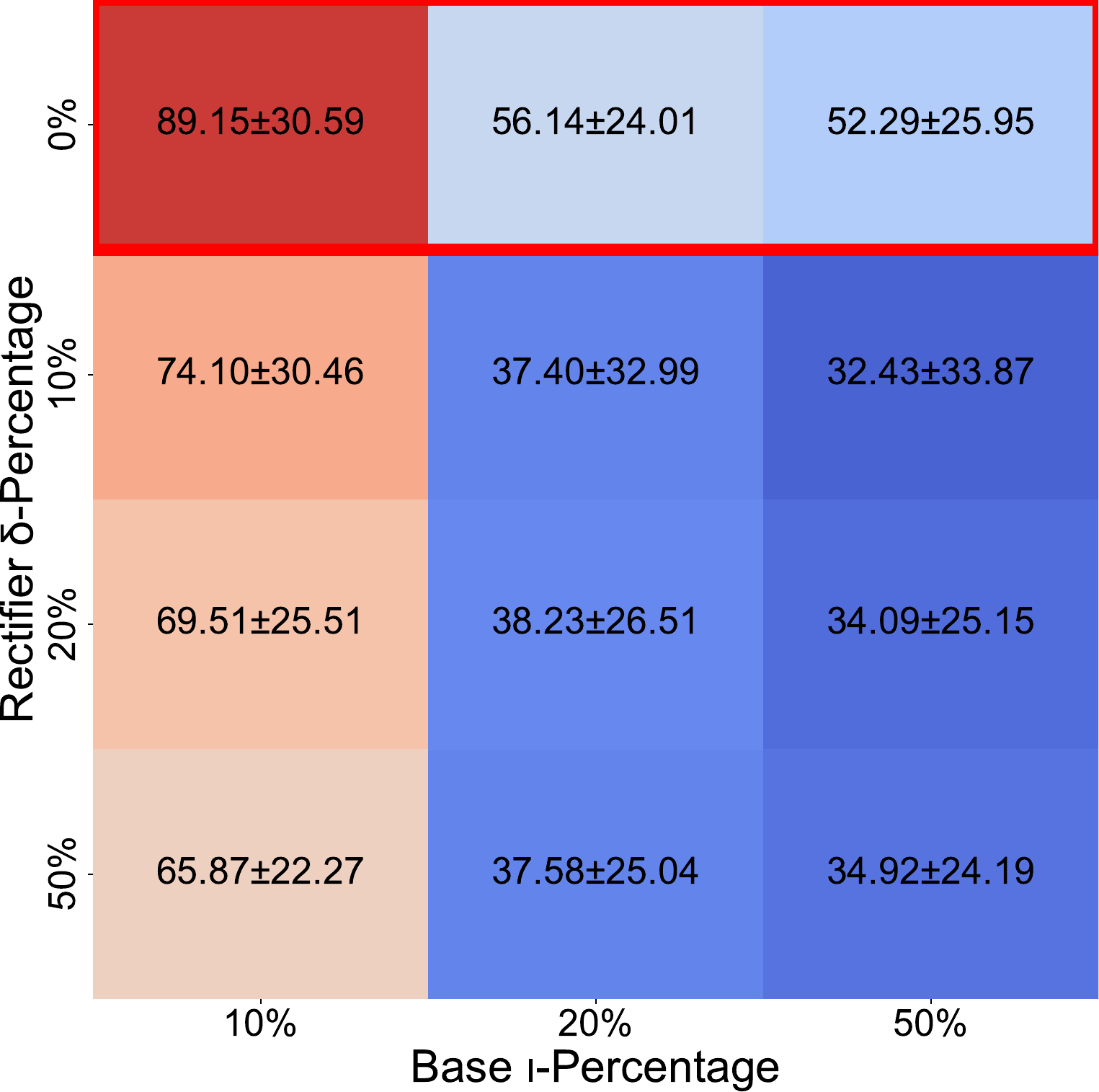} &
\includegraphics[width=0.31\textwidth]{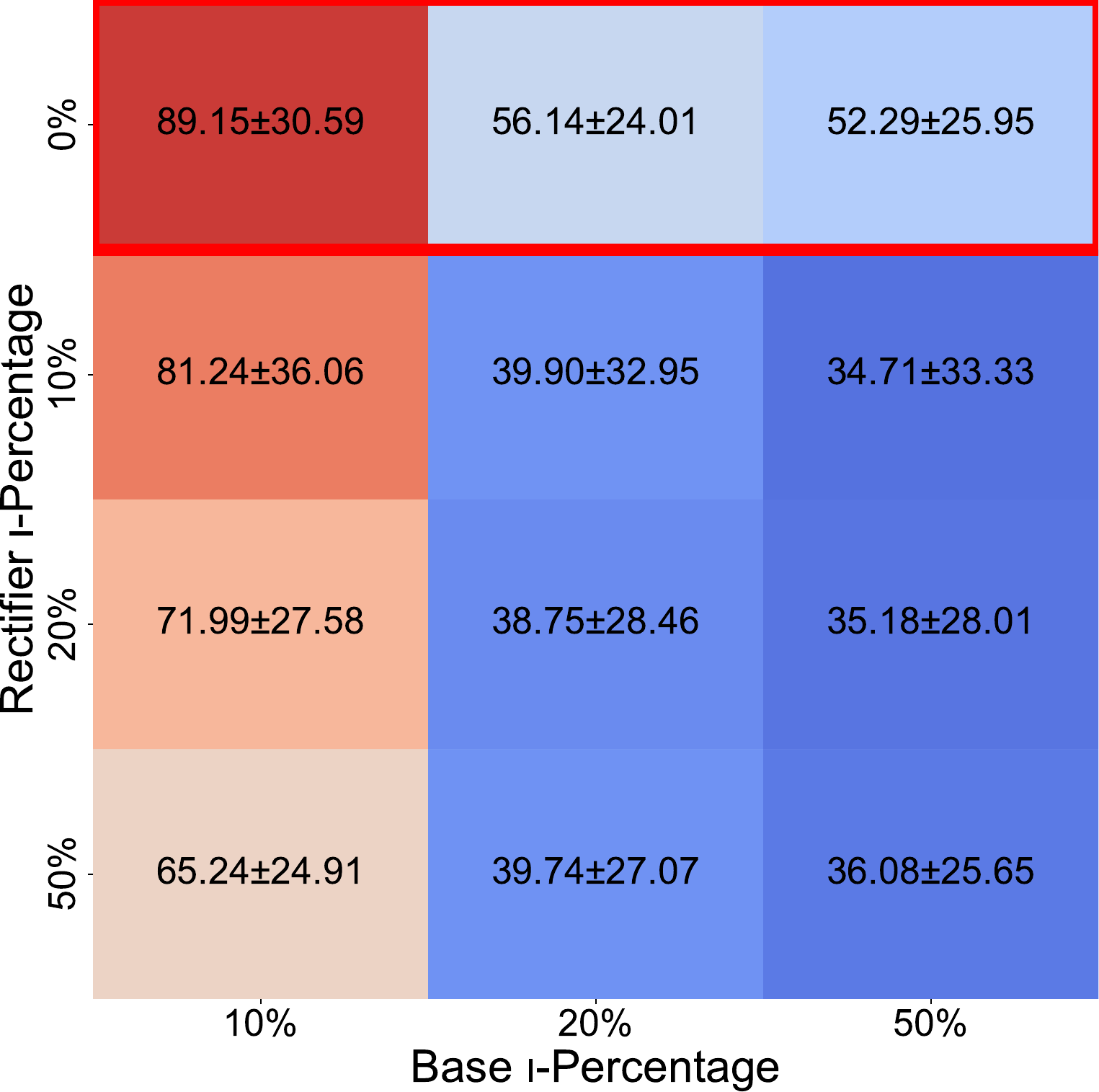} \\
\end{tabular}
\caption{All regions of the \textit{Stratify} plane for MLP forecaster on all datasets normalised over horizon length. Cells surrounded by red indicate previously existing strategies, represented by Stratify. Red signifies worse ranking error and blue shows a better ranking error. Exact mean ranks and the standard errors are shown in $\pm$. Values are calculated over  18 datasets, all horizons, and 3 seeds.}
\label{fig:stratify_plane}
\end{figure}

Figure \ref{fig:stratify_plane} illustrates the performance of all strategies within the \textit{Stratify} framework for MLP models, categorised by recursive ($\rho$), direct ($\delta$), and dirrec ($\iota$) forecasting strategies, normalised across datasets and forecast horizon lengths. The results reveal several key trends. 
The best-performing strategies are towards the bottom right of each plane. This supports the intuition of Rectify, where a direct strategy is used as the rectifier. In contrast, the upper regions of each plane show worse performance, which is where existing methods and rectifiers with small parameter values are.
The three best-performing strategies are novel, all have a $\delta:50$ base and $\delta:10$, $\delta:20$, $\delta:50$ rectifiers. However, all strategies exhibit relatively high variances, highlighting the task-specific nature of optimal configurations. While general trends are evident, the best strategy remains highly dependent on the dataset and task. This highlights the importance of a representation of the space of strategies for practitioners to select from.

\begin{figure}[t!]
\centering
\begin{tabular}{ccc}
\begin{subfigure}{0.48\textwidth}
    \centering
    \includegraphics[width=\textwidth]{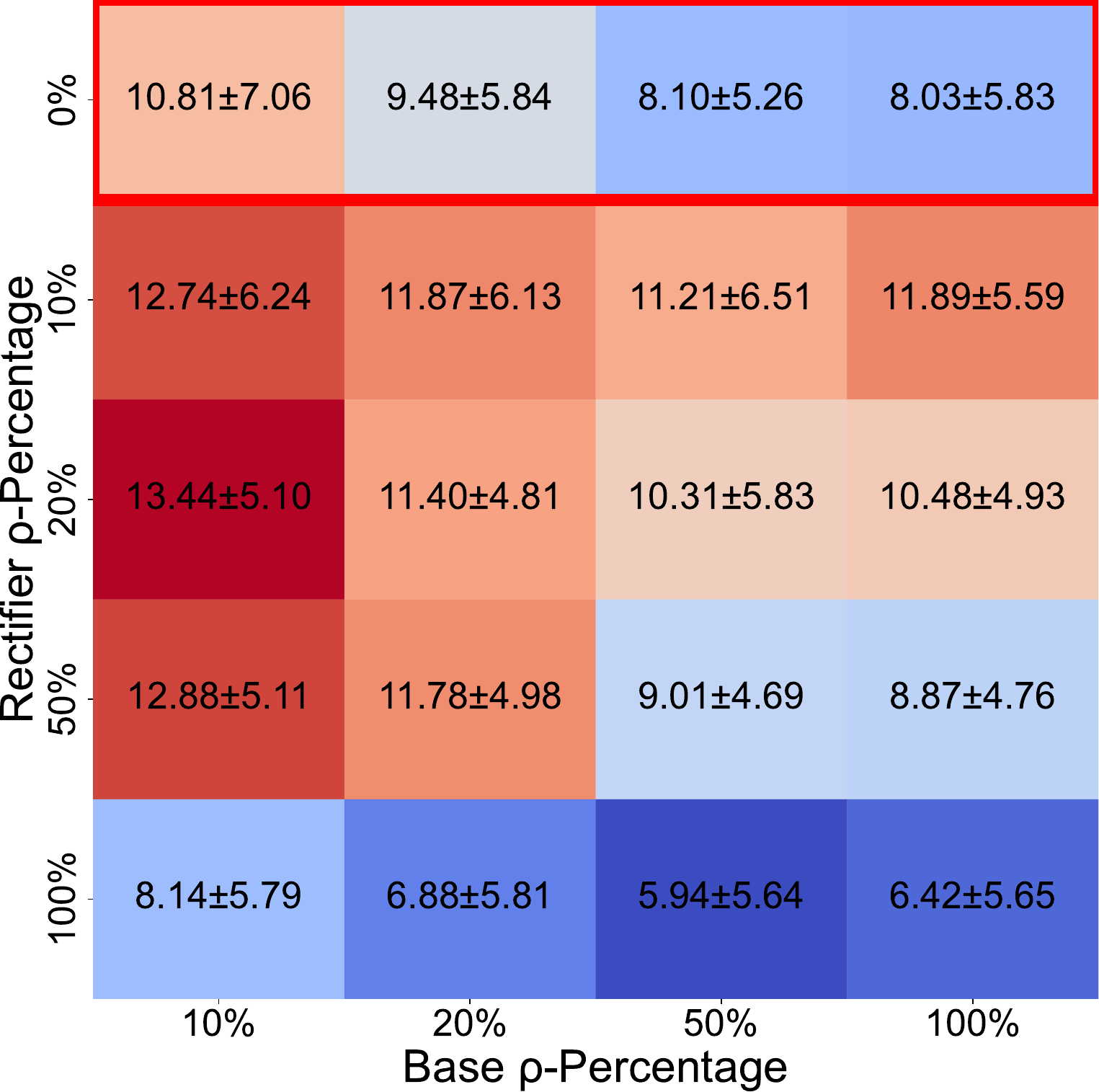}
    \caption{RF}
\end{subfigure} &
\begin{subfigure}{0.48\textwidth}
    \centering
    \includegraphics[width=\textwidth]{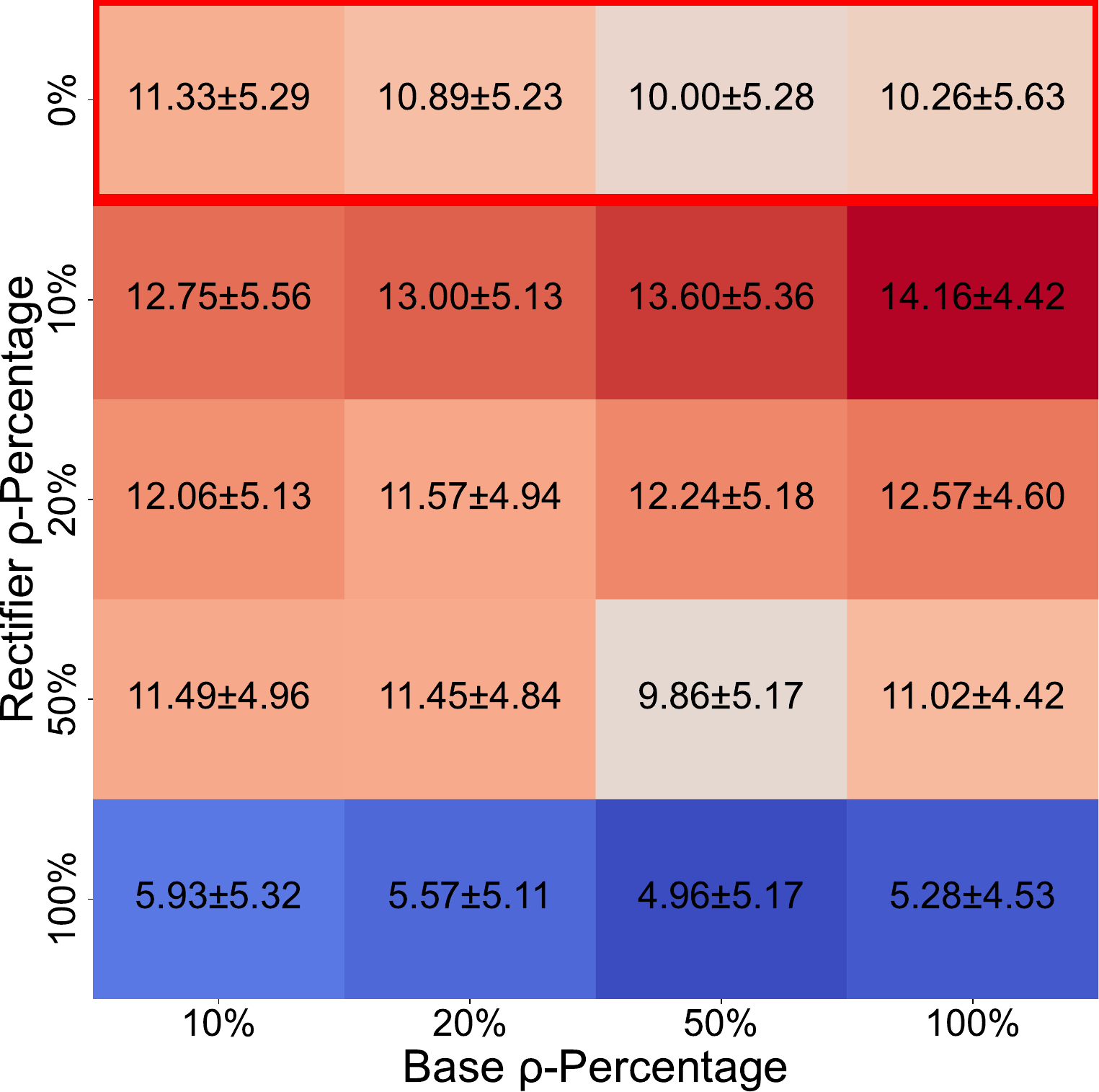}
    \caption{RNN}
\end{subfigure} \\
\begin{subfigure}{0.48\textwidth}
    \centering
    \includegraphics[width=\textwidth]{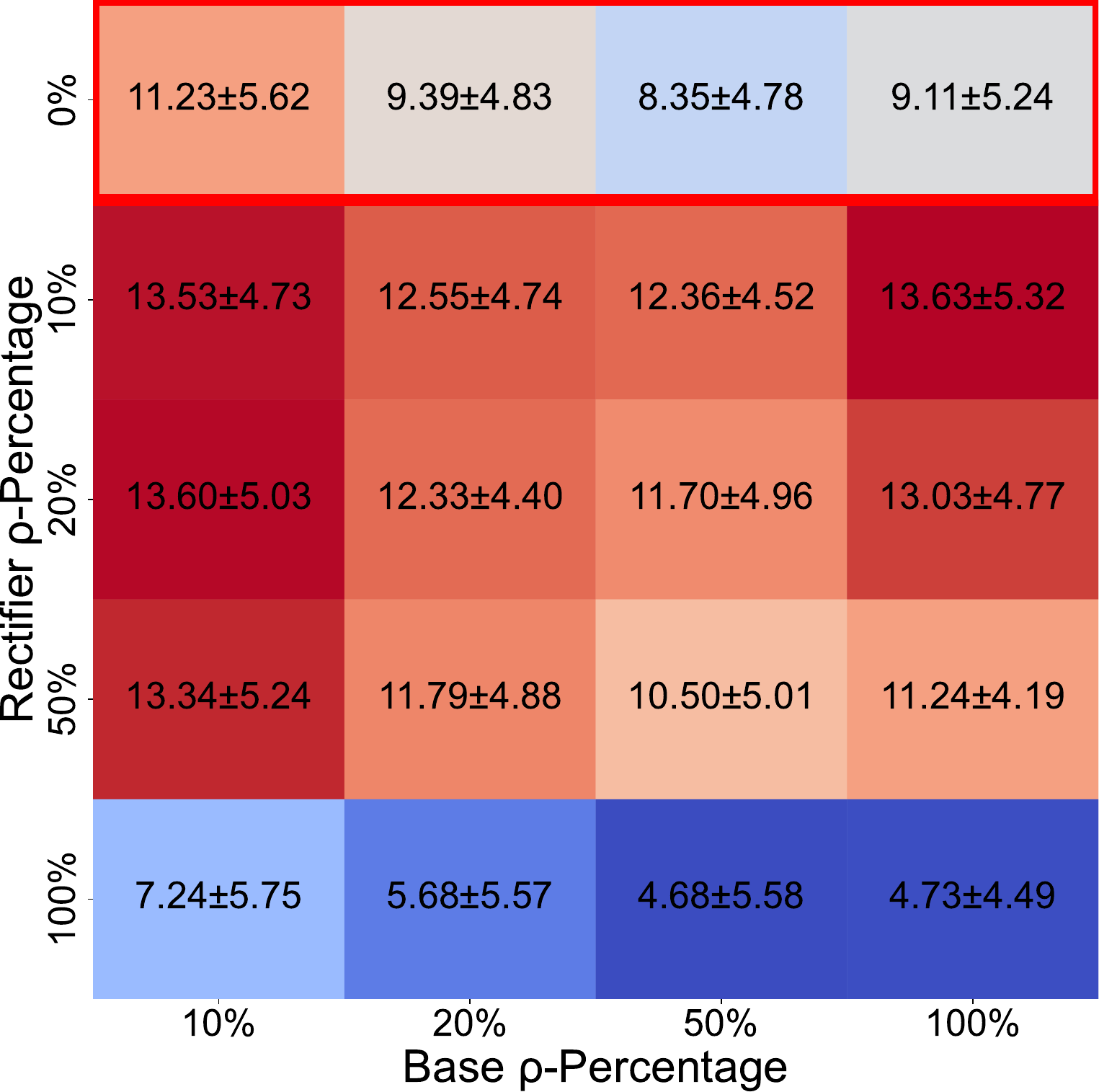}
    \caption{LSTM}
\end{subfigure} &
\begin{subfigure}{0.48\textwidth}
    \centering
    \includegraphics[width=\textwidth]{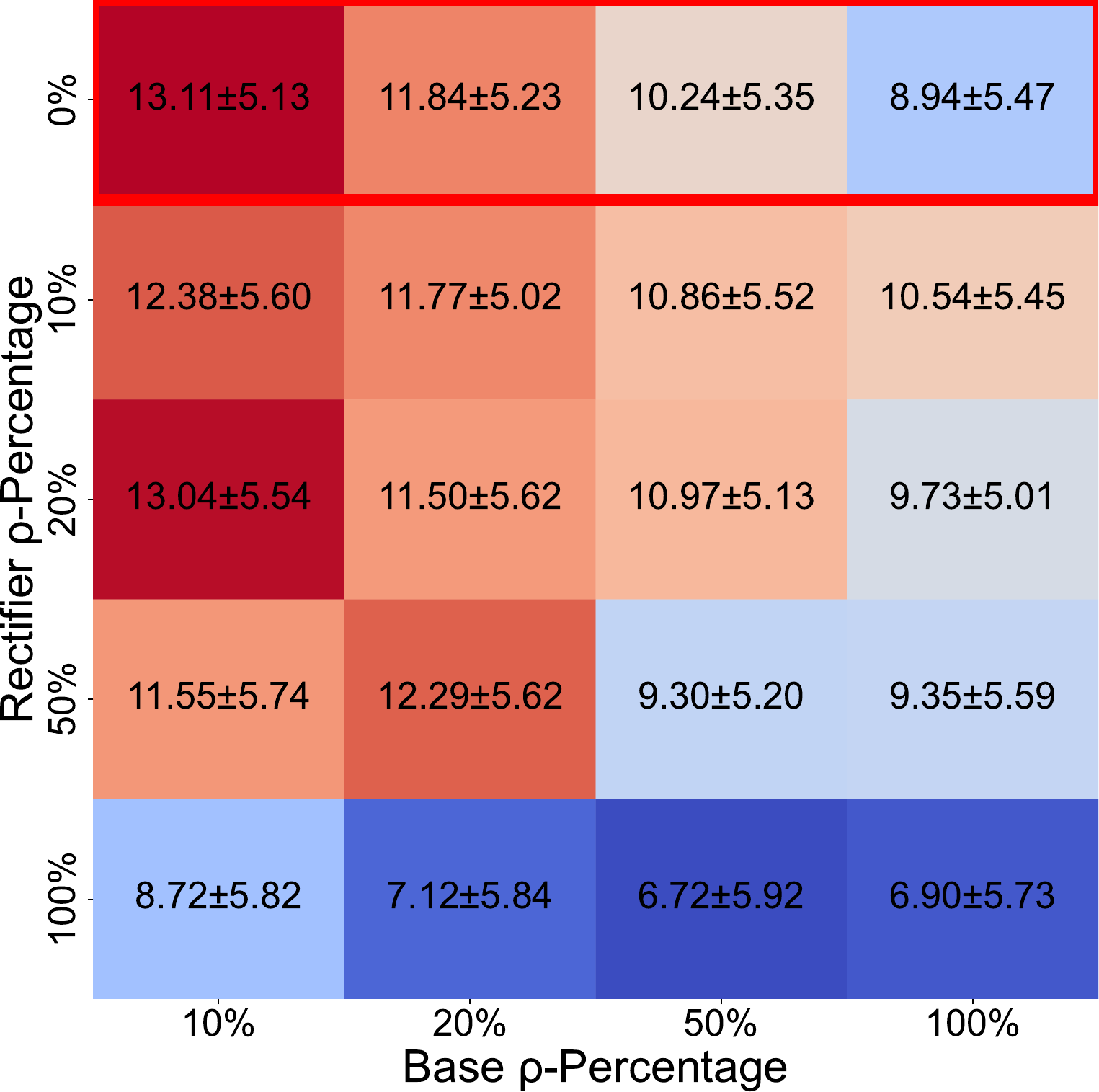}
    \caption{Transformer}
\end{subfigure} &
\\
\end{tabular}
\caption{The $\rho - \rho$ region for the RF (a) RNN (b) LSTM (c) and Transformer (d). 
Cells surrounded by red indicate previously existing strategies, represented by Stratify. Red signifies worse ranking error and blue shows better ranking error. The exact mean ranks and the standard errors are shown in $\pm$. Values are calculated over 18 datasets, all horizons, and 3 seeds.}

\label{fig:rr_region}
\end{figure}

Figure~\ref{fig:rr_region} presents the performance of the $\rho$-$\rho$ region across RF, RNN, LSTM, and Transformer function classes, with ranks normalised from 0 to 20 for consistency.
A similar trend is observed across all models, where the best-performing strategy is consistently a novel one in $\rho:50~\rho:100$. This highlights a strong preference for base functions that predict the forecast horizon in two steps, followed by a longer rectifier. This finding is consistent with Figure \ref{fig:stratify_plane}, where a two-step base strategy and a longer rectifier are generally effective, regardless of the underlying model architecture.

\section{Discussion and Conclusion}\label{discussion}
This paper introduced Stratify, a unified framework for multi-step forecasting 
strategies that unifies existing approaches while enabling the discovery of novel, improved strategies.
Novel \textit{Stratify} strategies consistently outperformed existing ones in over 84\% of experiments across 18 benchmark datasets (Table \ref{percentage_table}) and multiple function classes, with reductions in error between $5-25\%$ across multiple function classes (Table \ref{rel_mse_table}),  addressing our first research question (R1).
We showed that existing strategies failed to perform significantly differently with 95\% confidence under the Nemenyi test. However, novel strategies in \textit{Stratify} were shown to significantly improve forecast performance under this test (Figures \ref{fig:critical_differencemlp}, \ref{fig:critical_differencernn}). Both of these findings addressed our second research question (R2).
Our representations of \textit{Stratify} through the $\rho$-$\delta$-$\iota$ planes revealed general trends, suggesting them to be a reasonable representation of the space of strategies (R3). Despite the general trends of the heat maps, the high variances reported highlight the importance of task-specific selection of strategies.

We presented the most comprehensive benchmarking of multi-step forecasting strategies by evaluating all existing strategies and introducing a space of novel ones on 18 benchmarks, multiple horizon lengths, and five function classes. In this work we represented the parameterisation of strategies as a percentage of their forecasting length, which allowed for a more fair and intuitive comparison across strategies and their task settings. We hope for future works to consider the same when comparing strategies across different horizon lengths.

For practitioners, our work unifies the relationship between existing strategies. \textit{Stratify} offers a systematic methodology to discover high-performing forecasting strategies without treating each strategy in isolation. The planes from Figure \ref{fig:stratify_plane} can be searched via an optimisation routine to find an optimal strategy. Future work can investigate the use of various optimisation algorithms to navigate the vast \textit{Stratify} space, or utilise meta-learning to identify whether general task features have a relationship with the optimal strategy. This is particularly valuable for real-world applications where datasets exhibit diverse patterns and characteristics, making one-size-fits-all approaches ineffective. The insights provided by the framework, such as the preference for longer base strategies and rectifiers, also simplify the selection process for hyperparameters, reducing trial-and-error experimentation.

Whilst we computed the entire \textit{Stratify} plane for the MLP, we did not for the remaining function classes. Future work can investigate the relationship between the number of functions required for each strategy and the performance. From the performance heat-map in Figure \ref{fig:stratify_plane} and the computational time in Figure \ref{fig:compute_plane}, we hypothesise that compute-optimal strategy selection would be possible and highly beneficial for practitioners. More efficient exploration of the \textit{Stratify} space is expected to improve forecasting performances.

Lastly, the results demonstrate the effectiveness of the proposed \textit{Stratify} framework, but we acknowledge this work focuses exclusively on univariate time series data. Many real-world applications involve multivariate time series, where interactions between variables play a critical role in forecasting. Extending the framework to handle multivariate scenarios would significantly enhance its applicability and generality.


\backmatter
\bibliography{biblio}

\newpage
\begin{appendices}

\section{Training time over \textit{Stratify} space}

\begin{figure}[hbt!]
\centering
\begin{tabular}{ccc}
\includegraphics[width=0.31\textwidth]{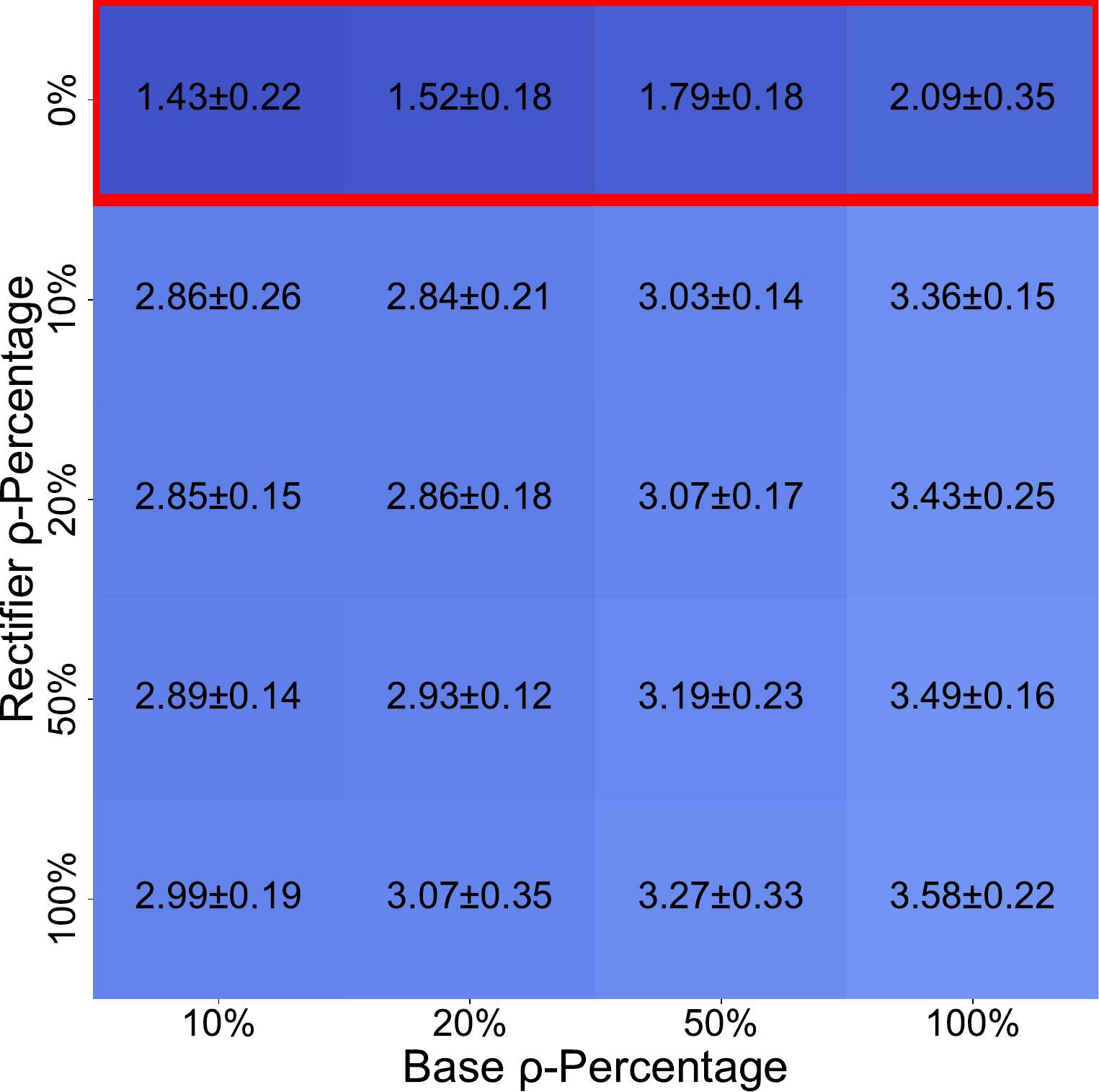} &
\includegraphics[width=0.31\textwidth]{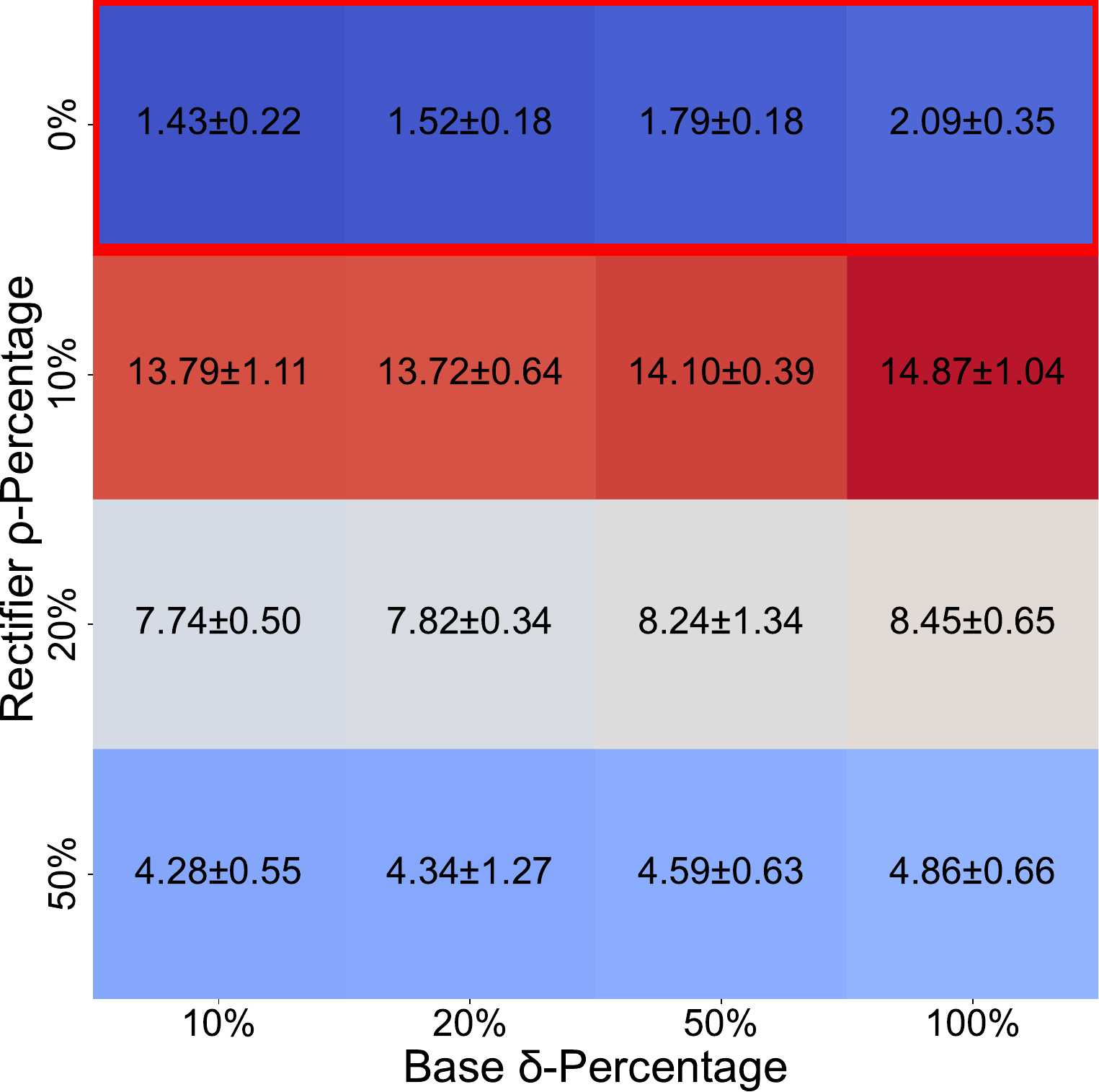} &
\includegraphics[width=0.31\textwidth]{
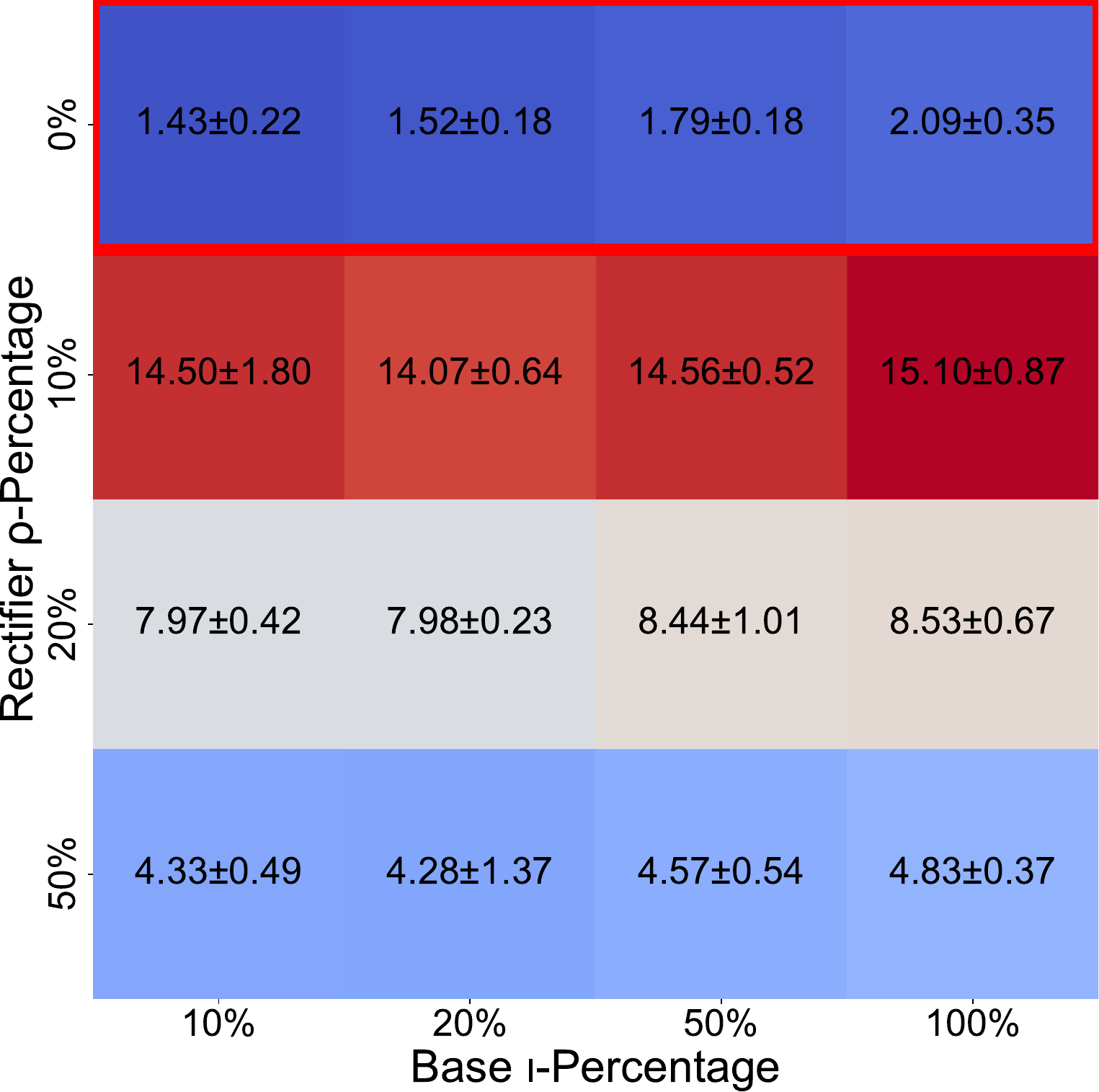} \\
\includegraphics[width=0.31\textwidth]{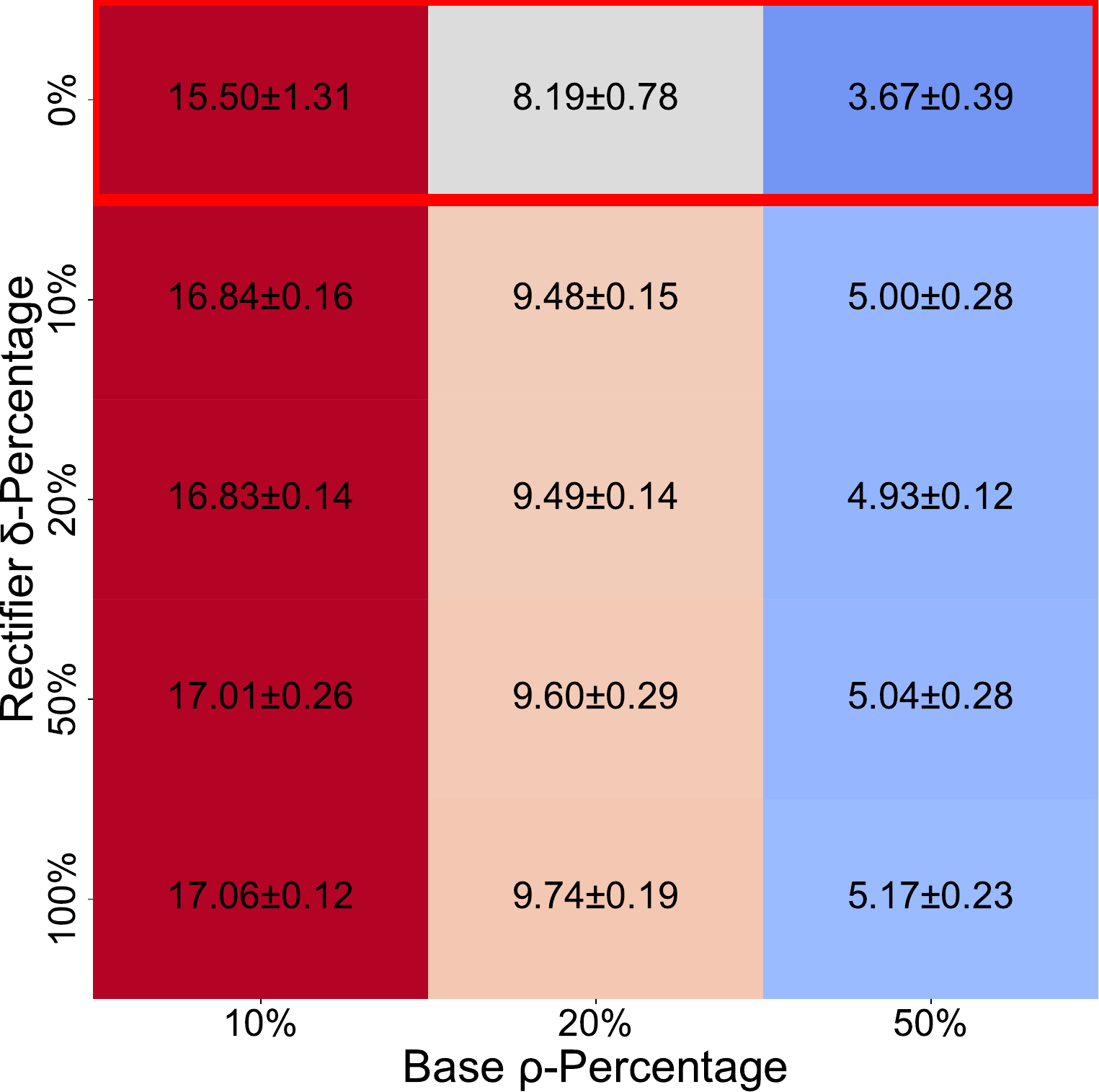} &
\includegraphics[width=0.31\textwidth]{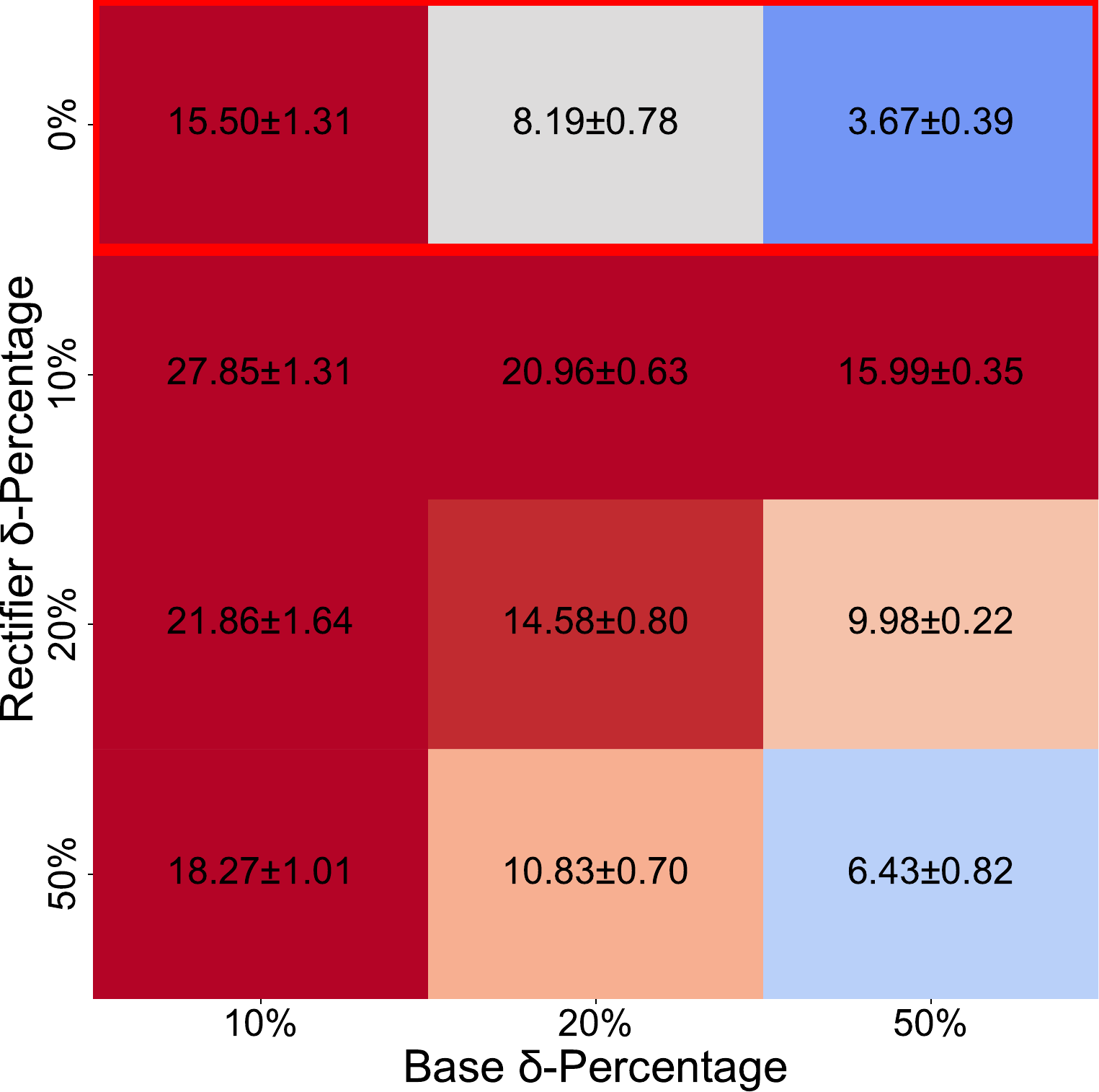} &
\includegraphics[width=0.31\textwidth]{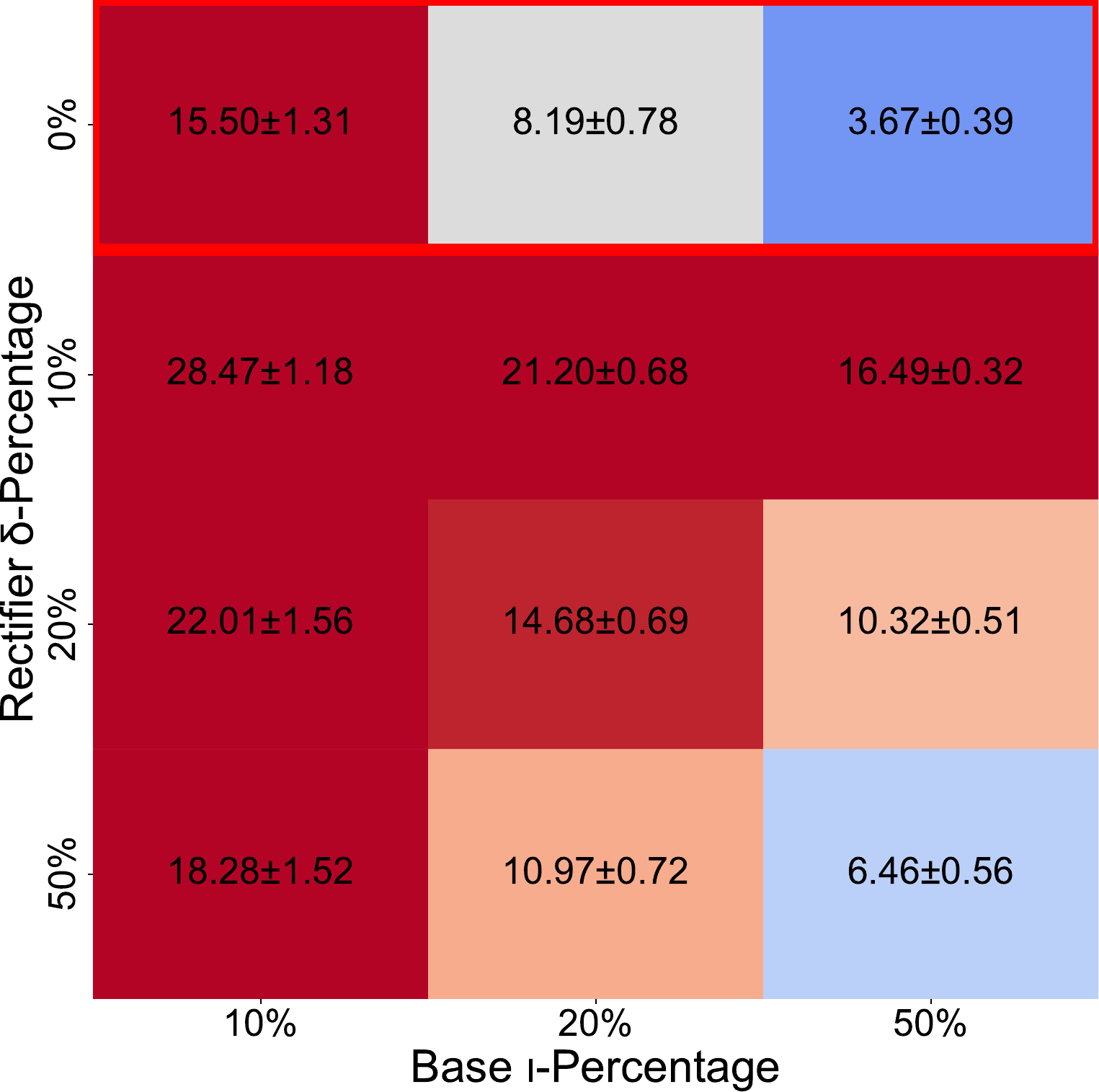} \\
\includegraphics[width=0.31\textwidth]{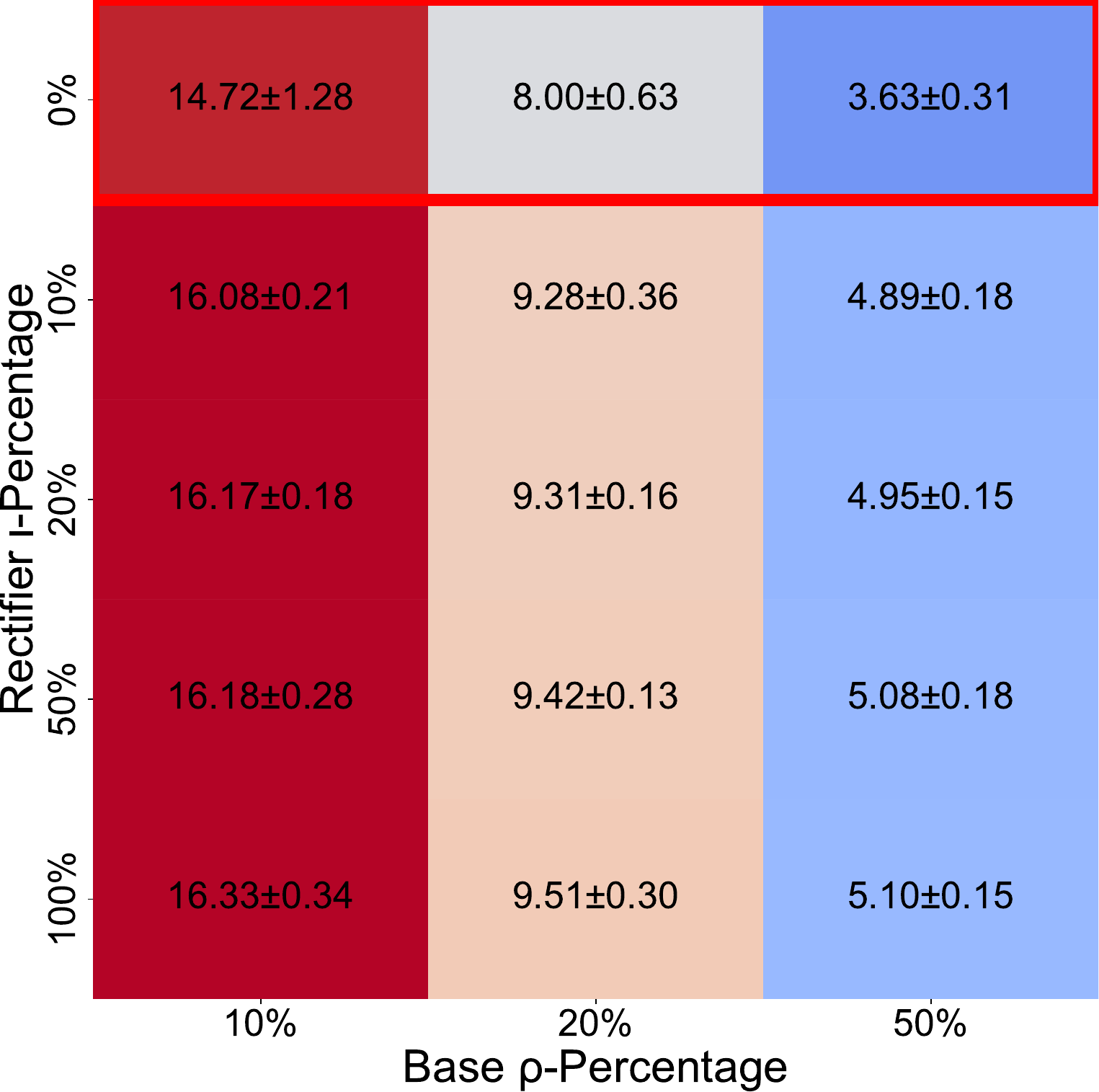} &
\includegraphics[width=0.31\textwidth]{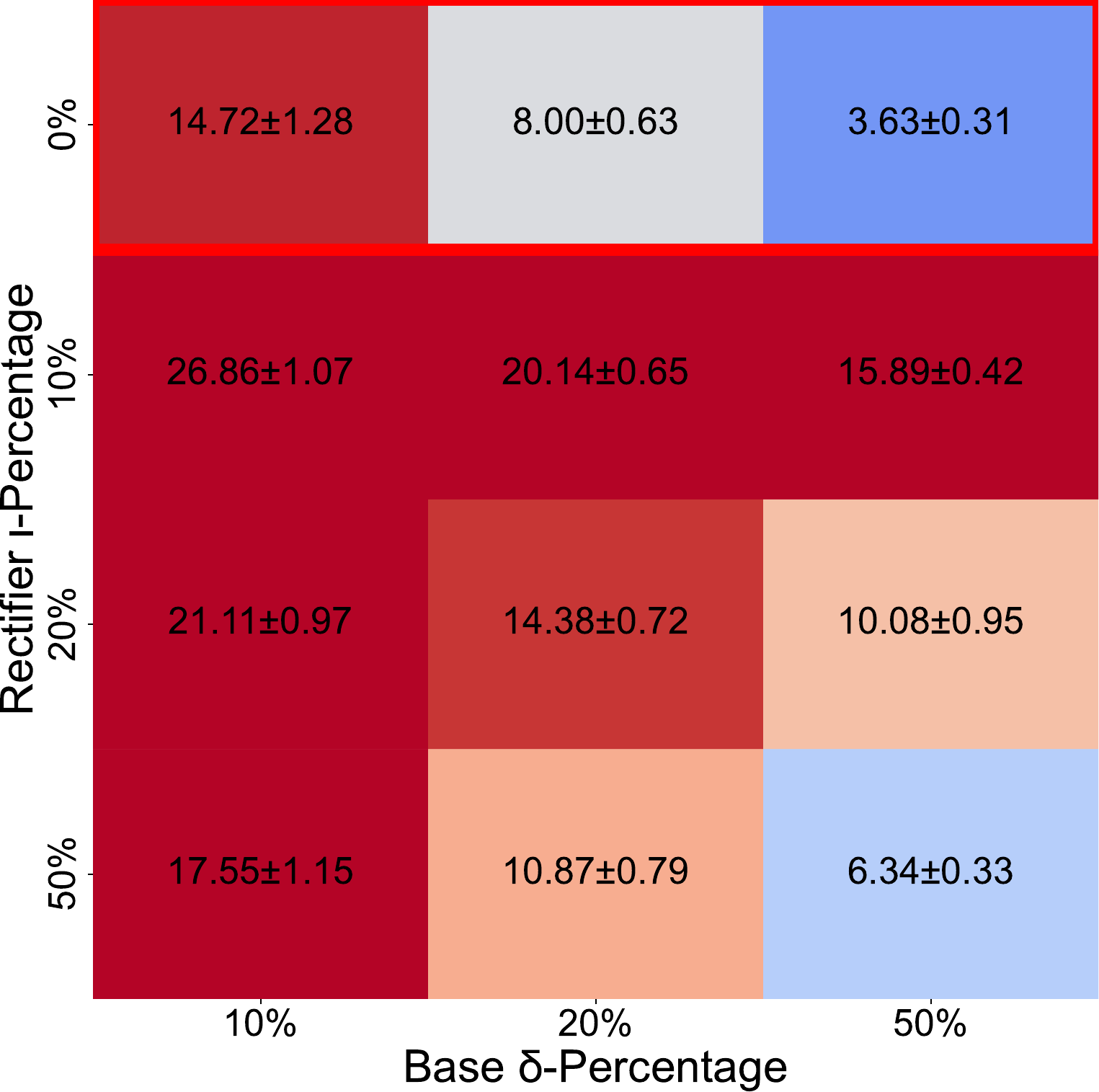} &
\includegraphics[width=0.31\textwidth]{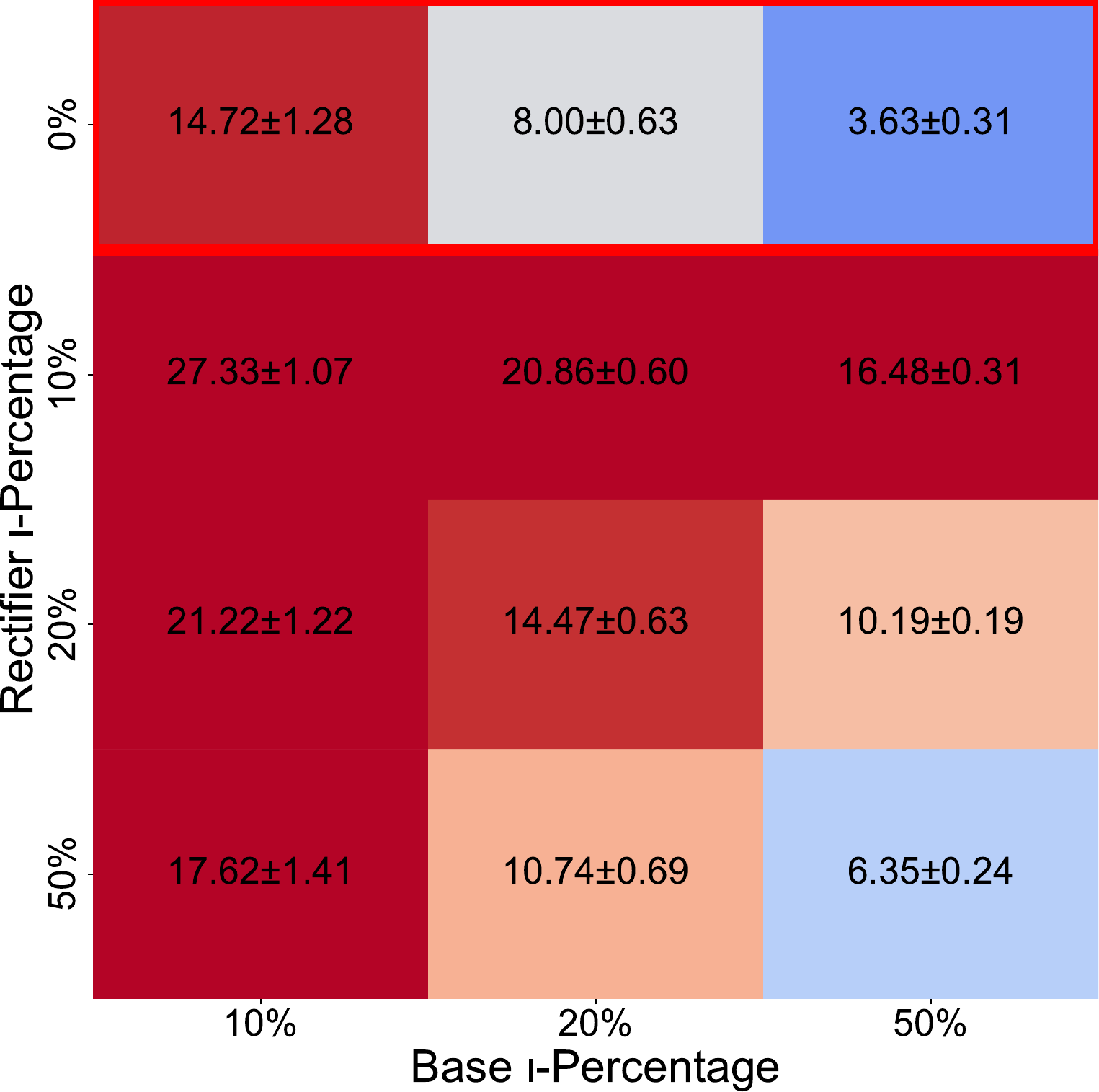} \\
\end{tabular}
\caption{Compute time for strategies over the \textit{Stratify} plane. Times to train MLP on Mackey Glass dataset for horizon 10.}
\label{fig:compute_plane}
\end{figure}

In Figure \ref{fig:compute_plane} we show the time taken to train each strategy in \textit{Stratify} for the MLP on the mg\_10000 dataset. We expect the qualitative times to be consistent across datasets and functions used. The training time is normalised by the minimum time taken for a single strategy to train. There is a clear relationship showing that strategies containing RecMO train much faster. This is because it is the only strategy where forecasts are computed using only one model.

\newpage

\section{Inference time over \textit{Stratify} space}

\begin{figure}[hbt!]
\centering
\begin{tabular}{ccc}
\includegraphics[width=0.31\textwidth]{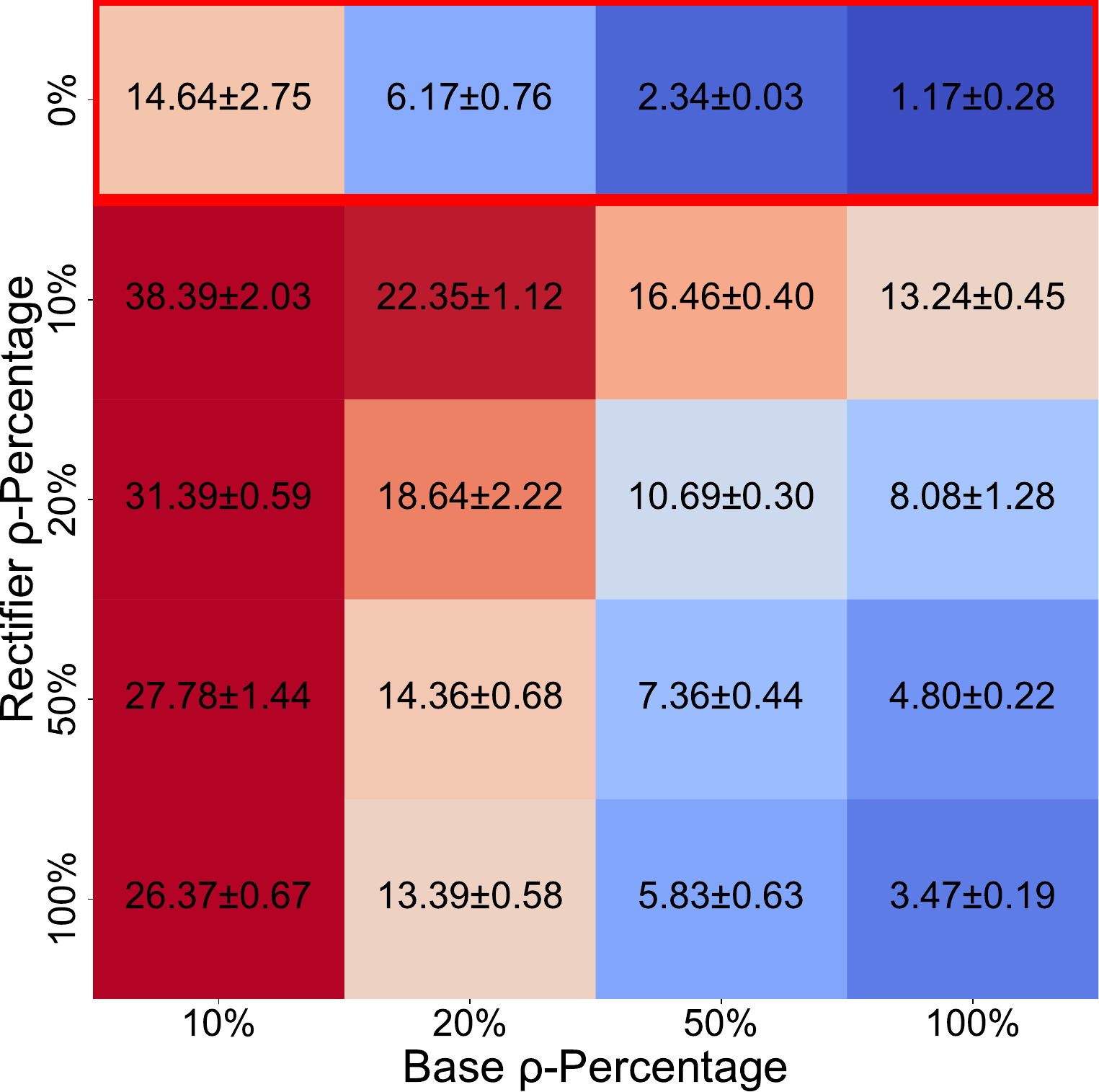} &
\includegraphics[width=0.31\textwidth]{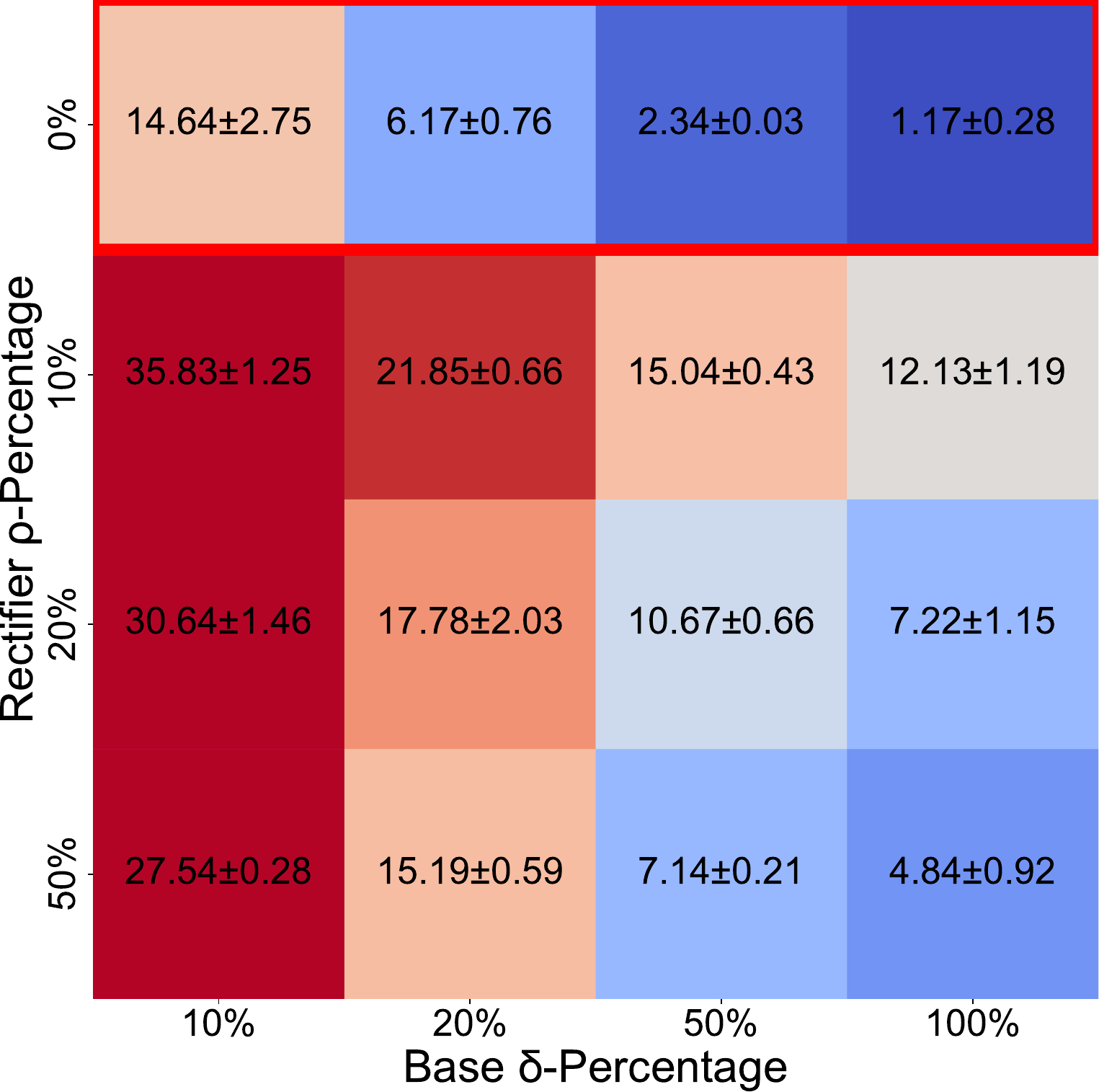} &
\includegraphics[width=0.31\textwidth]{
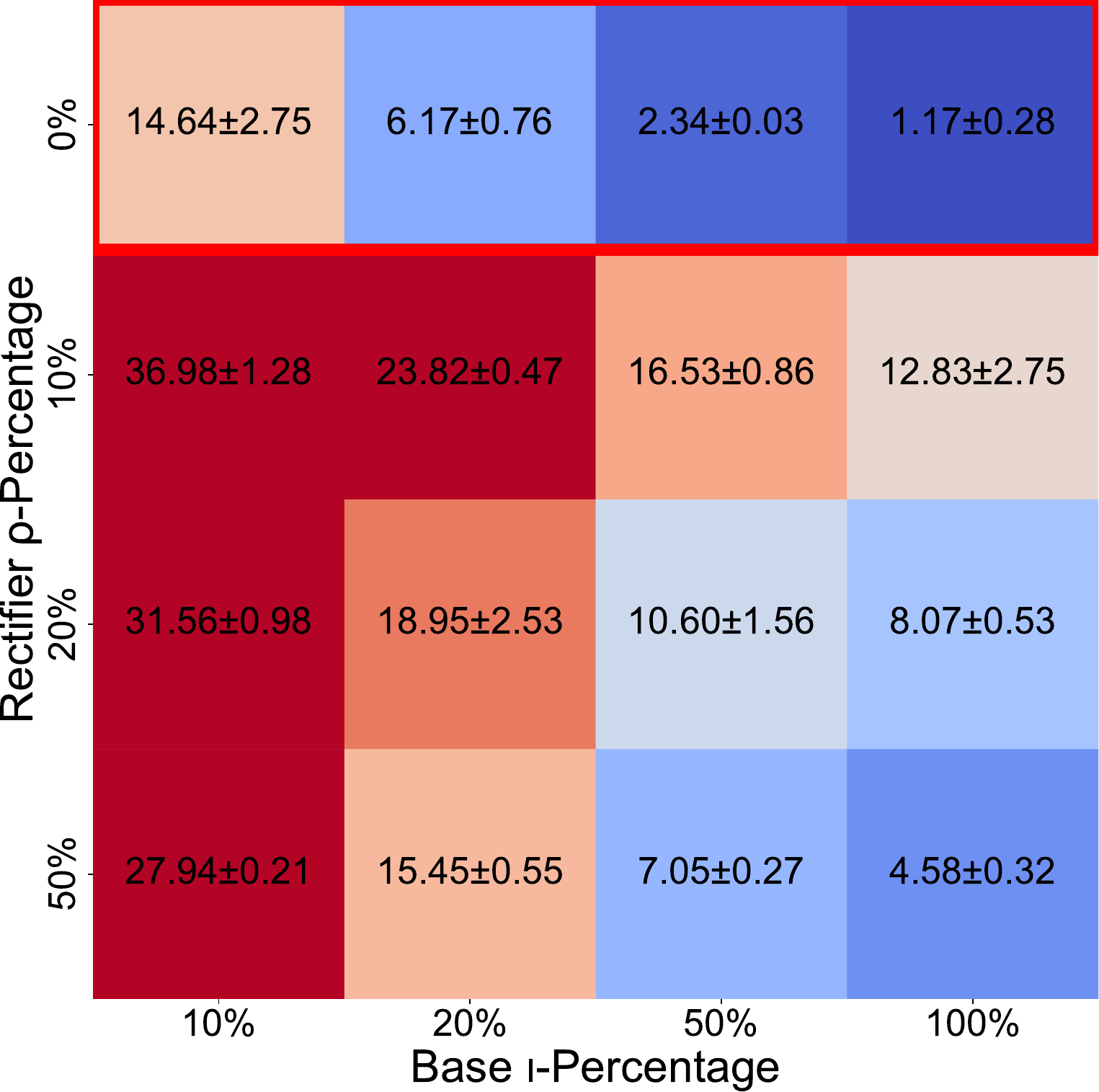} \\
\includegraphics[width=0.31\textwidth]{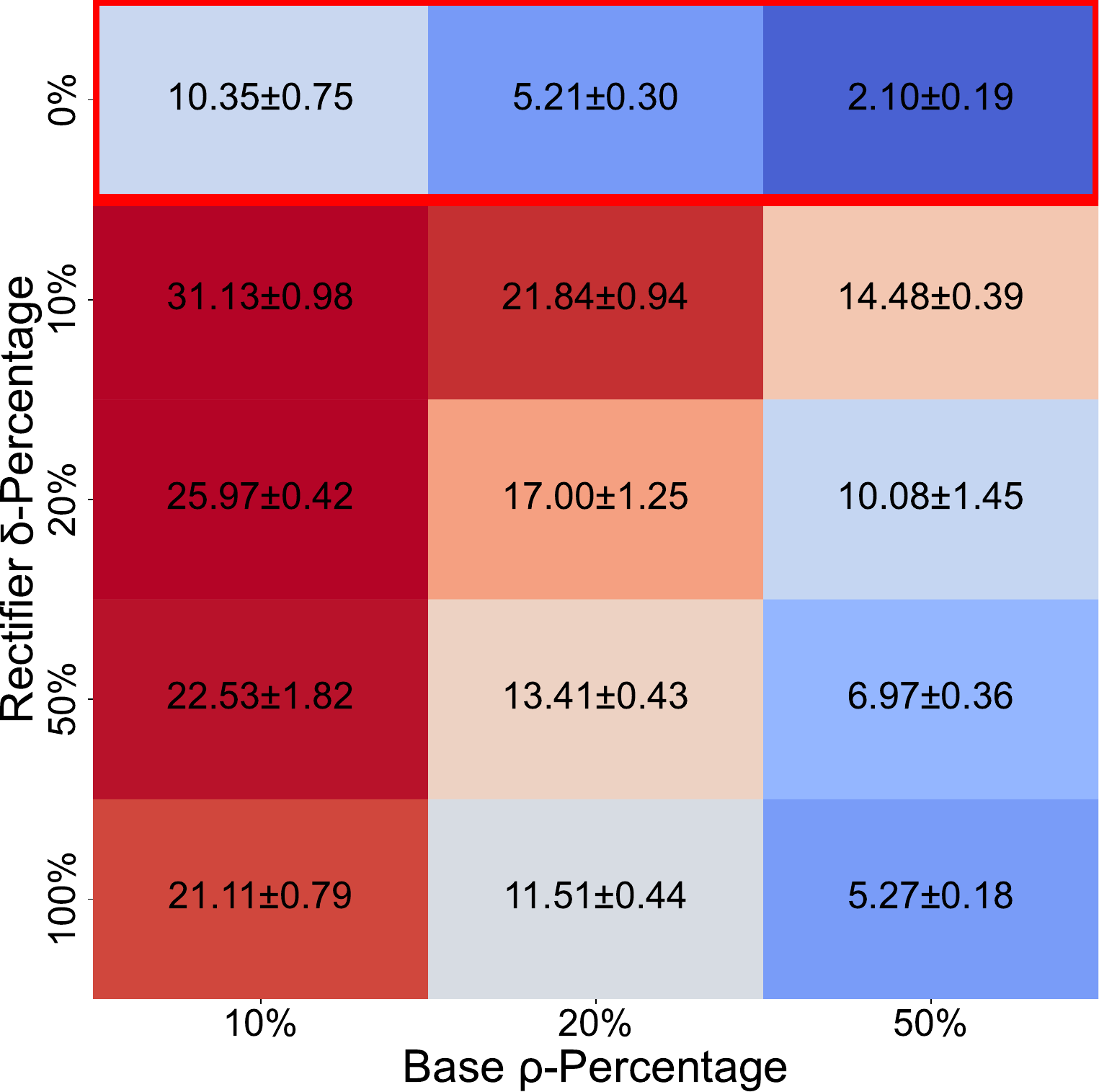} &
\includegraphics[width=0.31\textwidth]{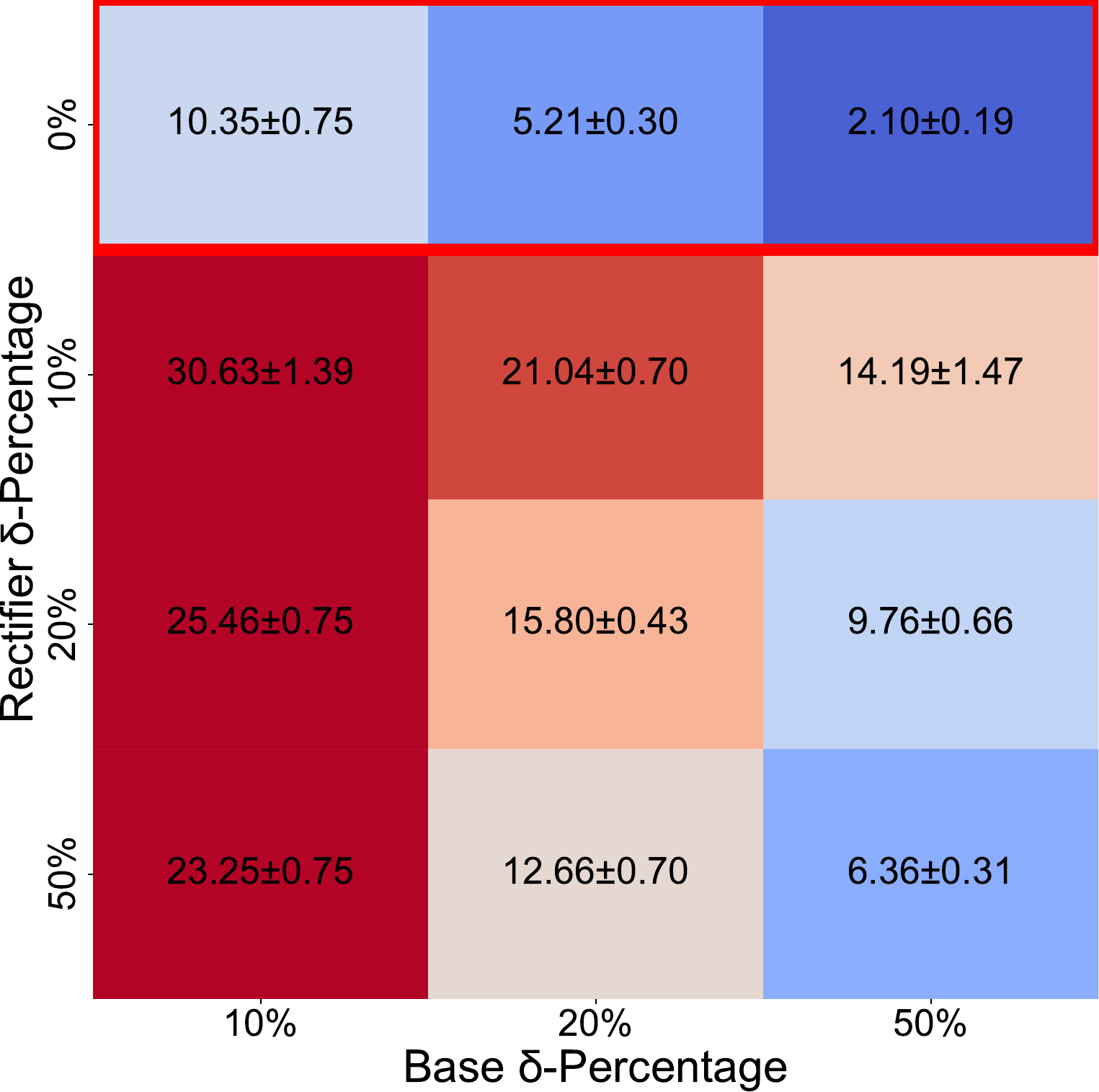} &
\includegraphics[width=0.31\textwidth]{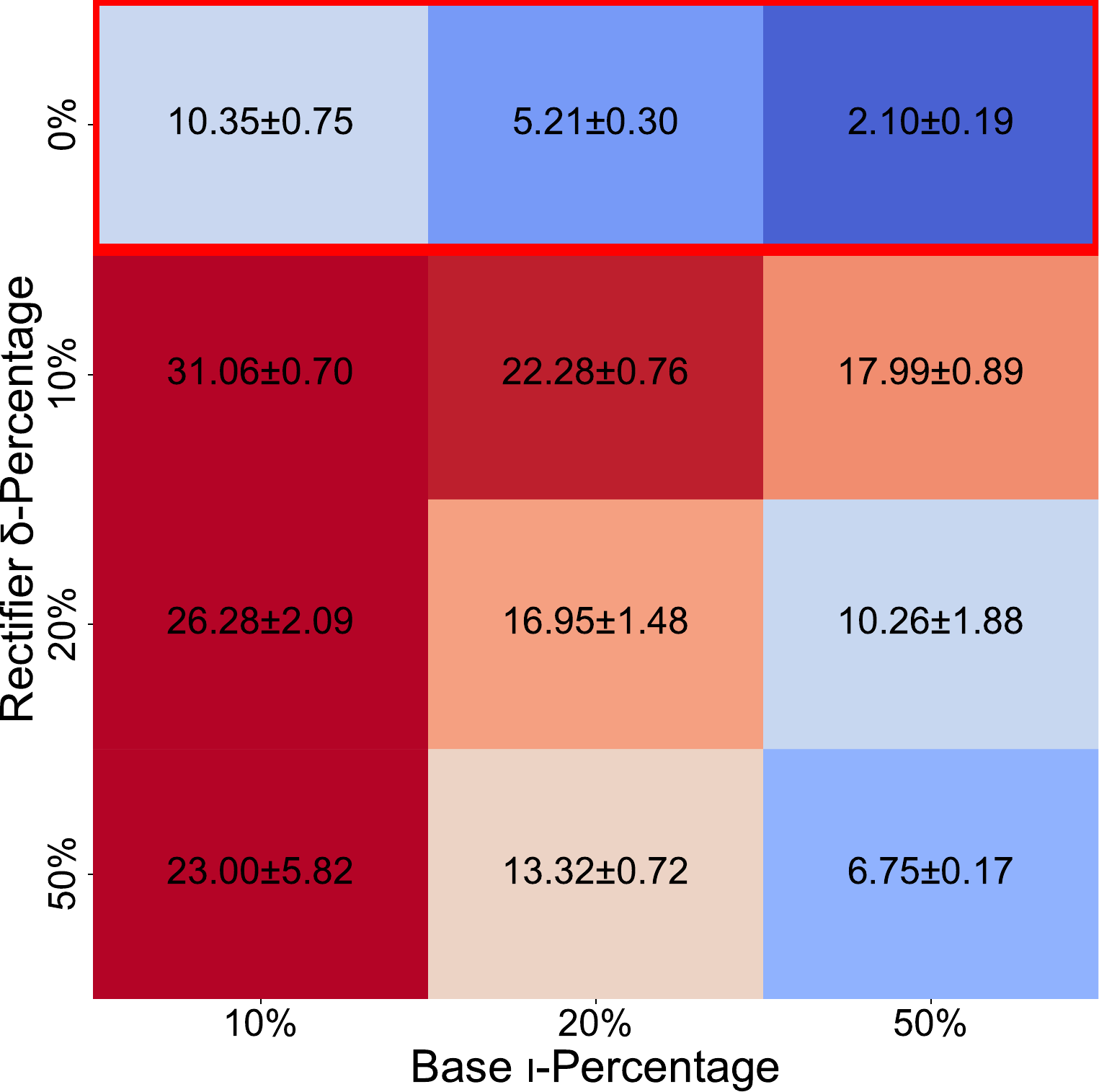} \\
\includegraphics[width=0.31\textwidth]{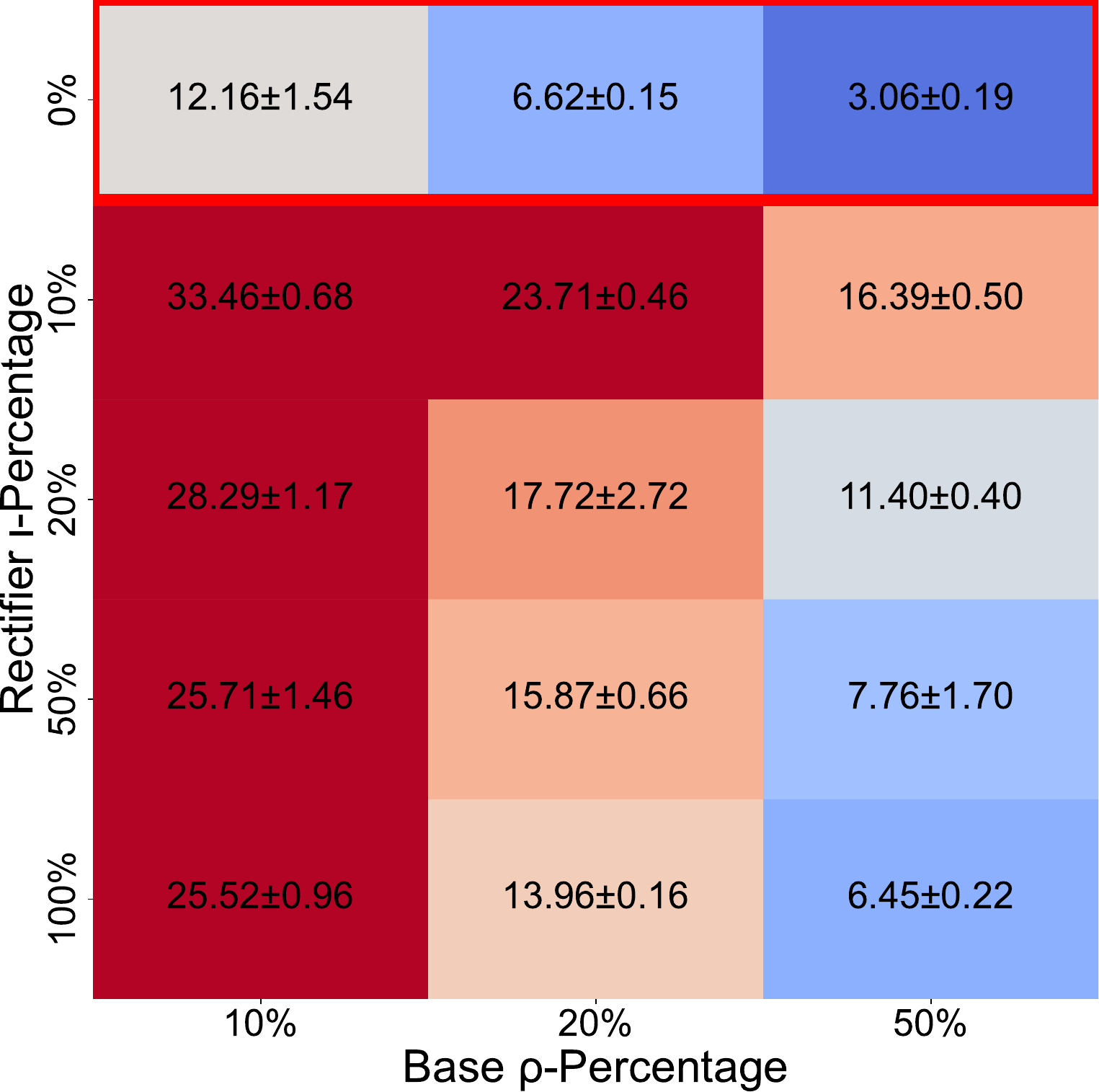} &
\includegraphics[width=0.31\textwidth]{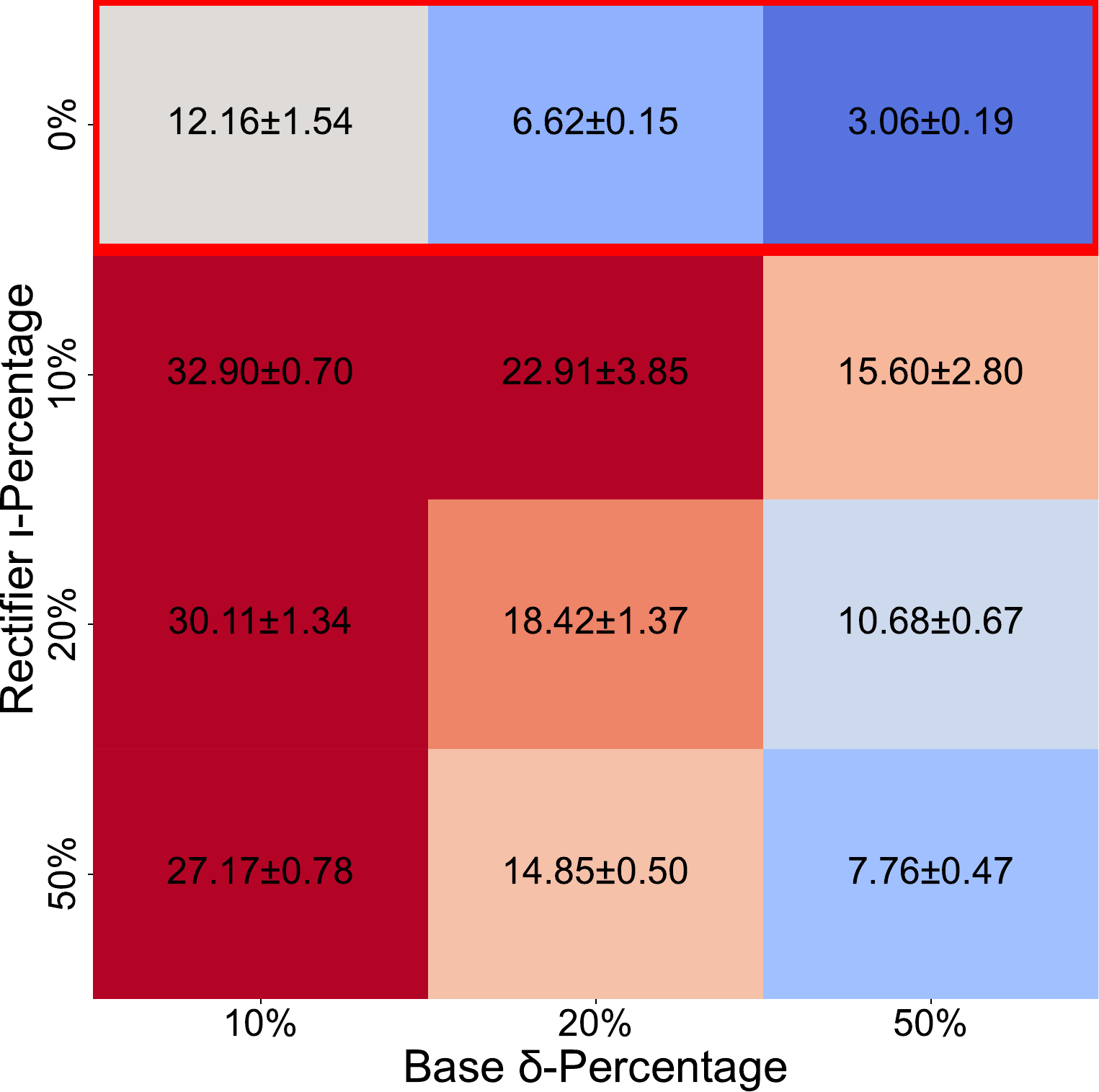} &
\includegraphics[width=0.31\textwidth]{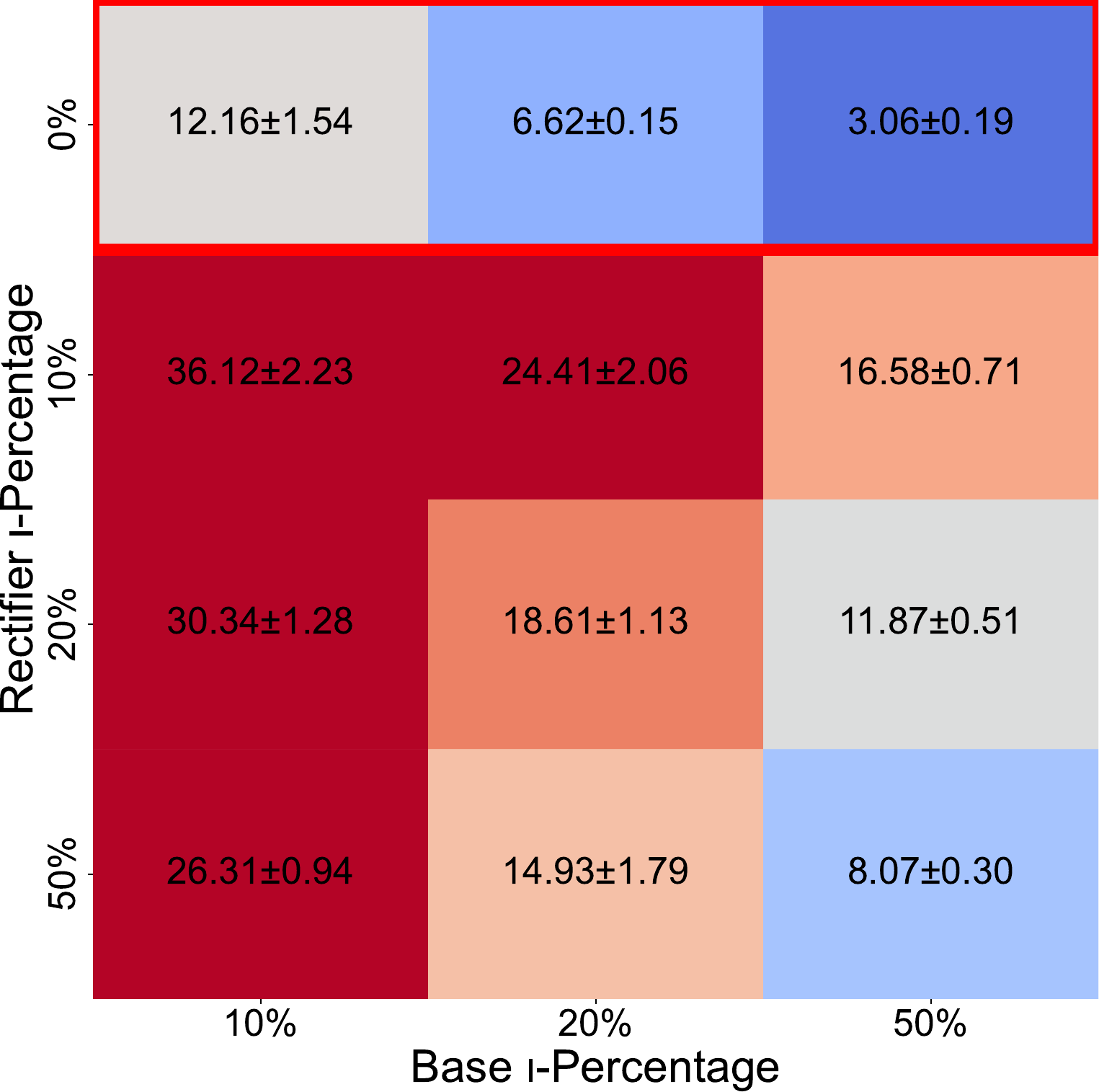} \\
\end{tabular}
\caption{Inference time for strategies over the \textit{Stratify} plane. Times to train MLP on Mackey Glass dataset for horizon 10.}
\label{fig:inference_plane}
\end{figure}

In Figure \ref{fig:inference_plane} we show the inference time taken for each strategy in \textit{Stratify} for the MLP on the mg\_10000 dataset. As with the compute times in Figure \ref{fig:compute_plane}, we expect the qualitative times to be consistent across datasets and functions used. Again, the time is normalised by the minimum time taken to produce forecasts for the task. We see a clear pattern over inference times, showing that the parameterisation is proportional to the inference time. This is to be expected for RecMO and DirRecMO strategies. This relationship would be alleviated for DirMO if the implementation is to allocate parallel computing methods over the set of functions of a DirMO-only strategy.

\newpage

\section{Remaining Critical Difference Diagrams}

In the main text we truncate the critical differencing diagram for space reasons. Below we show the full diagram of every strategy in \textit{Stratify} using the normalisation of parameter values by the horizon length. We find that many strategies in \textit{Stratify} rank on either side of existing strategies. However the majority of green is visible on the left of the blue, highlighting that \textit{Stratify} strategies are generally showing improved performance.

\begin{figure}[hbt!]
    \centering
    \includegraphics[width= 0.75\linewidth]{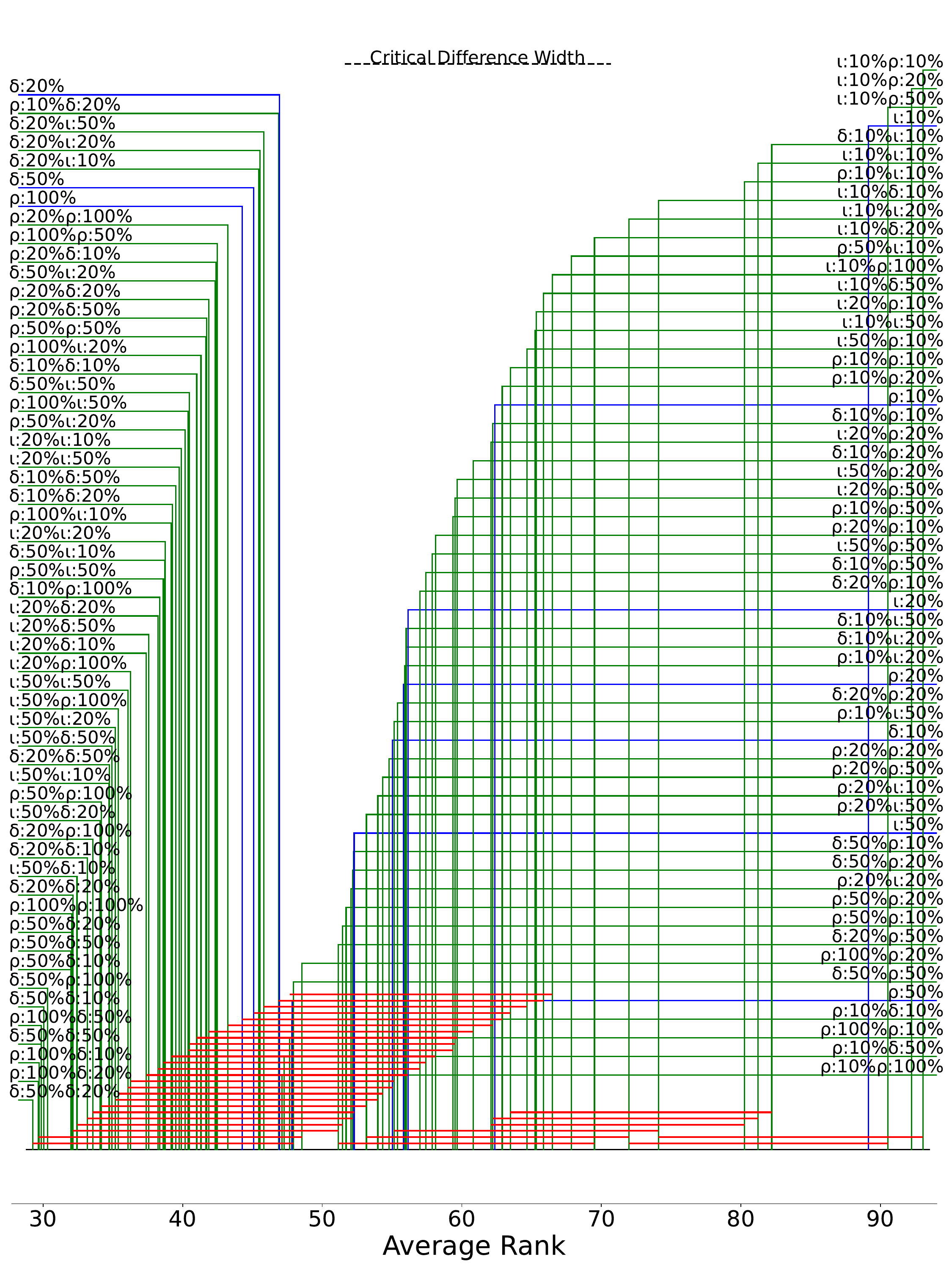}
    \caption{Critical difference diagram for MLP. \textit{Stratify} (in green) and all previous methods (in blue). Cliques at the 95\% confidence are shown by the red lines.}
    \label{fig:fullmlp_cd}
\end{figure}

\newpage


For the RF, the critical difference diagram is qualitatively similar to the MLP and Transformer, where the best strategies in \textit{Stratify} are significantly better than some existing methods, whereas no existing method is significantly better than any other.

\begin{figure}[hbt!]
    \centering
    \includegraphics[width=\linewidth]{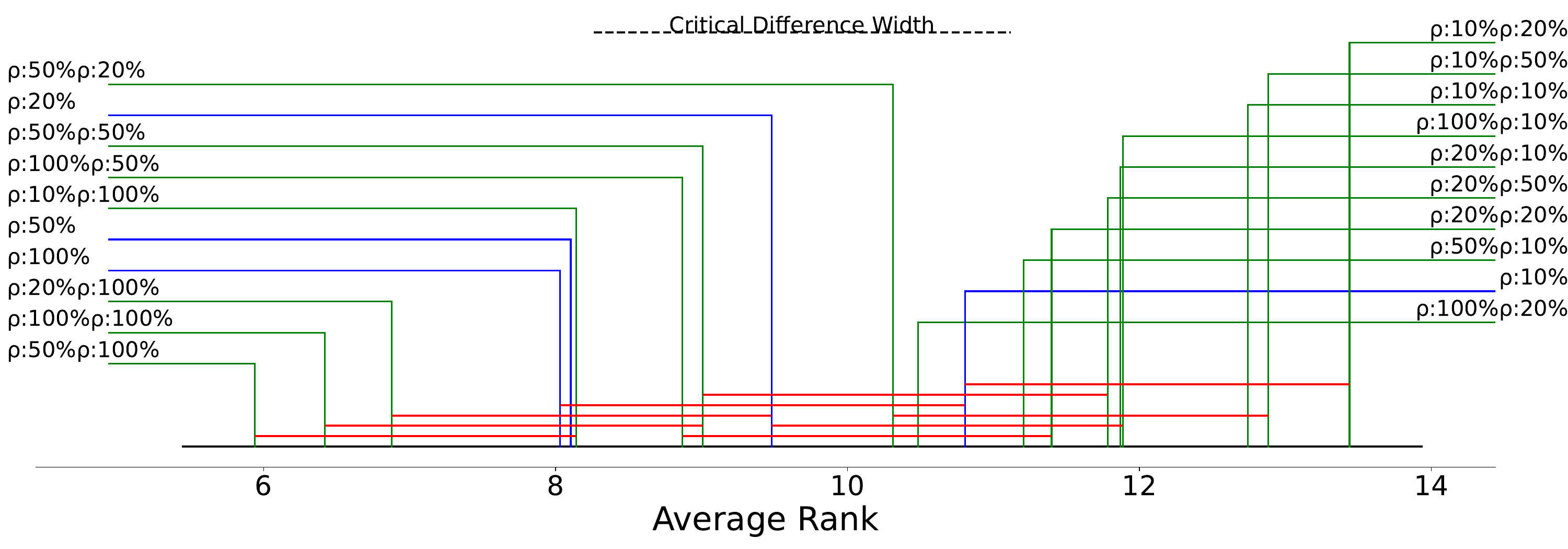}
    \caption{Critical difference diagram for RF. \textit{Stratify} (in green) and all previous methods (in blue). Cliques at the 95\% confidence are shown by the red lines.}
    \label{fig:critical_differencerf}
\end{figure}

For the Transformer, the critical difference diagram is qualitatively similar to the MLP and RF, where the best strategies in \textit{Stratify} are significantly better than some existing methods, whereas no existing method is significantly better than any other.

\begin{figure}[hbt!]
    \centering
    \includegraphics[width=\linewidth]{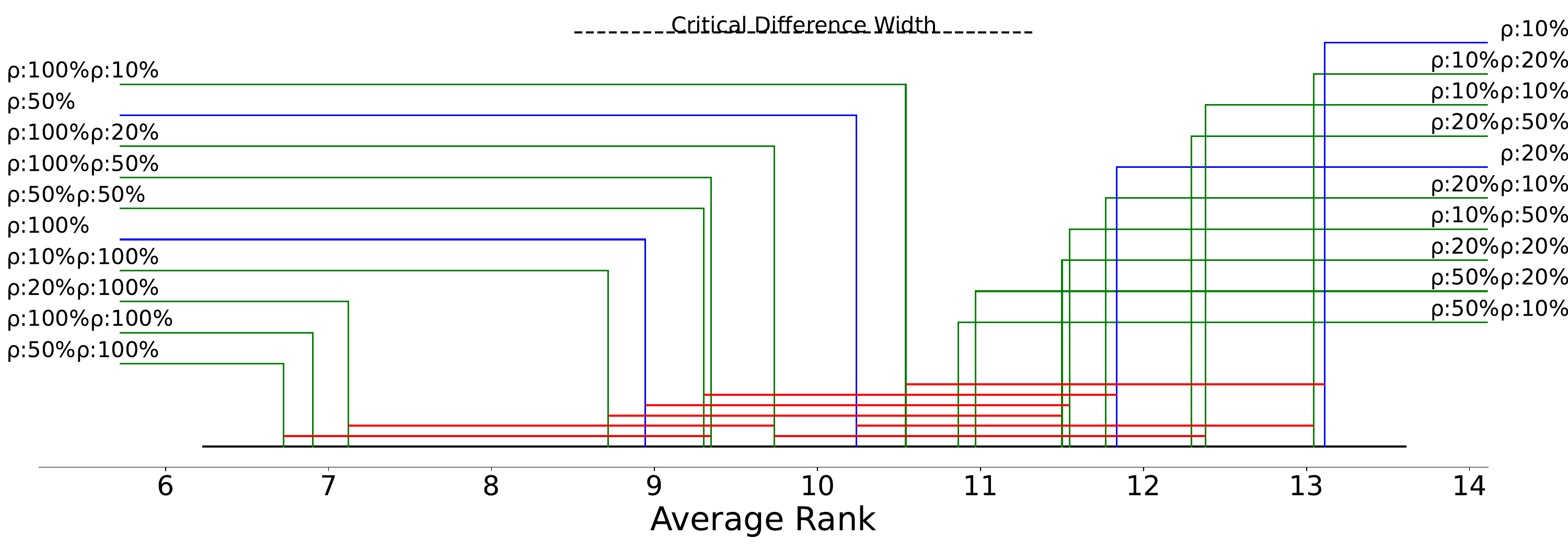}
    \caption{Critical difference diagram for Transformer. \textit{Stratify} (in green) and all previous methods (in blue). Cliques at the 95\% confidence are shown by the red lines. }
    \label{fig:critical_differencetrans}
\end{figure}

\newpage

\begin{figure}[hbt!]
    \centering
    \includegraphics[width=\linewidth]{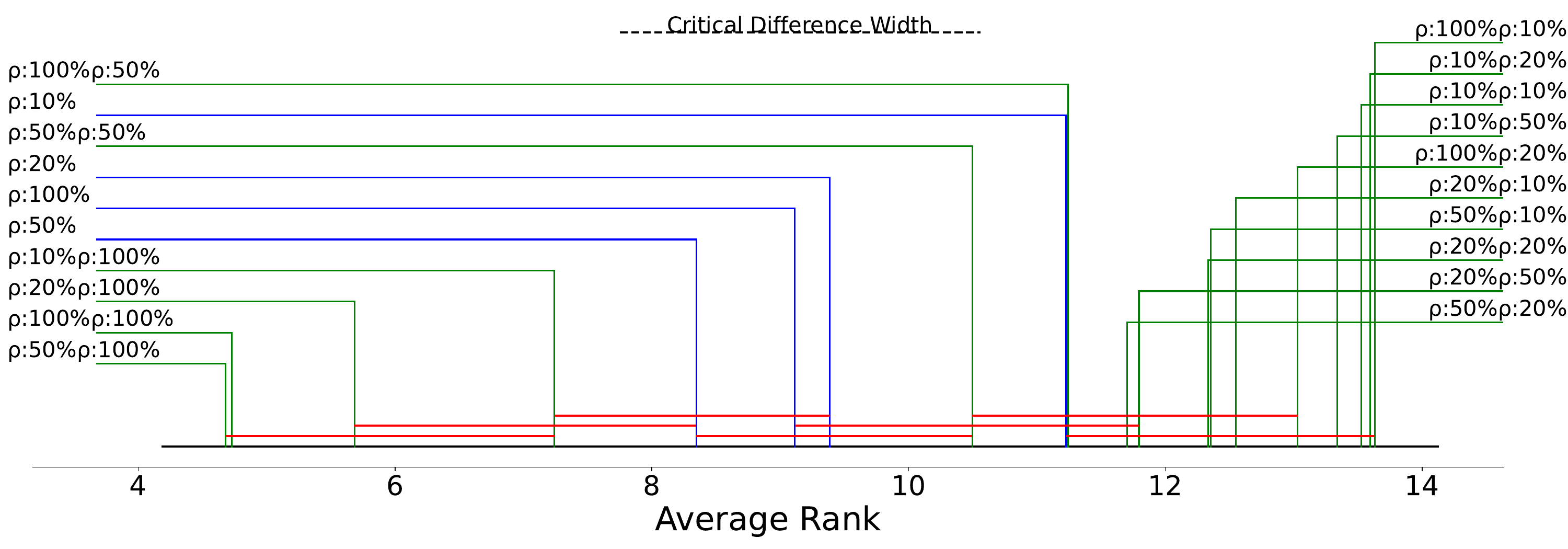}
    \caption{Critical difference diagram for LSTM. \textit{Stratify} (in green) and all previous methods (in blue). Cliques at the 95\% confidence are shown by the red lines. }
    \label{fig:critical_differencelstm}
\end{figure}
For the LSTM, the critical difference diagram is qualitatively similar to the RNN, where the best performing novel strategies are significantly improved compared to 100\% of existing methods.





\section{All raw errors}\label{rawerrors}

In the main text we only show the relative errors of the optimal strategies in the existing literature and novel ones in Stratify. For completeness, we present the raw errors alongside the error of a mean-forecast baseline. We find that all strategies generally were trained with relatively good generalisation performance. 


\begin{table}[hbt!]
    \centering
 \tiny
\begin{tabular}{llllll}
\toprule
Dataset name & 10 & 20 & 40 & 80 & MFE \\
\midrule
Traffic & 3.68e-5 $\pm$ 7e-7 & 3.93e-5 $\pm$ 2e-6 & 4.50e-5 $\pm$ 2e-6 & 5.04e-5 $\pm$ 2e-6 & 1.00e-3 \\
METR-LA & 8.12e+1 $\pm$ 2e-1 & 1.16e+2 $\pm$ 3e-1 & 1.47e+2 $\pm$ 1e+0 & 1.81e+2 $\pm$ 2e+0 & 2.24e+2 \\
Illness & 1.77e+10 $\pm$ 2e+9 & 1.80e+10 $\pm$ 2e+9 & 1.76e+10 $\pm$ 2e+9 & 1.52e+10 $\pm$ 3e+9 & 2.72e+9 \\
mg\_10000 & 5.67e-5 $\pm$ 6e-5 & 2.48e-5 $\pm$ 1e-6 & 4.04e-5 $\pm$ 8e-6 & 1.38e-4 $\pm$ 5e-5 & 1.20e-1 \\
ExchangeRate & 1.07e-4 $\pm$ 1e-5 & 1.33e-4 $\pm$ 7e-6 & 1.57e-4 $\pm$ 1e-5 & 2.44e-4 $\pm$ 7e-6 & 6.00e-3 \\
ETTm1 & 6.54e-1 $\pm$ 6e-4 & 9.85e-1 $\pm$ 2e-3 & 1.29e+0 $\pm$ 1e-2 & 1.48e+0 $\pm$ 0e+00 & 6.47e+0 \\
ETTh2 & 1.91e+0 $\pm$ 2e-2 & 2.48e+0 $\pm$ 2e-2 & 3.45e+0 $\pm$ 4e-2 & 4.39e+0 $\pm$ 7e-2 & 2.16e+1 \\
Pulse & 2.59e-16 $\pm$ 3e-16 & 2.54e-16 $\pm$ 3e-16 & 2.56e-16 $\pm$ 3e-16 & 2.67e-16 $\pm$ 3e-16 & 3.10e-2 \\
PEMS04 & 6.75e+1 $\pm$ 4e+0 & 1.11e+2 $\pm$ 1e+1 & 2.06e+2 $\pm$ 3e+1 & 3.19e+2 $\pm$ 3e+1 & 1.13e+4 \\
PEMS03 & 7.27e+1 $\pm$ 4e+0 & 1.33e+2 $\pm$ 4e+0 & 2.34e+2 $\pm$ 1e+1 & 3.58e+2 $\pm$ 1e+1 & 8.51e+3 \\
PEMS-BAY & 1.58e+0 $\pm$ 2e-1 & 3.34e+0 $\pm$ 3e-1 & 7.87e+0 $\pm$ 8e-1 & 1.67e+1 $\pm$ 0e+00 & 2.72e+1 \\
BeijingAirQuality & 3.50e+2 $\pm$ 6e+0 & 5.55e+2 $\pm$ 1e+1 & 7.43e+2 $\pm$ 1e+1 & 8.73e+2 $\pm$ 7e-1 & 1.30e+3 \\
Weather & 1.06e+2 $\pm$ 2e+0 & 1.34e+2 $\pm$ 3e+0 & 1.74e+2 $\pm$ 8e+0 & 2.11e+2 $\pm$ 0e+00 & 3.13e+3 \\
ETTh1 & 1.30e+0 $\pm$ 1e-2 & 1.42e+0 $\pm$ 3e-2 & 1.56e+0 $\pm$ 1e-2 & 1.69e+0 $\pm$ 4e-3 & 6.43e+0 \\
ETTm2 & 1.04e+0 $\pm$ 1e-2 & 1.42e+0 $\pm$ 3e-2 & 1.96e+0 $\pm$ 3e-2 & 2.55e+0 $\pm$ 0e+00 & 2.16e+1 \\
PEMS07 & 9.86e+1 $\pm$ 3e+0 & 1.48e+2 $\pm$ 1e+1 & 2.95e+2 $\pm$ 2e+1 & 4.84e+2 $\pm$ 1e+1 & 1.67e+4 \\
Electricity & 1.51e+4 $\pm$ 7e+2 & 1.78e+4 $\pm$ 4e+2 & 2.69e+4 $\pm$ 5e+2 & 4.40e+4 $\pm$ 3e+2 & 9.85e+5 \\
PEMS08 & 8.40e+1 $\pm$ 4e+0 & 1.56e+2 $\pm$ 7e+0 & 3.04e+2 $\pm$ 2e+1 & 4.85e+2 $\pm$ 3e+1 & 8.45e+3 \\
\bottomrule
\end{tabular}
    \caption{MLP New: The unscaled mean squared error of the best performing novel strategy in \textit{Stratify} is shown for each task and horizon. We show the mean MSE over three seeds with standard error in $\pm$. The `MFE' column shows the MSE of a forecast predicting the mean value of the time series. This is a useful benchmark to understand the scale of the errors across the datasets.}
    \label{tab:my_label}
\end{table}

\begin{table}[hbt!]
    \centering
 \tiny
\begin{tabular}{llllll}
\toprule
Dataset name & 10 & 20 & 40 & 80 & MFE \\
\midrule
Traffic & 2.07e-5 $\pm$ 4e-7 & 2.90e-5 $\pm$ 3e-7 & 3.44e-5 $\pm$ 2e-7 & 4.03e-5 $\pm$ 6e-7 & 1.00e-3 \\
METR-LA & 8.38e+1 $\pm$ 3e-1 & 1.15e+2 $\pm$ 6e-1 & 1.35e+2 $\pm$ 2e-1 & 1.59e+2 $\pm$ 0e+00 & 2.24e+2 \\
Illness & 5.60e+8 $\pm$ 9e+6 & 6.02e+8 $\pm$ 3e+7 & 1.25e+9 $\pm$ 1e+7 & 1.26e+9 $\pm$ 1e+7 & 2.72e+9 \\
mg\_10000 & 8.66e-5 $\pm$ 4e-5 & 5.66e-5 $\pm$ 2e-7 & 7.14e-5 $\pm$ 10e-7 & 1.51e-4 $\pm$ 9e-6 & 1.20e-1 \\
ExchangeRate & 4.62e-5 $\pm$ 2e-6 & 7.68e-5 $\pm$ 4e-6 & 1.17e-4 $\pm$ 2e-6 & 1.94e-4 $\pm$ 2e-6 & 6.00e-3 \\
ETTm1 & 6.76e-1 $\pm$ 1e-3 & 1.00e+0 $\pm$ 2e-3 & 1.32e+0 $\pm$ 4e-3 & 1.51e+0 $\pm$ 5e-3 & 6.47e+0 \\
ETTh2 & 3.24e+0 $\pm$ 3e-2 & 3.97e+0 $\pm$ 7e-2 & 4.48e+0 $\pm$ 5e-2 & 5.62e+0 $\pm$ 6e-2 & 2.16e+1 \\
Pulse & 0.00e+00 $\pm$ 0e+00 & 0.00e+00 $\pm$ 0e+00 & 0.00e+00 $\pm$ 0e+00 & 0.00e+00 $\pm$ 0e+00 & 3.10e-2 \\
PEMS04 & 4.18e+1 $\pm$ 8e-2 & 6.02e+1 $\pm$ 5e-1 & 8.69e+1 $\pm$ 2e+0 & 1.40e+2 $\pm$ 10e-1 & 1.13e+4 \\
PEMS03 & 4.47e+1 $\pm$ 1e-1 & 7.39e+1 $\pm$ 9e-1 & 1.10e+2 $\pm$ 6e-1 & 1.45e+2 $\pm$ 5e-1 & 8.51e+3 \\
PEMS-BAY & 3.76e-1 $\pm$ 2e-3 & 8.87e-1 $\pm$ 9e-3 & 2.01e+0 $\pm$ 1e-3 & 3.12e+0 $\pm$ 3e-2 & 2.72e+1 \\
BeijingAirQuality & 3.56e+2 $\pm$ 1e+0 & 5.42e+2 $\pm$ 2e+0 & 7.56e+2 $\pm$ 2e+0 & 9.46e+2 $\pm$ 0e+00 & 1.30e+3 \\
Weather & 9.03e+1 $\pm$ 6e-2 & 1.13e+2 $\pm$ 1e+0 & 1.36e+2 $\pm$ 5e-1 & 1.51e+2 $\pm$ 0e+00 & 3.13e+3 \\
ETTh1 & 1.27e+0 $\pm$ 1e-2 & 1.43e+0 $\pm$ 1e-2 & 1.57e+0 $\pm$ 2e-3 & 1.69e+0 $\pm$ 1e-2 & 6.43e+0 \\
ETTm2 & 1.46e+0 $\pm$ 1e-2 & 1.99e+0 $\pm$ 6e-2 & 2.99e+0 $\pm$ 3e-2 & 3.88e+0 $\pm$ 1e-1 & 2.16e+1 \\
PEMS07 & 1.69e+1 $\pm$ 1e-1 & 2.44e+1 $\pm$ 1e-1 & 4.08e+1 $\pm$ 2e-1 & 5.09e+1 $\pm$ 7e-1 & 1.67e+4 \\
Electricity & 1.35e+4 $\pm$ 2e+2 & 1.59e+4 $\pm$ 8e+1 & 2.32e+4 $\pm$ 6e+2 & 3.64e+4 $\pm$ 4e+2 & 9.85e+5 \\
PEMS08 & 1.47e+1 $\pm$ 5e-2 & 2.08e+1 $\pm$ 1e-1 & 3.20e+1 $\pm$ 3e-1 & 4.10e+1 $\pm$ 5e-1 & 8.45e+3 \\
\bottomrule
\end{tabular}

    \caption{RF New: The unscaled mean squared error of the best performing novel strategy in \textit{Stratify} is shown for each task and horizon. We show the mean MSE over three seeds with standard error in $\pm$. The `MFE' column shows the MSE of a forecast predicting the mean value of the time series. This is a useful benchmark to understand the scale of the errors across the datasets.}
    \label{tab:my_label}
\end{table}

\begin{table}[hbt!]
    \centering
    \tiny
\begin{tabular}{llllll}
\toprule
Dataset name & 10 & 20 & 40 & 80 & MFE \\
\midrule
Traffic & 8.91e-4 $\pm$ 3e-4 & 1.01e-3 $\pm$ 2e-4 & 1.13e-3 $\pm$ 10e-5 & 8.91e-4 $\pm$ 3e-4 & 1.00e-3 \\
METR-LA & 8.10e+1 $\pm$ 4e-1 & 1.21e+2 $\pm$ 1e+0 & 1.55e+2 $\pm$ 1e+0 & 1.84e+2 $\pm$ 1e+0 & 2.24e+2 \\
Illness & 2.52e+10 $\pm$ 7e+4 & 2.52e+10 $\pm$ 3e+4 & 2.43e+10 $\pm$ 9e+3 & 2.14e+10 $\pm$ 7e+3 & 2.72e+9 \\
mg\_10000 & 2.23e-2 $\pm$ 2e-2 & 5.32e-3 $\pm$ 4e-3 & 1.22e-2 $\pm$ 9e-3 & 3.40e-2 $\pm$ 2e-2 & 1.20e-1 \\
ExchangeRate & 1.35e-3 $\pm$ 2e-3 & 6.02e-4 $\pm$ 4e-4 & 4.20e-3 $\pm$ 8e-4 & 3.18e-4 $\pm$ 1e-4 & 6.00e-3 \\
ETTm1 & 1.38e+0 $\pm$ 7e-2 & 2.61e+0 $\pm$ 1e-1 & 3.30e+0 $\pm$ 2e-1 & 3.93e+0 $\pm$ 3e-2 & 6.47e+0 \\
ETTh2 & 4.32e+0 $\pm$ 7e-2 & 5.85e+0 $\pm$ 3e-1 & 6.30e+0 $\pm$ 5e-2 & 6.14e+0 $\pm$ 7e-2 & 2.16e+1 \\
Pulse & 3.12e-2 $\pm$ 9e-6 & 3.12e-2 $\pm$ 9e-6 & 2.76e-2 $\pm$ 1e-3 & 1.99e-2 $\pm$ 6e-3 & 3.10e-2 \\
PEMS04 & 2.29e+2 $\pm$ 4e+1 & 9.31e+2 $\pm$ 2e+2 & 2.97e+3 $\pm$ 1e+2 & 5.36e+3 $\pm$ 5e+2 & 1.13e+4 \\
PEMS03 & 2.69e+2 $\pm$ 2e+1 & 8.34e+2 $\pm$ 7e+1 & 2.52e+3 $\pm$ 2e+2 & 3.59e+3 $\pm$ 3e+2 & 8.51e+3 \\
PEMS-BAY & 1.66e+0 $\pm$ 5e-1 & 5.49e+0 $\pm$ 1e-2 & 1.08e+1 $\pm$ 1e-1 & 1.62e+1 $\pm$ 4e-2 & 2.72e+1 \\
BeijingAirQuality & 3.55e+2 $\pm$ 1e+0 & 5.29e+2 $\pm$ 2e+0 & 7.06e+2 $\pm$ 2e+0 & 8.61e+2 $\pm$ 4e+0 & 1.30e+3 \\
Weather & 1.81e+2 $\pm$ 9e+0 & 3.47e+2 $\pm$ 7e+0 & 6.57e+2 $\pm$ 2e+0 & 1.05e+3 $\pm$ 5e+1 & 3.13e+3 \\
ETTh1 & 3.42e+0 $\pm$ 2e-1 & 3.18e+0 $\pm$ 8e-1 & 2.90e+0 $\pm$ 7e-1 & 3.73e+0 $\pm$ 2e-1 & 6.43e+0 \\
ETTm2 & 1.65e+0 $\pm$ 7e-2 & 2.43e+0 $\pm$ 8e-2 & 4.46e+0 $\pm$ 4e-2 & 5.18e+0 $\pm$ 2e-1 & 2.16e+1 \\
PEMS07 & 4.55e+2 $\pm$ 10e+1 & 1.76e+3 $\pm$ 1e+2 & 4.36e+3 $\pm$ 5e+2 & 7.82e+3 $\pm$ 6e+2 & 1.67e+4 \\
Electricity & 1.54e+6 $\pm$ 7e+2 & 1.54e+6 $\pm$ 4e+1 & 1.53e+6 $\pm$ 1e+2 & 1.53e+6 $\pm$ 2e+1 & 9.85e+5 \\
PEMS08 & 2.22e+2 $\pm$ 4e+1 & 8.38e+2 $\pm$ 2e+1 & 2.20e+3 $\pm$ 1e+2 & 4.70e+3 $\pm$ 1e+2 & 8.45e+3 \\
\bottomrule
\end{tabular}
    \caption{RNN New: The unscaled mean squared error of the best performing novel strategy in \textit{Stratify} is shown for each task and horizon. We show the mean MSE over three seeds with standard error in $\pm$. The `MFE' column shows the MSE of a forecast predicting the mean value of the time series. This is a useful benchmark to understand the scale of the errors across the datasets.}
    \label{tab:my_label}
\end{table}

\begin{table}[hbt!]
    \centering
 \tiny
\begin{tabular}{llllll}
\toprule
Dataset name & 10 & 20 & 40 & 80 & MFE \\
\midrule
Traffic & 4.43e-5 $\pm$ 6e-6 & 4.85e-5 $\pm$ 2e-6 & 6.68e-5 $\pm$ 7e-6 & 7.94e-5 $\pm$ 2e-5 & 1.00e-3 \\
METR-LA & 7.64e+1 $\pm$ 9e-1 & 1.06e+2 $\pm$ 2e+0 & 1.30e+2 $\pm$ 10e-1 & 1.62e+2 $\pm$ 1e+0 & 2.24e+2 \\
Illness & 2.52e+10 $\pm$ 3e+4 & 2.52e+10 $\pm$ 1e+4 & 2.43e+10 $\pm$ 6e+3 & 2.14e+10 $\pm$ 3e+3 & 2.72e+9 \\
mg\_10000 & 6.19e-5 $\pm$ 5e-5 & 3.84e-5 $\pm$ 1e-5 & 3.29e-5 $\pm$ 9e-6 & 1.35e-4 $\pm$ 4e-5 & 1.20e-1 \\
ExchangeRate & 3.12e-5 $\pm$ 4e-7 & 5.61e-5 $\pm$ 5e-7 & 9.72e-5 $\pm$ 4e-7 & 1.74e-4 $\pm$ 2e-6 & 6.00e-3 \\
ETTm1 & 8.11e-1 $\pm$ 3e-2 & 1.25e+0 $\pm$ 4e-2 & 1.63e+0 $\pm$ 8e-3 & 1.79e+0 $\pm$ 4e-2 & 6.47e+0 \\
ETTh2 & 4.20e+0 $\pm$ 7e-1 & 4.95e+0 $\pm$ 1e+0 & 5.35e+0 $\pm$ 8e-1 & 6.28e+0 $\pm$ 1e+0 & 2.16e+1 \\
Pulse & 1.76e-2 $\pm$ 1e-2 & 7.91e-3 $\pm$ 1e-2 & 3.37e-15 $\pm$ 1e-16 & 3.45e-15 $\pm$ 1e-15 & 3.10e-2 \\
PEMS04 & 7.10e+1 $\pm$ 7e+0 & 1.21e+2 $\pm$ 2e+1 & 2.63e+2 $\pm$ 4e+0 & 3.99e+2 $\pm$ 2e+0 & 1.13e+4 \\
PEMS03 & 8.15e+1 $\pm$ 3e+0 & 1.24e+2 $\pm$ 3e+0 & 2.17e+2 $\pm$ 9e+0 & 3.55e+2 $\pm$ 6e+0 & 8.51e+3 \\
PEMS-BAY & 7.14e-1 $\pm$ 2e-2 & 1.53e+0 $\pm$ 3e-1 & 3.24e+0 $\pm$ 5e-1 & 6.18e+0 $\pm$ 2e+0 & 2.72e+1 \\
BeijingAirQuality & 3.50e+2 $\pm$ 5e-1 & 5.26e+2 $\pm$ 4e+0 & 7.00e+2 $\pm$ 9e+0 & 8.35e+2 $\pm$ 1e+1 & 1.30e+3 \\
Weather & 1.82e+2 $\pm$ 3e+0 & 3.42e+2 $\pm$ 2e+1 & 5.16e+2 $\pm$ 1e+2 & 6.03e+2 $\pm$ 6e+1 & 3.13e+3 \\
ETTh1 & 1.50e+0 $\pm$ 3e-2 & 1.67e+0 $\pm$ 4e-2 & 1.70e+0 $\pm$ 5e-2 & 1.78e+0 $\pm$ 8e-2 & 6.43e+0 \\
ETTm2 & 1.44e+0 $\pm$ 5e-2 & 2.13e+0 $\pm$ 1e-1 & 3.44e+0 $\pm$ 1e-1 & 4.58e+0 $\pm$ 2e-1 & 2.16e+1 \\
PEMS07 & 1.16e+2 $\pm$ 2e+0 & 2.14e+2 $\pm$ 3e+1 & 4.04e+2 $\pm$ 4e+1 & 5.07e+2 $\pm$ 6e+1 & 1.67e+4 \\
Electricity & 1.54e+6 $\pm$ 2e+3 & 1.54e+6 $\pm$ 3e+2 & 1.53e+6 $\pm$ 1e+2 & 1.53e+6 $\pm$ 3e+2 & 9.85e+5 \\
PEMS08 & 7.61e+1 $\pm$ 1e+1 & 1.52e+2 $\pm$ 1e+1 & 2.95e+2 $\pm$ 5e+1 & 4.44e+2 $\pm$ 3e+1 & 8.45e+3 \\
\bottomrule
\end{tabular}

    \caption{LSTM New: The unscaled mean squared error of the best performing novel strategy in \textit{Stratify} is shown for each task and horizon. We show the mean MSE over three seeds with standard error in $\pm$. The `MFE' column shows the MSE of a forecast predicting the mean value of the time series. This is a useful benchmark to understand the scale of the errors across the datasets.}
    \label{tab:my_label}
\end{table}

\begin{table}[hbt!]
    \centering
 \tiny
\begin{tabular}{llllll}
\toprule
Dataset name & 10 & 20 & 40 & 80 & MFE \\
\midrule
Traffic & 1.08e-3 $\pm$ 1e-4 & 9.88e-4 $\pm$ 1e-5 & 9.53e-4 $\pm$ 3e-5 & 1.03e-3 $\pm$ 9e-5 & 1.00e-3 \\
METR-LA & 2.16e+2 $\pm$ 10e+0 & 1.84e+2 $\pm$ 3e+1 & 1.86e+2 $\pm$ 5e+0 & 2.08e+2 $\pm$ 1e+1 & 2.24e+2 \\
Illness & 2.52e+10 $\pm$ 7e+4 & 2.52e+10 $\pm$ 3e+4 & 2.43e+10 $\pm$ 2e+4 & 2.14e+10 $\pm$ 2e+4 & 2.72e+9 \\
mg\_10000 & 7.71e-2 $\pm$ 5e-3 & 7.20e-2 $\pm$ 3e-4 & 7.88e-2 $\pm$ 5e-4 & 8.53e-2 $\pm$ 2e-4 & 1.20e-1 \\
ExchangeRate & 7.93e-3 $\pm$ 4e-3 & 1.69e-3 $\pm$ 4e-4 & 8.71e-4 $\pm$ 1e-4 & 7.55e-4 $\pm$ 8e-5 & 6.00e-3 \\
ETTm1 & 2.89e+0 $\pm$ 1e-1 & 3.73e+0 $\pm$ 6e-2 & 4.20e+0 $\pm$ 10e-2 & 4.30e+0 $\pm$ 1e-1 & 6.47e+0 \\
ETTh2 & 2.72e+1 $\pm$ 1e+1 & 1.79e+1 $\pm$ 9e+0 & 1.27e+1 $\pm$ 2e+0 & 9.43e+0 $\pm$ 2e+0 & 2.16e+1 \\
Pulse & 3.12e-2 $\pm$ 1e-5 & 3.12e-2 $\pm$ 2e-5 & 3.10e-2 $\pm$ 4e-4 & 3.11e-2 $\pm$ 2e-4 & 3.10e-2 \\
PEMS04 & 1.06e+4 $\pm$ 10e+1 & 9.81e+3 $\pm$ 3e+1 & 8.46e+3 $\pm$ 5e+2 & 5.92e+3 $\pm$ 5e+1 & 1.13e+4 \\
PEMS03 & 8.08e+3 $\pm$ 3e+2 & 7.37e+3 $\pm$ 9e+1 & 6.39e+3 $\pm$ 2e+2 & 4.35e+3 $\pm$ 7e+1 & 8.51e+3 \\
PEMS-BAY & 2.16e+1 $\pm$ 2e-2 & 2.21e+1 $\pm$ 5e-1 & 2.20e+1 $\pm$ 1e-1 & 2.26e+1 $\pm$ 1e-1 & 2.72e+1 \\
BeijingAirQuality & 1.01e+3 $\pm$ 2e+0 & 1.01e+3 $\pm$ 2e+1 & 1.03e+3 $\pm$ 5e+0 & 1.03e+3 $\pm$ 4e+0 & 1.30e+3 \\
Weather & 1.04e+3 $\pm$ 3e+0 & 8.99e+2 $\pm$ 2e+2 & 9.17e+2 $\pm$ 2e+2 & 1.05e+3 $\pm$ 0e+00 & 3.13e+3 \\
ETTh1 & 4.21e+0 $\pm$ 2e-2 & 4.36e+0 $\pm$ 2e-2 & 4.31e+0 $\pm$ 1e-2 & 4.34e+0 $\pm$ 9e-3 & 6.43e+0 \\
ETTm2 & 6.31e+0 $\pm$ 2e-1 & 5.73e+0 $\pm$ 8e-2 & 5.88e+0 $\pm$ 1e-1 & 5.96e+0 $\pm$ 2e-1 & 2.16e+1 \\
PEMS07 & 1.41e+4 $\pm$ 1e+2 & 1.30e+4 $\pm$ 2e+2 & 1.03e+4 $\pm$ 9e+1 & 7.20e+3 $\pm$ 4e+2 & 1.67e+4 \\
Electricity & 1.37e+6 $\pm$ 9e+3 & 1.36e+6 $\pm$ 5e+3 & 1.36e+6 $\pm$ 8e+3 & 1.36e+6 $\pm$ 6e+3 & 9.85e+5 \\
PEMS08 & 8.39e+3 $\pm$ 5e+1 & 7.99e+3 $\pm$ 5e+1 & 7.17e+3 $\pm$ 10e+2 & 4.83e+3 $\pm$ 4e+1 & 8.45e+3 \\
\bottomrule
\end{tabular}

    \caption{Transformer New: The unscaled mean squared error of the best performing novel strategy in \textit{Stratify} is shown for each task and horizon. We show the mean MSE over three seeds with standard error in $\pm$. The `MFE' column shows the MSE of a forecast predicting the mean value of the time series. This is a useful benchmark to understand the scale of the errors across the datasets.}
    \label{tab:my_label}
\end{table}



\begin{table}[hbt!]
    \centering
    \tiny
\begin{tabular}{llllll}
\toprule
Dataset name & 10 & 20 & 40 & 80 & MFE \\
\midrule
Traffic & 3.68e-5 $\pm$ 7e-7 & 3.93e-5 $\pm$ 2e-6 & 4.50e-5 $\pm$ 2e-6 & 5.04e-5 $\pm$ 2e-6 & 1.00e-3 \\
METR-LA & 8.12e+1 $\pm$ 2e-1 & 1.16e+2 $\pm$ 3e-1 & 1.47e+2 $\pm$ 1e+0 & 1.81e+2 $\pm$ 2e+0 & 2.24e+2 \\
Illness & 1.77e+10 $\pm$ 2e+9 & 1.80e+10 $\pm$ 2e+9 & 1.76e+10 $\pm$ 2e+9 & 1.52e+10 $\pm$ 3e+9 & 2.72e+9 \\
mg\_10000 & 5.67e-5 $\pm$ 6e-5 & 2.48e-5 $\pm$ 1e-6 & 4.04e-5 $\pm$ 8e-6 & 1.38e-4 $\pm$ 5e-5 & 1.20e-1 \\
ExchangeRate & 1.07e-4 $\pm$ 1e-5 & 1.33e-4 $\pm$ 7e-6 & 1.57e-4 $\pm$ 1e-5 & 2.44e-4 $\pm$ 7e-6 & 6.00e-3 \\
ETTm1 & 6.54e-1 $\pm$ 6e-4 & 9.85e-1 $\pm$ 2e-3 & 1.29e+0 $\pm$ 1e-2 & 1.48e+0 $\pm$ 0e+00 & 6.47e+0 \\
ETTh2 & 1.91e+0 $\pm$ 2e-2 & 2.48e+0 $\pm$ 2e-2 & 3.45e+0 $\pm$ 4e-2 & 4.39e+0 $\pm$ 7e-2 & 2.16e+1 \\
Pulse & 2.59e-16 $\pm$ 3e-16 & 2.54e-16 $\pm$ 3e-16 & 2.56e-16 $\pm$ 3e-16 & 2.67e-16 $\pm$ 3e-16 & 3.10e-2 \\
PEMS04 & 6.75e+1 $\pm$ 4e+0 & 1.11e+2 $\pm$ 1e+1 & 2.06e+2 $\pm$ 3e+1 & 3.19e+2 $\pm$ 3e+1 & 1.13e+4 \\
PEMS03 & 7.27e+1 $\pm$ 4e+0 & 1.33e+2 $\pm$ 4e+0 & 2.34e+2 $\pm$ 1e+1 & 3.58e+2 $\pm$ 1e+1 & 8.51e+3 \\
PEMS-BAY & 1.58e+0 $\pm$ 2e-1 & 3.34e+0 $\pm$ 3e-1 & 7.87e+0 $\pm$ 8e-1 & 1.67e+1 $\pm$ 0e+00 & 2.72e+1 \\
BeijingAirQuality & 3.50e+2 $\pm$ 6e+0 & 5.55e+2 $\pm$ 1e+1 & 7.43e+2 $\pm$ 1e+1 & 8.73e+2 $\pm$ 7e-1 & 1.30e+3 \\
Weather & 1.06e+2 $\pm$ 2e+0 & 1.34e+2 $\pm$ 3e+0 & 1.74e+2 $\pm$ 8e+0 & 2.11e+2 $\pm$ 0e+00 & 3.13e+3 \\
ETTh1 & 1.30e+0 $\pm$ 1e-2 & 1.42e+0 $\pm$ 3e-2 & 1.56e+0 $\pm$ 1e-2 & 1.69e+0 $\pm$ 4e-3 & 6.43e+0 \\
ETTm2 & 1.04e+0 $\pm$ 1e-2 & 1.42e+0 $\pm$ 3e-2 & 1.96e+0 $\pm$ 3e-2 & 2.55e+0 $\pm$ 0e+00 & 2.16e+1 \\
PEMS07 & 9.86e+1 $\pm$ 3e+0 & 1.48e+2 $\pm$ 1e+1 & 2.95e+2 $\pm$ 2e+1 & 4.84e+2 $\pm$ 1e+1 & 1.67e+4 \\
Electricity & 1.51e+4 $\pm$ 7e+2 & 1.78e+4 $\pm$ 4e+2 & 2.69e+4 $\pm$ 5e+2 & 4.40e+4 $\pm$ 3e+2 & 9.85e+5 \\
PEMS08 & 8.40e+1 $\pm$ 4e+0 & 1.56e+2 $\pm$ 7e+0 & 3.04e+2 $\pm$ 2e+1 & 4.85e+2 $\pm$ 3e+1 & 8.45e+3 \\
\bottomrule
\end{tabular}

    \caption{MLP Old: The unscaled mean squared error of the best performing existing strategy is shown for each task and horizon. We show the mean MSE over three seeds with standard error in $\pm$. The `MFE' column shows the MSE of a forecast predicting the mean value of the time series. This is a useful benchmark to understand the scale of the errors across the datasets.}
    \label{tab:my_label}
\end{table}

\begin{table}[hbt!]
    \centering
    \tiny
    \begin{tabular}{llllll}
\toprule
Dataset name & 10 & 20 & 40 & 80 & MFE \\
\midrule
Traffic & 2.13e-5 $\pm$ 5e-7 & 2.96e-5 $\pm$ 2e-7 & 3.51e-5 $\pm$ 4e-7 & 4.04e-5 $\pm$ 1e-6 & 1.00e-3 \\
METR-LA & 8.29e+1 $\pm$ 3e-1 & 1.14e+2 $\pm$ 2e-1 & 1.35e+2 $\pm$ 6e-1 & 1.61e+2 $\pm$ 0e+00 & 2.24e+2 \\
Illness & 9.46e+8 $\pm$ 3e+7 & 1.18e+9 $\pm$ 3e+7 & 1.90e+9 $\pm$ 1e+7 & 1.50e+9 $\pm$ 5e+6 & 2.72e+9 \\
mg\_10000 & 1.18e-4 $\pm$ 5e-5 & 7.62e-5 $\pm$ 8e-7 & 1.00e-4 $\pm$ 2e-6 & 2.08e-4 $\pm$ 1e-5 & 1.20e-1 \\
ExchangeRate & 4.51e-5 $\pm$ 2e-6 & 7.46e-5 $\pm$ 4e-6 & 1.14e-4 $\pm$ 1e-6 & 1.95e-4 $\pm$ 3e-6 & 6.00e-3 \\
ETTm1 & 6.87e-1 $\pm$ 2e-3 & 1.03e+0 $\pm$ 3e-3 & 1.37e+0 $\pm$ 1e-3 & 1.57e+0 $\pm$ 2e-3 & 6.47e+0 \\
ETTh2 & 3.24e+0 $\pm$ 4e-2 & 3.97e+0 $\pm$ 3e-2 & 4.54e+0 $\pm$ 6e-2 & 5.34e+0 $\pm$ 2e-2 & 2.16e+1 \\
Pulse & 0.00e+00 $\pm$ 0e+00 & 0.00e+00 $\pm$ 0e+00 & 0.00e+00 $\pm$ 0e+00 & 0.00e+00 $\pm$ 0e+00 & 3.10e-2 \\
PEMS04 & 4.37e+1 $\pm$ 9e-2 & 6.32e+1 $\pm$ 5e-1 & 9.05e+1 $\pm$ 2e+0 & 1.43e+2 $\pm$ 8e-1 & 1.13e+4 \\
PEMS03 & 4.62e+1 $\pm$ 2e-1 & 7.66e+1 $\pm$ 9e-1 & 1.13e+2 $\pm$ 6e-1 & 1.52e+2 $\pm$ 4e-1 & 8.51e+3 \\
PEMS-BAY & 3.80e-1 $\pm$ 5e-3 & 8.85e-1 $\pm$ 5e-3 & 2.06e+0 $\pm$ 1e-3 & 3.16e+0 $\pm$ 3e-2 & 2.72e+1 \\
BeijingAirQuality & 3.60e+2 $\pm$ 8e-1 & 5.43e+2 $\pm$ 2e+0 & 7.57e+2 $\pm$ 2e+0 & 9.44e+2 $\pm$ 0e+00 & 1.30e+3 \\
Weather & 9.03e+1 $\pm$ 6e-2 & 1.13e+2 $\pm$ 1e+0 & 1.36e+2 $\pm$ 6e-1 & 1.51e+2 $\pm$ 0e+00 & 3.13e+3 \\
ETTh1 & 1.33e+0 $\pm$ 1e-2 & 1.49e+0 $\pm$ 1e-2 & 1.66e+0 $\pm$ 6e-3 & 1.79e+0 $\pm$ 3e-3 & 6.43e+0 \\
ETTm2 & 1.42e+0 $\pm$ 3e-3 & 1.98e+0 $\pm$ 7e-2 & 2.72e+0 $\pm$ 7e-2 & 3.68e+0 $\pm$ 3e-2 & 2.16e+1 \\
PEMS07 & 1.78e+1 $\pm$ 2e-1 & 2.59e+1 $\pm$ 2e-1 & 4.55e+1 $\pm$ 3e-1 & 5.65e+1 $\pm$ 1e+0 & 1.67e+4 \\
Electricity & 1.47e+4 $\pm$ 6e+1 & 1.76e+4 $\pm$ 4e+1 & 2.52e+4 $\pm$ 2e+2 & 3.81e+4 $\pm$ 2e+2 & 9.85e+5 \\
PEMS08 & 1.65e+1 $\pm$ 9e-2 & 2.31e+1 $\pm$ 1e-1 & 3.54e+1 $\pm$ 3e-1 & 4.60e+1 $\pm$ 7e-1 & 8.45e+3 \\
\bottomrule
\end{tabular}

    \caption{RF Old: The unscaled mean squared error of the best performing existing strategy is shown for each task and horizon. We show the mean MSE over three seeds with standard error in $\pm$. The `MFE' column shows the MSE of a forecast predicting the mean value of the time series. This is a useful benchmark to understand the scale of the errors across the datasets.}
    \label{tab:my_label}
\end{table}

\begin{table}[hbt!]
    \centering
    \tiny
    \begin{tabular}{llllll}
\toprule
Dataset name & 10 & 20 & 40 & 80 & MFE \\
\midrule
Traffic & 1.20e-3 $\pm$ 2e-5 & 1.25e-3 $\pm$ 1e-5 & 1.21e-3 $\pm$ 6e-5 & 1.25e-3 $\pm$ 2e-5 & 1.00e-3 \\
METR-LA & 8.12e+1 $\pm$ 1e+0 & 1.23e+2 $\pm$ 4e+0 & 1.59e+2 $\pm$ 4e+0 & 1.87e+2 $\pm$ 2e+0 & 2.24e+2 \\
Illness & 2.52e+10 $\pm$ 3e+4 & 2.52e+10 $\pm$ 2e+4 & 2.43e+10 $\pm$ 8e+3 & 2.14e+10 $\pm$ 5e+3 & 2.72e+9 \\
mg\_10000 & 4.25e-2 $\pm$ 4e-2 & 3.11e-2 $\pm$ 4e-2 & 3.56e-2 $\pm$ 4e-2 & 5.39e-2 $\pm$ 4e-2 & 1.20e-1 \\
ExchangeRate & 1.07e-2 $\pm$ 1e-3 & 1.07e-2 $\pm$ 1e-3 & 5.32e-3 $\pm$ 4e-3 & 5.89e-3 $\pm$ 4e-3 & 6.00e-3 \\
ETTm1 & 1.55e+0 $\pm$ 2e-2 & 3.02e+0 $\pm$ 1e-2 & 3.96e+0 $\pm$ 2e-1 & 4.36e+0 $\pm$ 2e-2 & 6.47e+0 \\
ETTh2 & 1.39e+1 $\pm$ 1e+1 & 1.52e+1 $\pm$ 9e+0 & 1.29e+1 $\pm$ 5e+0 & 1.49e+1 $\pm$ 7e+0 & 2.16e+1 \\
Pulse & 3.14e-2 $\pm$ 1e-4 & 3.13e-2 $\pm$ 4e-5 & 3.05e-2 $\pm$ 7e-4 & 3.09e-2 $\pm$ 2e-4 & 3.10e-2 \\
PEMS04 & 3.48e+2 $\pm$ 4e+1 & 1.08e+3 $\pm$ 2e+1 & 3.71e+3 $\pm$ 1e+2 & 9.06e+3 $\pm$ 1e+2 & 1.13e+4 \\
PEMS03 & 2.79e+2 $\pm$ 2e+1 & 8.51e+2 $\pm$ 6e+1 & 2.68e+3 $\pm$ 2e+2 & 6.04e+3 $\pm$ 4e+2 & 8.51e+3 \\
PEMS-BAY & 4.70e+0 $\pm$ 2e+0 & 1.02e+1 $\pm$ 1e+0 & 2.14e+1 $\pm$ 2e+0 & 2.45e+1 $\pm$ 2e-2 & 2.72e+1 \\
BeijingAirQuality & 3.60e+2 $\pm$ 4e+0 & 5.34e+2 $\pm$ 5e+0 & 7.19e+2 $\pm$ 7e+0 & 8.93e+2 $\pm$ 2e+1 & 1.30e+3 \\
Weather & 1.09e+3 $\pm$ 8e-1 & 1.08e+3 $\pm$ 1e+1 & 1.07e+3 $\pm$ 5e+0 & 1.07e+3 $\pm$ 3e+0 & 3.13e+3 \\
ETTh1 & 3.64e+0 $\pm$ 10e-2 & 4.12e+0 $\pm$ 8e-2 & 4.43e+0 $\pm$ 2e-2 & 4.50e+0 $\pm$ 2e-2 & 6.43e+0 \\
ETTm2 & 2.09e+0 $\pm$ 5e-1 & 3.21e+0 $\pm$ 5e-1 & 5.19e+0 $\pm$ 3e-1 & 5.93e+0 $\pm$ 6e-1 & 2.16e+1 \\
PEMS07 & 7.31e+3 $\pm$ 5e+3 & 8.13e+3 $\pm$ 4e+3 & 1.04e+4 $\pm$ 2e+3 & 1.43e+4 $\pm$ 7e+2 & 1.67e+4 \\
Electricity & 2.35e+6 $\pm$ 4e+2 & 2.35e+6 $\pm$ 1e+2 & 2.34e+6 $\pm$ 9e+1 & 2.33e+6 $\pm$ 2e+2 & 9.85e+5 \\
PEMS08 & 7.24e+3 $\pm$ 10e+2 & 7.45e+3 $\pm$ 10e+2 & 7.80e+3 $\pm$ 9e+2 & 7.96e+3 $\pm$ 3e+2 & 8.45e+3 \\
\bottomrule
\end{tabular}

    \caption{RNN Old: The unscaled mean squared error of the best performing existing strategy is shown for each task and horizon. We show the mean MSE over three seeds with standard error in $\pm$. The `MFE' column shows the MSE of a forecast predicting the mean value of the time series. This is a useful benchmark to understand the scale of the errors across the datasets.}
    \label{tab:my_label}
\end{table}

\begin{table}[hbt!]
    \centering
    \tiny
\begin{tabular}{llllll}
\toprule
Dataset name & 10 & 20 & 40 & 80 & MFE \\
\midrule
Traffic & 4.71e-5 $\pm$ 4e-6 & 5.57e-5 $\pm$ 1e-6 & 6.94e-5 $\pm$ 7e-6 & 8.30e-5 $\pm$ 1e-5 & 1.00e-3 \\
METR-LA & 7.67e+1 $\pm$ 8e-1 & 1.11e+2 $\pm$ 3e+0 & 1.41e+2 $\pm$ 3e+0 & 1.73e+2 $\pm$ 4e-1 & 2.24e+2 \\
Illness & 2.52e+10 $\pm$ 5e+3 & 2.52e+10 $\pm$ 5e+3 & 2.43e+10 $\pm$ 5e+3 & 2.14e+10 $\pm$ 4e+3 & 2.72e+9 \\
mg\_10000 & 3.04e-4 $\pm$ 2e-4 & 1.92e-4 $\pm$ 3e-5 & 2.81e-4 $\pm$ 4e-5 & 5.94e-4 $\pm$ 2e-4 & 1.20e-1 \\
ExchangeRate & 3.37e-5 $\pm$ 1e-6 & 5.70e-5 $\pm$ 10e-7 & 9.74e-5 $\pm$ 6e-7 & 1.80e-4 $\pm$ 3e-6 & 6.00e-3 \\
ETTm1 & 8.60e-1 $\pm$ 4e-2 & 1.27e+0 $\pm$ 4e-2 & 1.64e+0 $\pm$ 1e-2 & 1.83e+0 $\pm$ 5e-2 & 6.47e+0 \\
ETTh2 & 4.57e+0 $\pm$ 4e-1 & 5.74e+0 $\pm$ 7e-1 & 6.49e+0 $\pm$ 8e-1 & 7.97e+0 $\pm$ 2e+0 & 2.16e+1 \\
Pulse & 2.75e-2 $\pm$ 2e-3 & 2.81e-2 $\pm$ 1e-3 & 8.11e-10 $\pm$ 1e-9 & 8.36e-10 $\pm$ 1e-9 & 3.10e-2 \\
PEMS04 & 1.11e+2 $\pm$ 5e+0 & 2.11e+2 $\pm$ 8e+0 & 4.19e+2 $\pm$ 7e+0 & 7.10e+2 $\pm$ 3e+1 & 1.13e+4 \\
PEMS03 & 1.06e+2 $\pm$ 8e+0 & 2.13e+2 $\pm$ 1e+1 & 3.98e+2 $\pm$ 2e+1 & 6.14e+2 $\pm$ 1e+1 & 8.51e+3 \\
PEMS-BAY & 9.19e-1 $\pm$ 6e-2 & 3.19e+0 $\pm$ 3e-1 & 1.03e+1 $\pm$ 1e+0 & 1.84e+1 $\pm$ 8e-1 & 2.72e+1 \\
BeijingAirQuality & 3.52e+2 $\pm$ 3e-1 & 5.29e+2 $\pm$ 4e+0 & 7.07e+2 $\pm$ 9e+0 & 8.52e+2 $\pm$ 2e+1 & 1.30e+3 \\
Weather & 1.91e+2 $\pm$ 8e+0 & 3.87e+2 $\pm$ 2e+1 & 6.55e+2 $\pm$ 3e+1 & 8.71e+2 $\pm$ 8e+1 & 3.13e+3 \\
ETTh1 & 1.60e+0 $\pm$ 6e-2 & 1.71e+0 $\pm$ 7e-2 & 1.86e+0 $\pm$ 3e-2 & 2.14e+0 $\pm$ 4e-3 & 6.43e+0 \\
ETTm2 & 1.52e+0 $\pm$ 4e-2 & 2.31e+0 $\pm$ 2e-1 & 3.38e+0 $\pm$ 10e-3 & 4.28e+0 $\pm$ 3e-2 & 2.16e+1 \\
PEMS07 & 1.43e+2 $\pm$ 4e+0 & 3.23e+2 $\pm$ 4e+1 & 7.34e+2 $\pm$ 1e+2 & 1.03e+3 $\pm$ 1e+2 & 1.67e+4 \\
Electricity & 2.51e+6 $\pm$ 5e+4 & 2.50e+6 $\pm$ 5e+4 & 2.47e+6 $\pm$ 5e+4 & 2.45e+6 $\pm$ 4e+4 & 9.85e+5 \\
PEMS08 & 1.35e+2 $\pm$ 5e+0 & 2.64e+2 $\pm$ 6e+0 & 5.46e+2 $\pm$ 3e+1 & 9.25e+2 $\pm$ 4e+1 & 8.45e+3 \\
\bottomrule
\end{tabular}

    \caption{LSTM Old: The unscaled mean squared error of the best performing existing strategy is shown for each task and horizon. We show the mean MSE over three seeds with standard error in $\pm$. The `MFE' column shows the MSE of a forecast predicting the mean value of the time series. This is a useful benchmark to understand the scale of the errors across the datasets.}
    \label{tab:my_label}
\end{table}

\begin{table}[hbt!]
    \centering
    \tiny
    \begin{tabular}{llllll}
\toprule
Dataset name & 10 & 20 & 40 & 80 & MFE \\
\midrule
Traffic & 1.27e-3 $\pm$ 4e-5 & 1.12e-3 $\pm$ 2e-5 & 1.11e-3 $\pm$ 6e-5 & 1.22e-3 $\pm$ 5e-5 & 1.00e-3 \\
METR-LA & 2.25e+2 $\pm$ 1e+0 & 2.26e+2 $\pm$ 2e+0 & 2.25e+2 $\pm$ 2e+0 & 2.14e+2 $\pm$ 1e+1 & 2.24e+2 \\
Illness & 2.52e+10 $\pm$ 8e+4 & 2.52e+10 $\pm$ 2e+4 & 2.43e+10 $\pm$ 3e+4 & 2.14e+10 $\pm$ 3e+4 & 2.72e+9 \\
mg\_10000 & 7.86e-2 $\pm$ 6e-3 & 7.33e-2 $\pm$ 10e-4 & 8.00e-2 $\pm$ 2e-4 & 8.81e-2 $\pm$ 2e-4 & 1.20e-1 \\
ExchangeRate & 1.30e-2 $\pm$ 6e-4 & 1.29e-2 $\pm$ 8e-4 & 1.14e-2 $\pm$ 2e-3 & 1.08e-2 $\pm$ 2e-3 & 6.00e-3 \\
ETTm1 & 3.10e+0 $\pm$ 2e-1 & 3.90e+0 $\pm$ 2e-1 & 4.25e+0 $\pm$ 9e-2 & 4.43e+0 $\pm$ 1e-1 & 6.47e+0 \\
ETTh2 & 5.07e+1 $\pm$ 5e-1 & 4.95e+1 $\pm$ 3e+0 & 4.76e+1 $\pm$ 3e+0 & 4.58e+1 $\pm$ 2e-1 & 2.16e+1 \\
Pulse & 3.13e-2 $\pm$ 5e-5 & 3.13e-2 $\pm$ 8e-6 & 3.10e-2 $\pm$ 4e-4 & 3.11e-2 $\pm$ 2e-4 & 3.10e-2 \\
PEMS04 & 1.11e+4 $\pm$ 3e+2 & 9.87e+3 $\pm$ 4e+1 & 9.19e+3 $\pm$ 7e+1 & 7.94e+3 $\pm$ 1e+3 & 1.13e+4 \\
PEMS03 & 8.21e+3 $\pm$ 3e+2 & 7.93e+3 $\pm$ 6e+2 & 7.66e+3 $\pm$ 8e+2 & 4.86e+3 $\pm$ 3e+2 & 8.51e+3 \\
PEMS-BAY & 2.27e+1 $\pm$ 1e+0 & 2.28e+1 $\pm$ 1e+0 & 2.24e+1 $\pm$ 6e-2 & 2.32e+1 $\pm$ 3e-1 & 2.72e+1 \\
BeijingAirQuality & 1.05e+3 $\pm$ 1e+1 & 1.05e+3 $\pm$ 1e+1 & 1.05e+3 $\pm$ 1e+1 & 1.04e+3 $\pm$ 3e+0 & 1.30e+3 \\
Weather & 9.91e+2 $\pm$ 3e+1 & 8.62e+2 $\pm$ 1e+2 & 8.64e+2 $\pm$ 1e+2 & 9.63e+2 $\pm$ 0e+00 & 3.13e+3 \\
ETTh1 & 4.25e+0 $\pm$ 2e-2 & 4.49e+0 $\pm$ 5e-2 & 4.44e+0 $\pm$ 1e-1 & 4.53e+0 $\pm$ 6e-2 & 6.43e+0 \\
ETTm2 & 1.92e+1 $\pm$ 1e+1 & 1.86e+1 $\pm$ 10e+0 & 1.86e+1 $\pm$ 8e+0 & 1.40e+1 $\pm$ 1e+1 & 2.16e+1 \\
PEMS07 & 1.48e+4 $\pm$ 1e+3 & 1.35e+4 $\pm$ 4e+2 & 1.29e+4 $\pm$ 7e+2 & 1.09e+4 $\pm$ 3e+3 & 1.67e+4 \\
Electricity & 1.37e+6 $\pm$ 7e+3 & 1.37e+6 $\pm$ 6e+3 & 1.36e+6 $\pm$ 5e+3 & 1.36e+6 $\pm$ 5e+3 & 9.85e+5 \\
PEMS08 & 8.54e+3 $\pm$ 1e+2 & 8.50e+3 $\pm$ 2e+2 & 7.34e+3 $\pm$ 1e+3 & 5.11e+3 $\pm$ 3e+2 & 8.45e+3 \\
\bottomrule
\end{tabular}

    \caption{Transformer Old: The unscaled mean squared error of the best performing existing strategy is shown for each task and horizon. We show the mean MSE over three seeds with standard error in $\pm$. The `MFE' column shows the MSE of a forecast predicting the mean value of the time series. This is a useful benchmark to understand the scale of the errors across the datasets.}
    \label{tab:my_label}
\end{table}

\

\end{appendices}



\end{document}